\documentclass[runningheads]{llncs}

\usepackage{eccv}

\usepackage{eccvabbrv}

\usepackage{graphicx}
\usepackage{booktabs}
\usepackage{graphicx}
\usepackage{verbatim}
\usepackage{float}
\usepackage{wrapfig}
\usepackage[table]{xcolor}

\usepackage[accsupp]{axessibility}  % Improves PDF readability for those with disabilities.

\usepackage{hyperref}

\usepackage{orcidlink}

\usepackage{shortbold}

\newcommand{\parnobf}[1]{\vspace{0.5mm} \par \noindent {\bf #1.}}
\newcommand{\method}{Z3D}

\definecolor{oodblue}{RGB}{240,248,255}

\newcommand{\ZThreeDVGGTDD}{\textsc{VGGT-DD}\xspace}
\newcommand{\ZThreeDWMDD}{\textsc{WM-DD}\xspace}
\newcommand{\ZThreeDVGGT}{\textsc{Z3D-VGGT}}
\newcommand{\ZThreeDWM}{\textsc{Z3D-WM}}
\newcommand{\VGGTLDI}{\textsc{VGGT-LDI}\xspace}
\newcommand{\WMTLDI}{\textsc{WM-LDI}\xspace}
\newcommand{\LVSMVGGT}{\textsc{LVSM+VGGT}\xspace}
\newcommand{\LVSMVWM}{\textsc{LVSM+WM}\xspace}

\newcommand{\VZThreeVGGTOmega}{\textsc{Z3D-VGGT-$\Omega$}\xspace}

\begin{document}
\title{Zero-Shot Novel Depth Synthesis Using 3D Foundation Models Scene Representations} 

\titlerunning{Zero-Shot Novel Depth Synthesis Using 3DFMs}

\author{Denis M. Akola\inst{1} \and
David F. Fouhey\inst{1, 2}}

\authorrunning{D. Akola and D. Fouhey}

\institute{New York University, Tandon School of Engineering,  1 MetroTech Center, Brooklyn, NY 11201, USA  \and
Courant Institute of Mathematical Sciences,  251 Mercer St, New York, NY 10012 }
\maketitle

\begin{abstract}
3D Foundation Models (3DFMs) such as VGGT have recently pushed the boundaries of 3D vision by predicting rich unified representations with feed-foward transformers. The scene representations learned by these models enable strong performance on multiple 3D vision tasks. In this paper, we investigate using their internal representations to infer 3D in the scene from new views. Our hypothesis is that in order to solve the task of 3D reconstruction, these models need to learn a representation that includes a large amount of general knowledge about 3D scenes. After showing that it is possible to decode hidden surfaces from internal 3DFM representations, we propose a method, Z3D, that estimates pointmaps in unseen views by doing latent diffusion on 3DFM representation. We show that Z3D can predict realistic depth maps for new views across multiple datasets. 
Project page: \url{https://akola-mbey-denis.github.io/Z3D-page}

\end{abstract}

\section{Introduction}
\label{sec:intro}
Understanding the full 3D structure of a scene from limited observations remains a fundamental challenge in computer vision. In many real-world settings like robot navigation, AR/VR and autonomous systems, we only observe partial views of a scene, yet must reason about the geometry that lies outside the visible surface, including regions occluded by foreground objects. Novel depth synthesis addresses this challenge. Given one or a few input views, the goal is to predict plausible and geometrically consistent depth for unseen viewpoints, including content hidden behind objects. Beyond view rendering, this capability enables physical interaction, planning, and spatial reasoning in incomplete environments, where missing geometry must be inferred rather than directly observed.

Recent advances in neural scene representations such as NeRF\cite{10.1145/3503250},3D Gaussian Splatting\cite{10.1145/3592433} and other neural methods\cite{Guizilini2023TowardsZS, Guizilini2024GRINZM, Guizilini_2025_CVPR, jin2025lvsm} have significantly improved novel view synthesis by optimizing continuous 3D representations from multi-view imagery. However, these methods primarily interpolate between observed views and are not designed to hallucinate geometry behind occlusions, especially in sparse-view settings. Prior works\cite{Guizilini_2025_CVPR, 10.1007/978-3-031-19824-3_15} on novel depth synthesis often jointly learn view and depth prediction from multi-view data, but progress toward generalizable depth synthesis from sparse inputs remains limited. More recently, generative models \cite{ke2023repurposing, Guizilini2024GRINZM} have been explored for depth prediction, but these often operate in 2D image space and lack strong 3D consistency priors, limiting their ability to produce coherent geometry across viewpoints. The core challenges stem from geometric ambiguity under limited observations and the difficulty of learning representations that transfer across diverse scenes.

The key insight of this paper is that the internal representations of pretrained 3D foundation models (3DFMs) \cite{wang2025vggt, liu2025worldmirror, dust3r_cvpr24, Yang_2025_Fast3R} can be effectively combined with diffusion-based models for generalizable novel-view depth synthesis. Rather than treating novel depth synthesis as pure reconstruction or view interpolation, we formulate it as a conditional generation task over scene geometry. Specifically, we use the intermediate representations of 3DFMs for conditioning as well as the latent space in which to do latent diffusion~\cite{rombach2021highresolution}. After diffusion, the predicted latents can be converted to geometry by the original decoder. By decoupling representation learning (which is taken care of by the 3DFM) from generative completion (handled by diffusion), our approach enables plausible hallucination of occluded or unobserved structure, and encourages multi-view consistency. To further support our premise that 3DFMs encode geometry beyond directly visible surfaces, we show that simply attaching a lightweight decoder to their features enables inference of hidden surfaces, demonstrating that the learned scene representations are well-suited for reasoning about occluded geometry.

We propose \method– \textbf{Z}ero-Shot \textbf{3D} \textbf{D}epth, a simple and effective method that leverages the rich geometric scene representations learned by 3DFMs in combination with a diffusion framework to generate novel-view depth maps. Z3D integrates these pretrained 3DFM representations with the relative pose of a novel view to predict consistent and accurate depth for novel view cameras. We evaluate our method on multiple datasets. We compare Z3D with state-of-art methods~\cite{jin2025lvsm} for novel view depth synthesis.

\section{Related Work}
We aim to generate consistent depth maps for novel views of a scene, given a sparse set of images (2 or more views). We first infer a geometric scene representation from 3DFMs and, using a diffusion model conditioned on this representation and the poses of the target views, predict novel-view depths.

\parnobf{Novel View and Depth Synthesis}
NeRF~\cite{10.1145/3503250} and its extensions \cite{yu2021pixelnerf, mueller2022instant, Niemeyer2021RegNeRFRN} render novel views of observed surfaces but often require many input images and per-scene optimization. In sparse view settings, they tend to produce incomplete geometry and cannot infer surfaces outside the original field of view or behind occlusions. In contrast, our method infers novel-view geometry from sparse inputs without per-scene optimization.
Methods such as PixelNeRF \cite{yu2021pixelnerf}, SRT \cite{srt22}, RUST \cite{10204153}, LVSM \cite{jin2025lvsm}, and RayZer \cite{Jiang_2025_ICCV} generalize across scenes but focus on photometric quality rather than accurate depth, limiting their ability to hallucinate unseen structure. Our approach explicitly models geometry, generating consistent depth maps beyond visible surfaces. CUT3R\cite{cut3r} is a stateful 3D reconstruction model capable of recovering scene geometry from RGB observations, predicting depth for novel views, and inferring unobserved scene regions by querying virtual camera viewpoints. Unlike CUT3R, our method incorporates pre-trained 3DFM representations within a diffusion framework. This design leverages the structured geometric knowledge learned by 3DFM while benefiting from the expressive generative power of diffusion models, resulting in more accurate and realistic novel-view depth synthesis.

A few works jointly predict novel views and depth \cite{Guizilini_2025_CVPR,10.1007/978-3-031-19824-3_15}. For example, \cite{Guizilini_2025_CVPR} learns scene representations within the diffusion model and predicts depth for a single target view under restrictive camera assumptions. In contrast, we use pretrained 3DFM features, decoupling representation from generation, and produce plausible depths for multiple novel viewpoints without assuming that the target camera(s) are always placed at the origin.

\parnobf{3D Foundation Models (3DFMs)}
Traditional 3D reconstruction methods such as Structure-from-Motion (SfM)\cite{7780814,10.1145/2001269.2001293,10.5555/861369} and Multi-View Stereo (MVS)\cite{Furukawa2007AccurateDA,Furukawa2015MultiViewSA,Galliani_2015_ICCV} recover camera poses and geometry through local feature matching and joint optimization via bundle adjustment. Recent 3D foundation models instead predict camera poses, depth, and point maps in a single feed-forward pass from multiple input images. Representative works in this line of research include DUSt3R \cite{dust3r_cvpr24} and its extensions \cite{mast3r_eccv24,Yang_2025_Fast3R,cut3r,Jang2025Pow3REU}, as well as multiview models such as VGGT \cite{wang2025vggt} and WorldMirror \cite{liu2025worldmirror}.
%, which learn scalable, multi-view geometric representations using unified transformer backbones to jointly predict 3D scene geometry (consisting of camera poses, depth, and point maps). 
These models are not designed for novel-view synthesis out of the box, but we repurpose their representations for this task by pairing them with diffusion models.  Specifically, we treat their internal features as strong scene-level geometry representation. We empirically show that a lightweight decoder can use these features to predict hidden surfaces (see \S~\ref{subsec:ldis}), demonstrating that 3DFMs encode information useful for reasoning about occluded geometry and unseen viewpoints.

\parnobf{Diffusion Models} Denoising Diffusion Probabilistic Models (DDPMs) \cite{10.5555/3495724.3496298} generate samples by iteratively reversing a diffusion process, which involves denoising Gaussian noise into structured outputs. Conditional DDPMs incorporate  information like text \cite{10.5555/3600270.3602913}, reference images \cite{10.1145/3528233.3530757}, or semantic maps \cite{Zhang2023AddingCC} to guide generation. Latent Diffusion Models (LDMs) \cite{10377858,rombach2021highresolution,10.5555/3692070.3692573} perform denoising in a compact latent space. Diffusion has recently been applied to monocular depth \cite{ke2023repurposing,Guizilini2024GRINZM,10.5555/3666122.3667835,saxena2023zeroshot,zhao2023unleashing,xu2025pixel} and camera pose \cite{10378090, zhang2024raydiffusion} prediction. These models operate in a compact latent space similar to original LDMs, allowing standard noise schedulers to work out of the box.
These approaches \cite{ke2023repurposing,Guizilini2024GRINZM,10.5555/3666122.3667835,saxena2023zeroshot,zhao2023unleashing,xu2025pixel} predict depth only for surfaces explicitly visible in the input images and cannot hallucinate occluded or unobserved geometry. In contrast, our method conditions diffusion on high dimensional geometric features extracted from pretrained 3DFMs that requires careful tuning of noise schedulers.

\section{Method}
Our goal is to predict the depth of a novel view given one or more input images as well as the pose of set of target camera views. 
In the base case of two views where one view is the source and the other view is target, the model should be able to plausibly hallucinate about the depth of the target view. With more views, the model should be able to predict the depth of the target view(s) more accurately. 
We propose Z3D, a simple and effective framework for novel view depth synthesis that integrates diffusion-based generative modeling with 3DFMs to leverage their complementary strengths. In what follows, we analyze how the scene representations learned by 3DFMs encode geometric structure beyond what is immediately apparent, using a lightweight probe to predict layered depth for each scene. Building upon our findings, we introduce Z3D method that leverages the encoding of a scene from 3DFMs within a diffusion framework to learn to generate consistent depth of novel views. 

\parnobf{Background of 3DFMs}
3DFMs~\cite{wang2025vggt,liu2025worldmirror} encode input images into patch tokens using a shared transformer backbone.

Each image is divided into patches, producing tokens that are processed independently in frame attention and jointly across images in global attention. The pretrained patch tokens from the transformer backbone capture rich multi-view geometric information and are fed into task-specific heads, such as depth, point map, or camera pose. In our method, we extract these pretrained patch tokens and use them as a scene representation to condition a diffusion model which allows us to predict the depth for novel viewpoints. These viewpoints reveal occluded and unobserved surfaces.

\begin{figure}[!t]
    \centering
    \includegraphics[page=1,width=0.90\linewidth]{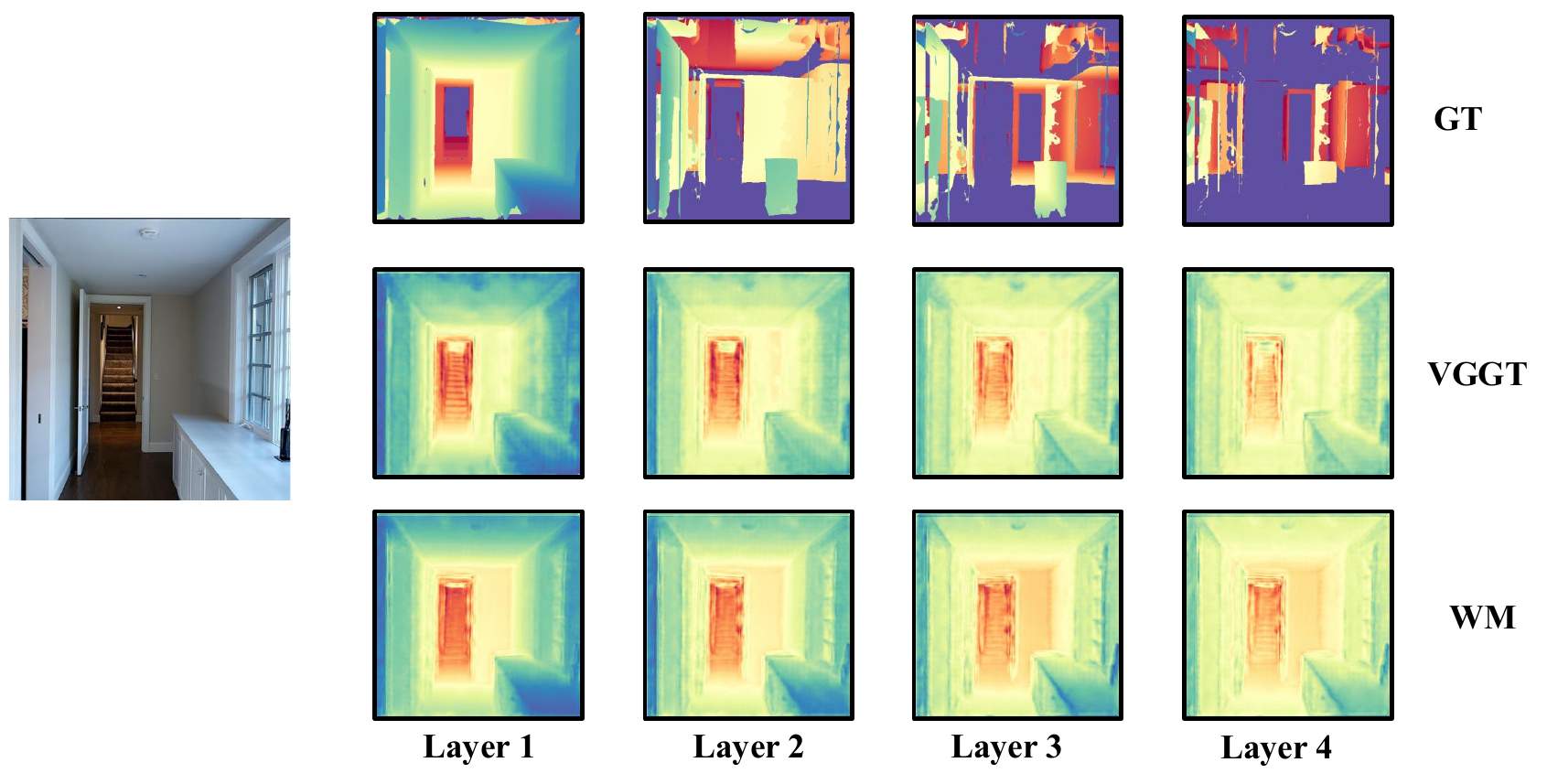}
    \caption{
    {\bf Linear Probe Layered Depth Image (LDI) prediction capability of 3DFMs.}
    To test whether 3DFMs implicitly contain information about unseen views, we train a linear model to predict LDIs. Good performance on LDI prediction after the first layer suggests that the model understands the full 3D scene, including hidden surfaces. For each example: (left) input RGB; (top) ground-truth LDI layers; (bottom) linearly decoded LDI layers. The ability to recover the non-trivial structure of the scene with only a linear layer suggests that 3DFMs implicitly encode this information.
    }
    \label{fig:ldi_results}
\end{figure}

\subsection{Do 3DFMs Know about Unseen Structure?}
\label{subsec:ldis}
%Before training the model to 3D from new views, w
We first test whether knowledge of this information is already embedded in the 3DFM representations. To accomplish this, we follow the work of Banani et al.~\cite{10656175} and probe the knowledge of these networks. As our target representation, we train a linear probe to predict layered depth images (LDI)~\cite{Shade1998LayeredDI}, a simple representation of occluded parts of the scene. LDIs generalize depth maps and represent, at each pixel, the first $k$ surfaces that the ray through the pixel passes through. Accurately estimating the 2nd surfaces and subsequent surfaces of an LDI requires knowledge of the occluded parts of the scene. 

\begin{wraptable}{l}{50mm}
\centering
\vspace{-10mm}
\caption{LDI results. }
\small
\begin{tabular}{lcc}
\toprule
Method & AbsRel $\downarrow$ & $\delta<1.25$ $\uparrow$ \\
\midrule
Avg LDI & 0.319 & 0.564 \\
\VGGTLDI & \underline{0.197} & \underline{0.717} \\
\WMTLDI & \textbf{0.167} & \textbf{0.789} \\
\bottomrule
\end{tabular}
\label{tab:ldi_qual_results}
\end{wraptable}
We test this by evaluating how well the last layer of a 3DFM's DPT head can predict LDIs. Specifically, we expand the DPT head of VGGT~\cite{wang2025vggt} to produce four depth layers and confidence maps and train the model while keeping all other layers fixed. Figure~\ref{fig:ldi_results} shows that 3DFM produces plausible layered depth predictions that extend beyond the first visible surface. These results suggest, that 3DFMs do contain implicit knowledge for reasoning about occluded and hidden scene structure. We also quantitatively confirm that linear probes on 3DFM features predict LDIs. We report results with VGGT and WM in Table \ref{tab:ldi_qual_results}. Despite being a weak linear model, they halve the error rate compared to using the
dataset average, showing the hidden/unseen information information is embedded in 3DFMs.

\subsection{Network Architecture}
\label{subsec:z3d_architecture}

\begin{figure}[!t]
    \centering
    \includegraphics[width=0.90\linewidth]{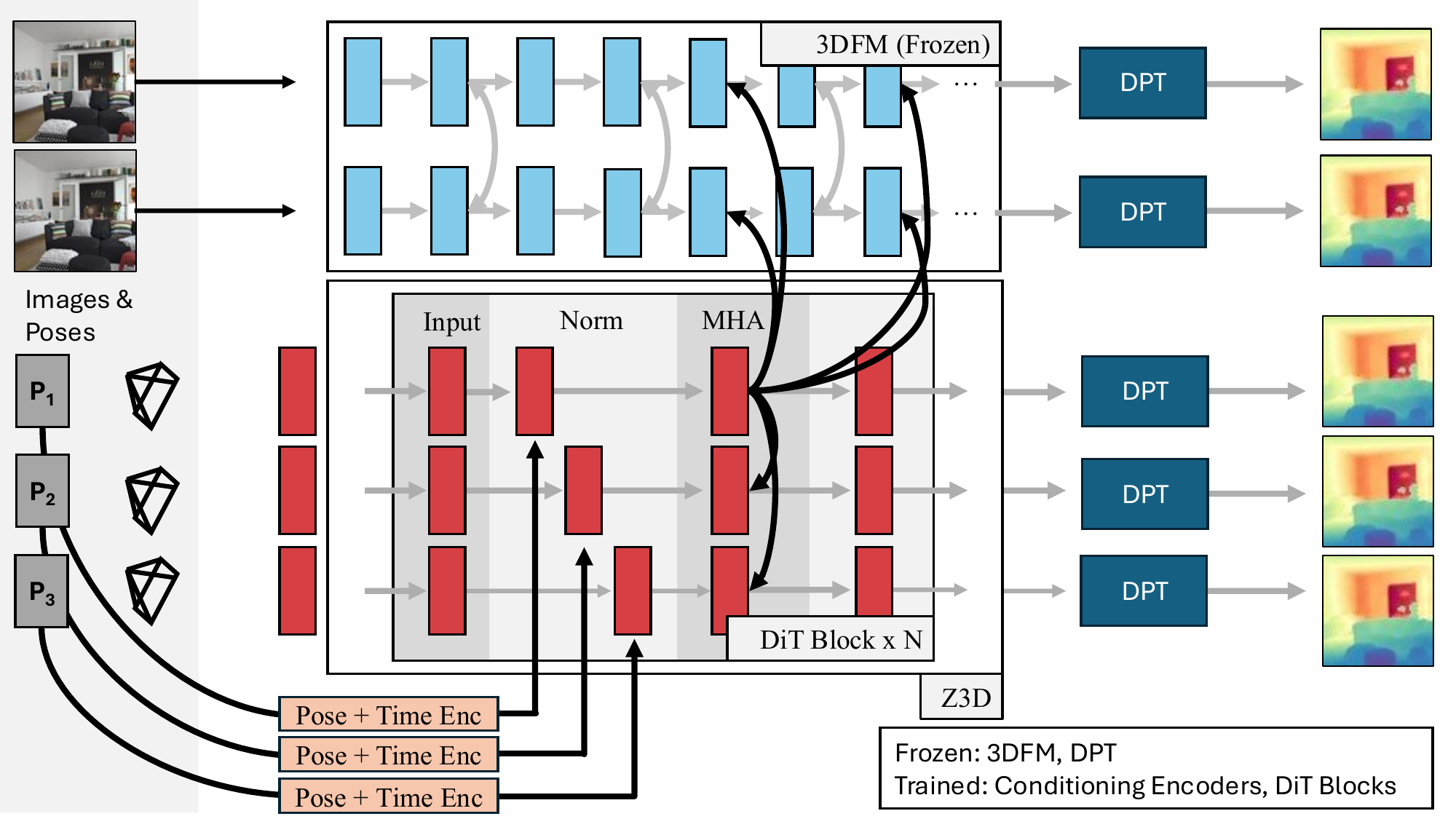}
    \caption{{\bf Z3D Inference Pipeline.} Z3D takes one or more input images (top left images in this figure) and multiple query poses ($p_{1}, p_{2}$, $p_{3}$ etc). The input images are passed into a 3DFM. Z3D infers the depthmaps at the query poses via latent diffusion using    3DFM features. In each DiT block, Z3D: (i) conditions on the encoded pose (plus diffusion timestep) via adaptive layer normalization following DiT practice; and (ii) attends to the 3DFM intermediate activations while doing multiheaded attention.  The diffusion-predicted 3DFM feature is passed directly into the frozen DPT head. }
    \label{fig:model-architecture}
\end{figure}

Our goal is to predict novel-view depth maps by performing conditional diffusion on 3DFM patch tokens, conditioned on relative poses and source view images. Z3D achieves this by combining a pretrained 3DFM with a Pose Encoder, a conditional Diffusion Model, and a DPT head. The pretrained 3DFM provides a high-dimensional latent representation of the scene, capturing rich multi-view geometry. The Pose Encoder encodes the relative camera poses of target views, while the diffusion model predicts the target view scene representation from noise conditioned on both the source views and target poses. Finally, the DPT head decodes the sampled scene representation into depth maps. Conceptually, this setup can be seen as latent diffusion operating in the space of 3DFM patch tokens, enabling hallucination of occluded surfaces and multi-view consistency.
During training, both source and target images are available. We first compute scene representations for the source and target views using the frozen 3DFM backbone. Gaussian noise is added to the target view representation, and the diffusion model learns to predict this noise conditioned on the source view representations and relative target view poses. Conditioning is implemented via a cross-attention block in each DiTBlock, while target pose embeddings are added directly to the timestep embeddings, allowing the model to reason about depth from relative camera positions.
\parnobf{3DFM Backbone, $\phi$} The 3DFM backbone tokenizes each source image into patches and produces patch tokens, which are processed through alternating frame-level and global attention blocks to capture both per-frame and multi-frame geometry. The resulting patch tokens are concatenated to form a scene representation for the input images. Only tokens from layer 17 are used for conditioning the diffusion model, leveraging the most informative geometric features while keeping computation efficient.
\parnobf{Pose Encoder, $p_{\theta}$} Relative camera poses of target views are encoded as 12-dimensional vectors per view. Each pose is Fourier-encoded and processed through an MLP to produce embeddings that match the feature dimension of the 3DFM patch tokens. All poses are represented relative to the first view in the source set, simplifying spatial reasoning for the diffusion model.
\parnobf{Diffusion Model, $f_{\theta}$} 
Our diffusion model follows
%is inspired by 
the DiT architecture~\cite{10377858} and operates directly in the high-dimensional latent space of 3DFM patch tokens. 
Let $I_s$ denote the source view image(s), $I_t$ the target view image(s), and $P_t$ the relative camera pose(s) of the target view(s). We use a shared image backbone $\phi$ to extract features from both target and source images:
$f_t = \phi(I_t)$ and $f_s = \phi(I_s)$.
We denote the target representation as $x_0 = f_t$ and the source representations as $z_{s} = f_s$. Also, we encode the relative pose using an MLP $p = p_\theta(P_t)$.

Recent 3DFM architectures such as VGGT~\cite{wang2025vggt} and WorldMirror~\cite{liu2025worldmirror} aggregate transformer tokens from multiple intermediate layers for depth prediction. 
Each of these token representations consists of high-dimensional embeddings (e.g., 2048-dimensional in both VGGT and WorldMirror), making it computationally prohibitive to apply diffusion directly to the concatenated multi-layer outputs used for depth estimation.

Following an analysis similar to~\cite{zhang2025emergentextremeviewgeometry3d}, we observe that, for depth prediction, tokens from layer 17 contribute significantly more than those from the other three aggregation layers. 
Based on this observation, we restrict diffusion to the scene tokens extracted from layer 17 for the target view(s), substantially reducing computational cost while preserving the most depth-relevant representation.
To mitigate potential information loss due to this constraint, we use the outputs from all four aggregation layers of the source views as conditioning signals. 
This allows the model to access richer multi-scale geometric information during cross-attention, while keeping the diffusion process tractable. We provide additional details of this analysis in the supplementary material.

During training, only the target-view tokens $x_0$ are noised, while the source-view tokens $z_s$ remain clean and serve solely as conditioning signals. 
The forward diffusion process corrupts the target tokens over timesteps $t \in \{1, \dots, T\}$:
%\begin{equation}
$x_t = \sqrt{\alpha_t}\, x_0 + \sqrt{1 - \alpha_t}\, \epsilon$,
%\end{equation}
where $\epsilon \sim \mathcal{N}(0, I)$ and
%\[
$\alpha_t = \prod_{s=1}^{t} (1 - \beta_s)$,
%\]
with $\{\beta_t\}_{t=1}^{T}$ defining the variance schedule.

The denoiser $f_\theta$ is trained to predict the velocity of injected noise:
%\begin{equation}
$\hat{v}_t = f_\theta(x_t,\, t,\, z_s,\, p)$,
%\end{equation}
where $x_t$ are the noised target tokens. 
Conditioning on $z_s$ is implemented through cross-attention layers inserted in each DiTBlock, allowing the model to extract geometry most relevant to the target view. The pose embedding $p$ is added to the timestep embedding.
%to enable viewpoint-aware denoising.

Because standard noise schedulers are typically designed for lower-dimensional latent spaces (e.g., dimension $\lesssim 1024$), directly applying them to the higher-dimensional 3DFM token space (typically $\geq 1024$) can lead to instability. 
Following~\cite{zheng2025diffusiontransformersrepresentationautoencoders}, we apply a timestep shift to the noise scheduler, inspired by~\cite{10.5555/3692070.3692573}, which stabilizes training and enables effective denoising in this richer latent space.

Figure \ref{fig:model-architecture} shows \method’s inference pipeline. 
We initialize $x_T \sim \mathcal{N}(0, I)$ and iteratively apply $f_\theta$ to recover an estimate of the clean target tokens $x_0$. 
The reconstructed tokens are then passed through the pretrained decoder $D$ to produce the final target-view depth map:
%\[
$\hat{d} = D(x_0)$.
%\]

\parnobf{DPT Head, $D$} The DPT head acts as a decoder, analogous to the decoder in a variational autoencoder (VAE)~\cite{10.5555/3295222.3295378}. It is the depth prediction heads of the 3DFMs. The sampled target-view scene representation tokens, $x_{0}$ produced by the diffusion model are passed through the DPT head, which decodes them into depth maps for the target views.

\subsection{Implementation Details}
We implement Z3D using PyTorch and the modified DiT model outlined in \S~\ref{subsec:z3d_architecture}). Our diffusion model is trained using a flow-matching noise scheduler with a flow-velocity prediction objective, following the formulation introduced in Stable Diffusion v3 \cite{10.5555/3692070.3692573}. During training , we apply \texttt{FlowMatchEulerDiscreteScheduler} \cite{10.5555/3692070.3692573} noise scheduler with 1000 timesteps. We adjust the timestep shift factor using a dimension-dependent scaling rule,
$\alpha = \sqrt{m/n}$, where the scaling compensates for changes in the latent dimensionality. The timestep shift factor in an Euler flow matching scheduler is a small offset applied to time steps to avoid evaluating the model exactly at the boundary times (0 and 1), improving numerical stability during sampling.
Following \cite{10.5555/3692070.3692573,zheng2025diffusiontransformersrepresentationautoencoders}, we use $n = 4096$ as the reference base dimension and set $m$ to the effective data dimension of the 3DFM backbone representation. At inference time, we apply the \texttt{FlowMatchEulerDiscreteScheduler} \cite{zheng2025diffusiontransformersrepresentationautoencoders} scheduler and only sample 50 steps.
We train our model in two stages. In the first stage, the diffusion model is trained under a one-source–one-target view setting using an effective batch size of 128 for 98k training steps. In the second stage, the model is initialized with the weights obtained from Stage~1 and further trained using a multi-view configuration consisting of two source views and four target views, with an effective batch size of 32 for 156k steps. 

Our model is not sensitive to batch size; however, we adopt this incremental training strategy because training the Stage~2 configuration from scratch was observed to be unstable and led to slower convergence. Initializing from Stage~1  led to much stable optimization and faster convergence. 
We train the model using the AdamW optimizer with $\beta_1 = 0.9$ and $\beta_2 = 0.95$, no weight decay, and a learning rate of $2 \times 10^{-4}$ for both training stages. We apply a linear warmup over the first 10\% of the total training steps, followed by a learning rate decay schedule that begins at 30\% of the training progress and gradually decreases the learning rate to zero by the end of training.

\subsection{Training Datasets}
\label{subsec:dataset-preparation}

Z3D is trained on sequences of overlapping camera views, typically containing 2--6 images per sequence. The supplement has more data preparation details, but briefly:
Training samples are constructed by mining 3D datasets to identify groups of views with sufficient geometric overlap (defined via  pairwise camera frustum overlap).

To ensure robustness and broad applicability, Z3D is trained on a large-scale aggregation of five diverse datasets that are both real and synthetic. This combined corpus provides extensive coverage across indoor and outdoor environments.

The specific datasets include MegaDepth \cite{Li_2018_CVPR}, Hypersim \cite{Roberts_2021_ICCV}, Taskonomy \cite{Zamir_2018_CVPR}, Replica \cite{replica19arxiv}, and Habitat HM3D \cite{9010745}.

\section{Experiments}
In this section, we present the experimental framework for \method. Our system is designed to predict novel-view depths from sparse input views using 3DFMs. Given the novelty of this problem, there are no existing methods that directly address this setting. To evaluate Z3D, we construct datasets as described in \S~\ref{subsec:dataset-preparation} and perform a comprehensive experimental analysis. We consider two recent 3DFMs: VGGT~\cite{wang2025vggt} and WorldMirror~\cite{liu2025worldmirror}.
\subsection{Baselines}
Several works attempt to predict novel-view depths jointly with novel-view synthesis~\cite{10.1007/978-3-031-19824-3_15, Guizilini_2025_CVPR}. However, to the best of our knowledge, there are no existing methods that repurpose 3DFMs specifically for this task. To address this, we construct strong baselines from methods that tackle parts of the problem. Our work is closely related to MVGD \cite{Guizilini_2025_CVPR}. Unfortunately, since the code for MVGD \cite{Guizilini_2025_CVPR} is not available, we were unable to perform a direct comparison.

\parnobf{LVSM + 3DFM}
We use LVSM \cite{jin2025lvsm} as a baseline because it is a state-of-the-art, scene-agnostic novel view synthesis model that generates novel RGB views from sparse observations and camera parameters (intrinsics and extrinsics). Since LVSM does not directly predict depth, we apply pre-trained 3DFMs (e.g., VGGT and WorldMirror) to the synthesized images to estimate depth for the posed cameras. This baseline evaluates the extent to which a general-purpose image synthesis model, combined with a strong monocular 3D foundation model, can be used for novel-view depth prediction. We denote these variants as \LVSMVGGT{} and \LVSMVWM{}, corresponding to LVSM followed by VGGT and WorldMirror depth estimation, respectively.

\parnobf{Depth Diffusion (DD)} As an additional baseline, we modify our approach by training the diffusion model directly in pixel space on the predicted depth maps from 3DFMs, rather than modeling the distribution of the target scene’s latent representation. To separate out the effects of the architecture from the space in which inference is done, we use the Z3D architecture but change the output space. We consider two variants of this baseline: Depth Diffusion (VGGT), hereafter referred to as \ZThreeDVGGTDD{}, and Depth Diffusion (WorldMirror), hereafter referred to as \ZThreeDWMDD.

\begin{figure}[!t]
    \centering
    \includegraphics[page=1,width=0.90\linewidth]{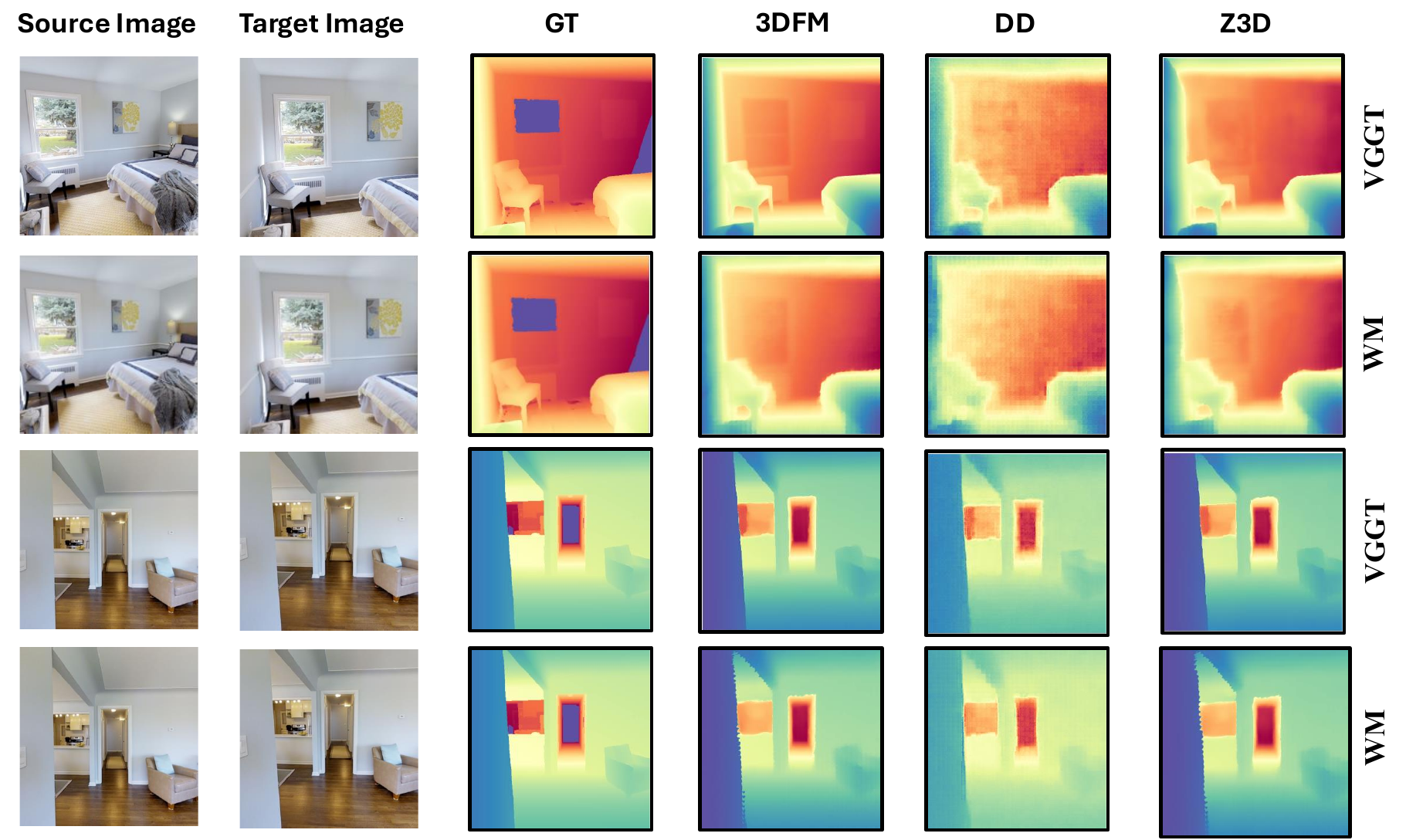}
   \caption{Qualitative comparison of Novel View Depth Predictions. Columns show: source RGB, target RGB, ground truth depth, 3DFM predicted depth, Depth Diffusion prediction, and Z3D prediction. Z3D results are substantially more smooth as compared to results from a method that works directly in patch-space.}
    \label{fig:qualitative_depth}
\end{figure}

\subsection{Evaluation}
In this section, we describe our quantitative and qualitative evaluations.

\parnobf{Datasets}
We follow a similar procedure as outlined in \S~\ref{subsec:dataset-preparation} to preprocess datasets for evaluation. We evaluate our model on both indoor and outdoor scenes, covering a diverse set of environments. Specifically, we perform quantitative evaluation on DTU~\cite{aanaes2016large}, NRGBD~\cite{Azinovic_2022_CVPR}, 7-Scenes~\cite{10.1109/CVPR.2013.377}, and an in-domain evaluation set sampled from the test splits of training datasets.

\parnobf{Depth Estimation Metrics} Following the affine-invariant depth evaluation protocol \cite{Ranftl2019TowardsRM}, we first align the predicted depth maps $\hat{d}$ to the ground truth $d$ using least squares fitting (Weiszfeld method~\cite{10.1007/s10957-014-0586-7}), producing the absolute aligned depth map $a = \hat{d} \times s + t$ in the same units as the ground truth. The depth quality is then assessed using two widely adopted metrics~\cite{Ranftl2019TowardsRM,9711226,9577432}. The first, Absolute Mean Relative Error (AbsRel), is computed as $\text{AbsRel} = \frac{1}{M} \sum_{i=1}^{M} |a_i - d_i| / d_i$, where $M$ is the total number of pixels. The second metric, $\delta < 1.25$, measures the fraction of pixels satisfying $\max(a_i / d_i, d_i / a_i) < 1.25$.

\parnobf{Multiview 3D Reconstruction Metrics} Since we predict novel depths for multiple cameras in a scene, we evaluate multiview 3D reconstruction by projecting the predicted depthmaps to point clouds. Consistent with prior works \cite{Azinovic_2022_CVPR,wang20243d,cut3r,dust3r_cvpr24}, we report Accuracy (Acc.), and Completion (Comp.), which together quantify both the fidelity and completeness of the reconstructed 3D geometry.

\parnobf{Qualitative Results}
Figure~\ref{fig:qualitative_depth} presents qualitative results of \method{}, and Depth Diffusion (DD) baselines. Both \ZThreeDVGGT{} and \ZThreeDWM{} predict reasonable novel-view depth when conditioned on the target camera pose. However, DD methods yield noticeably less smooth depth maps, likely because they operate directly in depth space rather than a latent representation.

\begin{figure}[!t]
    \centering
    \includegraphics[page=1,width=0.90\linewidth]{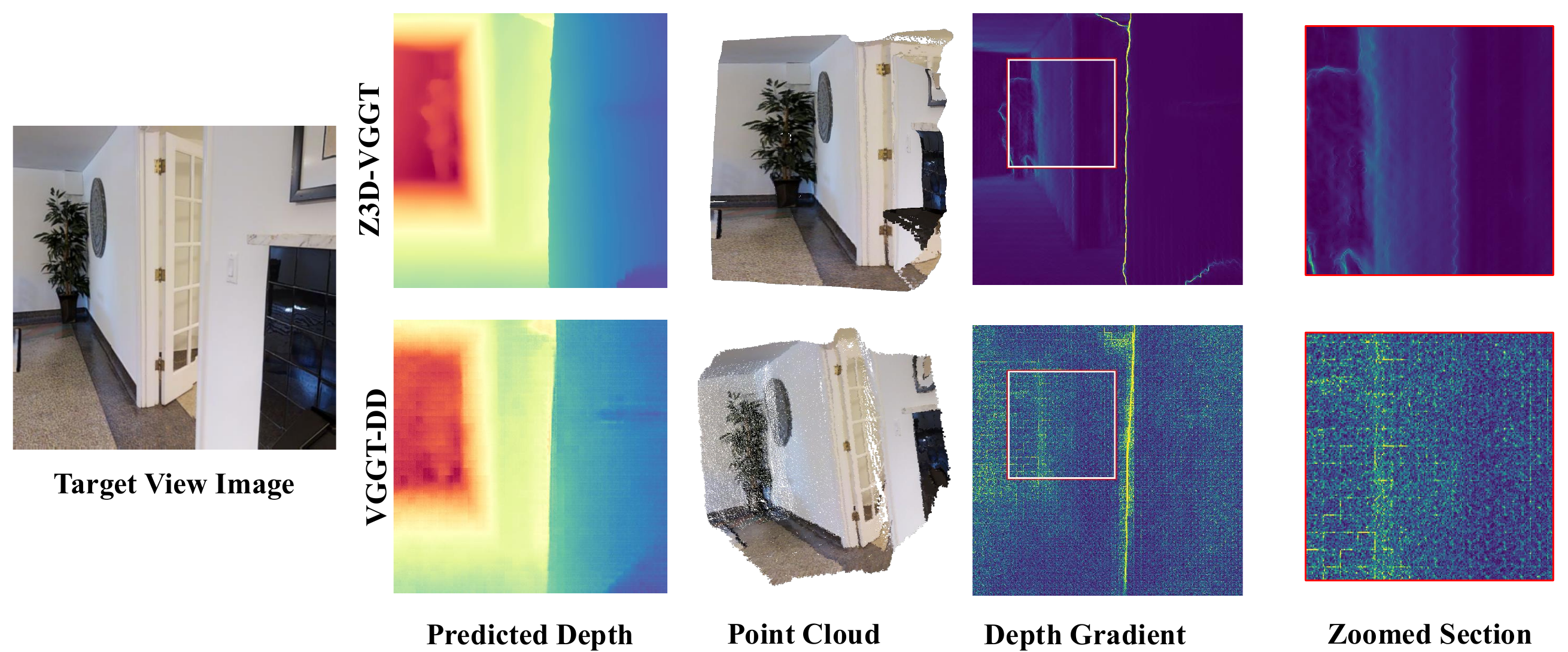}
    \caption{Depth gradient analysis of \ZThreeDVGGT{} and \ZThreeDVGGTDD{} model predictions}
    \label{fig:depth-gradient-results}
\end{figure}

To further substantiate this observation, we visualize the depth gradients of predictions from Z3D and DD baselines in Figure \ref{fig:depth-gradient-results}. In Figure \ref{fig:depth-gradient-results}, we observed pronounced gradient spikes in \ZThreeDVGGTDD depths in comparison to \ZThreeDVGGT{}, a trend that is consistently observed when comparing \ZThreeDWMDD{} and \ZThreeDWM{}. 

\begin{figure*}[!t]
\centering
\includegraphics[page=1, width=0.90\textwidth]{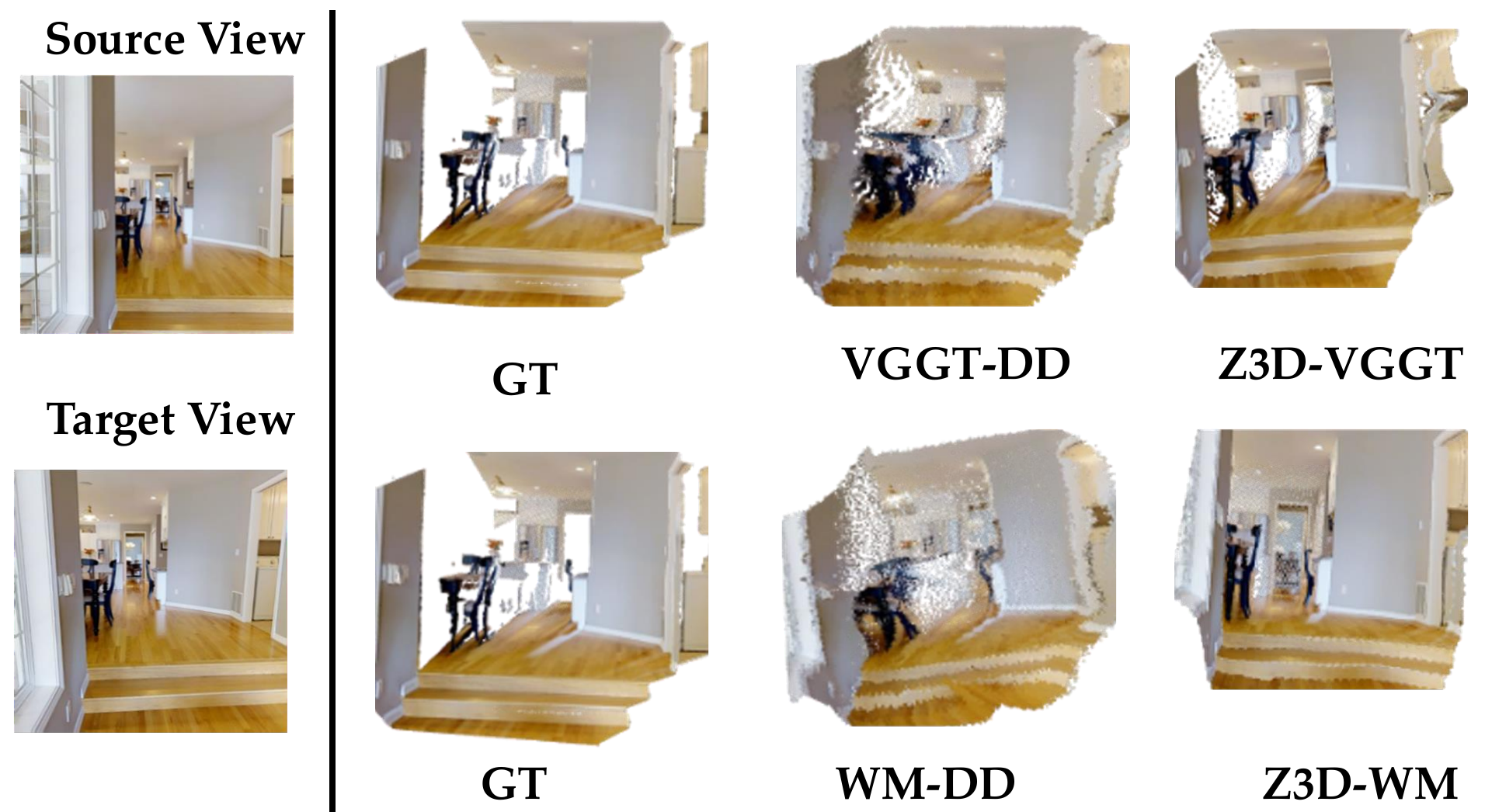}
\caption{Qualitative comparison of novel-view point cloud reconstructions from a single source image and target pose. Although \ZThreeDVGGTDD{} and \ZThreeDWMDD{}  achieves competitive quantitative performance, its reconstructed point clouds exhibit noticeably higher noise and reduced geometric sharpness compared to \ZThreeDVGGT{} and \ZThreeDWM{}, which produce cleaner and more coherent 3D structures. Top left: source view (left of the vertical line). Bottom left: target view (not observed by the model). First row: Predictions of VGGT , \ZThreeDVGGTDD{} and \ZThreeDVGGT{} respectively. Second row: Predictions of WM , \ZThreeDWMDD{} and \ZThreeDWM{} respectively.
}
\label{3d_pointmaps-example-2}
\end{figure*}

\begin{figure*}[!t]
\centering
\includegraphics[page=1, width=0.90\textwidth]{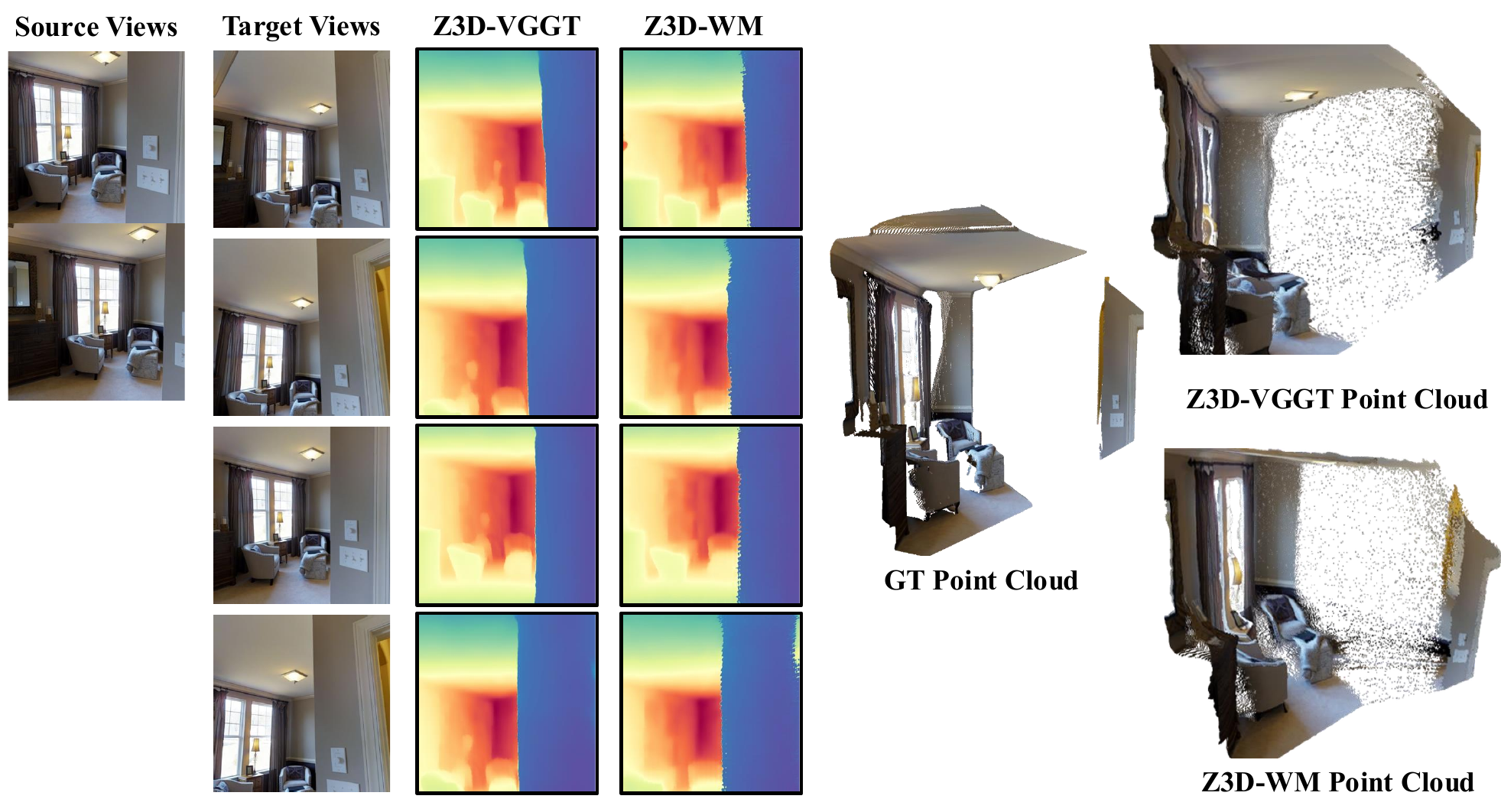}
\caption{Novel-view depth prediction from 2 source views $\rightarrow$ 4 target views. Conditioned on the target camera poses, \ZThreeDVGGT{} and \ZThreeDWM{} generate geometrically consistent depth maps across all target views, resulting in a coherent reconstructed point cloud.}
\label{multi-source_3d_pointmaps-example-2}
\end{figure*}

We also visualize point clouds obtained by projecting predicted depth maps for novel viewpoints in Figures~\ref{3d_pointmaps-example-2} and \ref{multi-source_3d_pointmaps-example-2}. In Figure~\ref{3d_pointmaps-example-2}, DD-based methods produce noticeably noisier point clouds compared to \method{} predictions. In the multi-view setting (Figure~\ref{multi-source_3d_pointmaps-example-2}), \method{} generates consistent depth across novel views given only a few source views (two in this case).

\subsubsection{Quantitative Evaluation}
% \parnobf{Quantitative Results}
We evaluate our method on out-of-domain datasets under two settings: 1-source $\to$ 1-target and 2-sources $\to$ 4-targets, to assess its generalization ability across unseen domains. Our approach is flexible with respect to the number of input views and performs well in both few-view and sparse-view settings.
Under 2-view setting, our models often outperform our baselines on depth map metrics and are comparable in point cloud reconstruction metrics. However, we qualitatively find that Z3D produces smoother results than the depth diffusion methods (see Figure \ref{fig:depth-gradient-results}).

\parnobf{In-domain Evaluation} In this setting, we evaluate Z3D and DD models on subsets of the test splits from the training datasets. As shown in Tables~\ref{tab:pointcloud_metrics-1-1}, \ref{tab:depth_metrics-1-1}, \ref{tab:two-4_targets_GT_point-eval}, and \ref{tab:two-4_targets_GT_depth-eval}, all models achieve strong performance on their respective training distributions.

\parnobf{Out-of-domain Evaluation} Here, we evaluate our method and baselines on the test splits of datasets not seen during the training of our model and baselines. Our LVSM+3DFM baselines provide a strong comparison since LVSM is trained end-to-end for novel-view synthesis and is expected to encode scene geometry accurately. However, we observe that LVSM sometimes struggles to generate novel views that are fully consistent with the target pose, suggesting that optimizing for perceptual quality can come at the expense of faithfully preserving the underlying scene geometry.

From Tables~\ref{tab:pointcloud_metrics-1-1} and \ref{tab:depth_metrics-1-1}, our Depth Diffusion baselines (\ZThreeDWMDD{} and \ZThreeDVGGTDD{}) are competitive but still struggle to produce accurate novel-view depths. This stems from the fact that they learn a direct mapping from observed depth to novel-view depth, where depth alone is not a sufficiently rich representation of full 3D scene geometry. In contrast, 3DFMs predict depth from a richer latent scene representation encoded in the backbone, with the depth head acting as a lightweight decoder. While this head captures some geometric cues, it is less expressive than the full latent features, which better support reasoning about scene structure from novel viewpoints.
\ZThreeDVGGT{} and \ZThreeDWM{} consistently outperform our baselines in out-of-domain evaluations in point clouds accuracy  and depth map metrics and perform on par with each other across all experiments. This indicates that the 3D representations learned by both 3DFMs capture similar geometric structures of the scene. We attribute the strong performance of \ZThreeDWM{} and \ZThreeDVGGT{} to the rich geometric scene representations learned by these models, as well as our implicit modeling of novel-depth synthesis through the diffusion framework. 

\parnobf{Outdoor Evaluation} Our training/evaluation data includes outdoor data from MegaDepth. This was included in the `In Domain' grouping; we explicitly pull out the MegaDepth results in Tab.~\ref{tab:megadepth}. Z3D substantially outperforms the DD models in depth metrics, and has substantially higher point cloud accuracy with slightly lower completeness than the DD. This also shows that Z3D method is domain agnostic and works for both indoor and outdoor scenes.

\begin{table}[!t]
\centering
\caption{Point Cloud Metrics (1 source $\to$ 1 target). Best results per dataset in \textbf{bold}, second best \underline{underlined}. \method{} performs well, especially in accuracy, and often has comparable completeness. Blue color  indicates out of domain evaluation. Note: M means mean and Md means median \scriptsize}
\label{tab:pointcloud_metrics-1-1}
\setlength{\tabcolsep}{1.8pt}
\resizebox{\columnwidth}{!}{%
\begin{tabular}{
l
cccc
*{12}{>{\columncolor{oodblue}}c}
}
\toprule

& \multicolumn{4}{c}{~} 
& \multicolumn{12}{c}{\cellcolor{oodblue}\bf Out of Domain}
\\

& \multicolumn{4}{c}{\textbf{In-Domain}}
& \multicolumn{4}{c}{\cellcolor{oodblue}\textbf{DTU}}
& \multicolumn{4}{c}{\cellcolor{oodblue}\textbf{7-Scenes}}
& \multicolumn{4}{c}{\cellcolor{oodblue}\textbf{NRGBD}} \\

\cmidrule(lr){2-5}
\cmidrule(lr){6-9}
\cmidrule(lr){10-13}
\cmidrule(lr){14-17}

\textbf{Model}
& \multicolumn{2}{c}{Acc $\downarrow$}
& \multicolumn{2}{c}{Comp $\downarrow$}
& \multicolumn{2}{c}{\cellcolor{oodblue}Acc $\downarrow$}
& \multicolumn{2}{c}{\cellcolor{oodblue}Comp $\downarrow$}
& \multicolumn{2}{c}{\cellcolor{oodblue}Acc $\downarrow$}
& \multicolumn{2}{c}{\cellcolor{oodblue}Comp $\downarrow$}
& \multicolumn{2}{c}{\cellcolor{oodblue}Acc $\downarrow$}
& \multicolumn{2}{c}{\cellcolor{oodblue}Comp $\downarrow$} \\

& M & Md & M & Md
& M & Md & M & Md
& M & Md & M & Md
& M & Md & M & Md \\

\midrule

\LVSMVGGT
& - & - & - & -
& 7.681 & 6.503 & 23.396 & 11.958
& 0.065 & 0.042 & 0.184 & 0.091
& 0.168 & 0.097 & 0.369 & \underline{0.204} \\

\LVSMVWM
& - & - & - & -
& 7.590 & 6.304 & 23.242 & 11.512
& 0.063 & 0.038 & 0.178 & 0.083
& 0.168 & 0.095 & 0.385 & \textbf{0.217} \\

\ZThreeDVGGTDD
& 0.154 & 0.044 & 0.149 & \textbf{0.013}
& 6.337 & 4.386 & \textbf{1.644} & \textbf{1.155}
& 0.046 & 0.029 & \underline{0.029} & \underline{0.010}
& 0.255 & 0.092 & 0.192 & 0.743 \\

\ZThreeDWMDD{}
& 0.130 & 0.053 & \textbf{0.068} & \textbf{0.013}
& 11.282 & 7.379 & \underline{1.737} & \underline{1.335}
& 0.042 & 0.029 & \textbf{0.024} & \textbf{0.009}
& \underline{0.159} & 0.084 & \underline{0.127} & 0.837 \\

\ZThreeDVGGT
& \underline{0.121} & \textbf{0.020} & 0.181 & \underline{0.021}
& \textbf{2.725} & \textbf{1.782} & 2.938 & 2.225
& \underline{0.033} & \underline{0.018} & 0.046 & 0.022
& 0.340 & \underline{0.056} & 0.260 & 0.676 \\

\ZThreeDWM
& \textbf{0.083} & \underline{0.022} & \underline{0.100} & \underline{0.021}
& \underline{5.021} & \underline{2.408} & 3.827 & 2.610
& \textbf{0.029} & \textbf{0.017} & 0.038 & 0.019
& \textbf{0.154} & \textbf{0.050} & \textbf{0.133} & 0.826 \\

\bottomrule
\end{tabular}%
}

\end{table}

\begin{table}[!t]
\centering
\caption{Depth Estimation Metrics (1 source $\to$ 1 target). Best results per dataset in \textbf{bold}, second best \underline{underlined}. Blue color  indicates out of domain evaluation.}
\label{tab:depth_metrics-1-1}

\resizebox{\textwidth}{!}{%
\begin{tabular}{
l
cc
>{\columncolor{oodblue}}c
>{\columncolor{oodblue}}c
>{\columncolor{oodblue}}c
>{\columncolor{oodblue}}c
>{\columncolor{oodblue}}c
>{\columncolor{oodblue}}c
}
\toprule

& \multicolumn{2}{c}{~}
& \multicolumn{6}{>{\columncolor{oodblue}}c}{\textbf{Out-of-Domain}} \\

\cmidrule(lr){2-3}
\cmidrule(lr){4-9}

& \multicolumn{2}{c}{\textbf{In-Domain}}
& \multicolumn{2}{c}{\cellcolor{oodblue}\textbf{DTU}}
& \multicolumn{2}{c}{\cellcolor{oodblue}\textbf{7-Scenes}}
& \multicolumn{2}{c}{\cellcolor{oodblue}\textbf{NRGBD}} \\

\cmidrule(lr){2-3}
\cmidrule(lr){4-5}
\cmidrule(lr){6-7}
\cmidrule(lr){8-9}

Model
& AbsRel$\downarrow$ & $\delta < 1.25$ $\uparrow$
& AbsRel$\downarrow$ & $\delta < 1.25$ $\uparrow$
& AbsRel$\downarrow$ & $\delta < 1.25$ $\uparrow$
& AbsRel$\downarrow$ & $\delta < 1.25$ $\uparrow$ \\

\midrule

\LVSMVGGT
& -- & --
& 0.468 & 0.024
& 0.342 & 0.745
& 0.214 & 0.666 \\

\LVSMVWM
& -- & --
& 0.468 & 0.025
& 0.328 & 0.749
& 0.210 & 0.672 \\

\ZThreeDVGGTDD
& 0.098 & 0.914
& 0.022 & \underline{0.998}
& 0.274 & \underline{0.963}
& 0.192 & 0.743 \\

\ZThreeDWMDD
& 0.108 & 0.902
& 0.036 & 0.997
& 0.272 & \textbf{0.967}
& \underline{0.127} & \textbf{0.837} \\

\ZThreeDVGGT
& \underline{0.076} & \underline{0.935}
& \textbf{0.012} & \textbf{0.999}
& \textbf{0.260} & \underline{0.963}
& 0.260 & 0.676 \\

\ZThreeDWM
& \textbf{0.075} & \textbf{0.939}
& \underline{0.021} & \textbf{0.999}
& \underline{0.266} & \textbf{0.967}
& \textbf{0.133} & \underline{0.826} \\

\bottomrule
\end{tabular}%
}
\end{table}

\begin{table}[h]
\centering
\caption{Point Cloud Metrics (2 source $\to$ 4 target). \textbf{Bold}: best (lowest) per column per dataset, \underline{underline}: second best.}\scriptsize
\setlength{\tabcolsep}{1.8pt}
\resizebox{\columnwidth}{!}{%
\begin{tabular}{
l
cccc
*{12}{>{\columncolor{oodblue}}c}
}
\toprule

& \multicolumn{4}{c}{~}
& \multicolumn{12}{c}{\cellcolor{oodblue}\bf Out of Domain}
\\

& \multicolumn{4}{c}{\textbf{In-Domain}}
& \multicolumn{4}{c}{\cellcolor{oodblue}\textbf{DTU}}
& \multicolumn{4}{c}{\cellcolor{oodblue}\textbf{7-Scenes}}
& \multicolumn{4}{c}{\cellcolor{oodblue}\textbf{NRGBD}} \\

\cmidrule(lr){2-5}
\cmidrule(lr){6-9}
\cmidrule(lr){10-13}
\cmidrule(lr){14-17}

Model
& \multicolumn{2}{c}{Acc $\downarrow$}
& \multicolumn{2}{c}{Cmp $\downarrow$}
& \multicolumn{2}{c}{\cellcolor{oodblue}Acc $\downarrow$}
& \multicolumn{2}{c}{\cellcolor{oodblue}Cmp $\downarrow$}
& \multicolumn{2}{c}{\cellcolor{oodblue}Acc $\downarrow$}
& \multicolumn{2}{c}{\cellcolor{oodblue}Cmp $\downarrow$}
& \multicolumn{2}{c}{\cellcolor{oodblue}Acc $\downarrow$}
& \multicolumn{2}{c}{\cellcolor{oodblue}Cmp $\downarrow$} \\

& M & Md & M & Md
& M & Md & M & Md
& M & Md & M & Md
& M & Md & M & Md \\

\midrule

\LVSMVGGT{}
& - & - & - & -
& 9.32 & 7.90 & 16.45 & 7.31
& \underline{0.025} & \underline{0.014} & 0.055 & 0.017
& \underline{0.062} & \underline{0.028} & 0.140 & 0.033 \\

\LVSMVWM{}
& - & - & - & -
& \underline{9.20} & 7.570 & 16.510 & 6.810
& \textbf{0.022} & \textbf{0.013} & 0.053 & 0.017
& \textbf{0.055} & \underline{0.028} & 0.138 & 0.039 \\

\ZThreeDVGGTDD{}
& 0.266 & 0.148  & 0.107 &  0.018
& 17.03 & 12.22 & \textbf{1.60} & \textbf{1.05}
& 0.046 & 0.029 & \underline{0.023} & \textbf{0.006}
& 0.215 & 0.080 & \underline{0.091} & \underline{0.018} \\

\ZThreeDWMDD{}{}
& 0.237 &  0.123 &  0.079  & 0.015
& 19.20 & 12.34 & \underline{1.690} & \textbf{1.05}
& 0.054 & 0.037 & \underline{0.023} & 0.007
& 0.113 & 0.069 & \textbf{0.074} & \textbf{0.014} \\

\ZThreeDVGGT{}
& \underline{0.131} & \textbf{0.037} & 0.112 & \textbf{0.018}
& \textbf{6.14} & \textbf{3.71} & 2.79 & \underline{1.96}
& 0.027 & \underline{0.014} & 0.032 & 0.012
& 0.211 & 0.045 & 0.109 & 0.037 \\

\ZThreeDWM{}
& \textbf{0.120} & \underline{0.041} & \textbf{0.088} & \underline{0.019}
& 10.32 & \underline{5.67} & 3.23 & 2.18
& 0.026 & \underline{0.014} & \textbf{0.029} & \underline{0.010}
& 0.091 & \textbf{0.035} & 0.108 & 0.031 \\

\bottomrule
\label{tab:two-4_targets_GT_point-eval}

\end{tabular}%
}
\end{table}

% -------------------- Depth --------------------
\begin{table}[h]
\centering
\caption{Depth Estimation Metrics (2 sources $\to$ 4 targets). \textbf{Bold}: best per column per dataset, \underline{underline}: second best.}
\scriptsize
\setlength{\tabcolsep}{3pt}
\resizebox{\columnwidth}{!}{%
\begin{tabular}{
l
cc
>{\columncolor{oodblue}}c
>{\columncolor{oodblue}}c
>{\columncolor{oodblue}}c
>{\columncolor{oodblue}}c
>{\columncolor{oodblue}}c
>{\columncolor{oodblue}}c
}
\toprule

& \multicolumn{2}{c}{~}
& \multicolumn{6}{>{\columncolor{oodblue}}c}{\textbf{Out-of-Domain}} \\

& \multicolumn{2}{c}{\textbf{In-Domain}}
& \multicolumn{2}{c}{\cellcolor{oodblue}\textbf{DTU}}
& \multicolumn{2}{c}{\cellcolor{oodblue}\textbf{7Scenes}}
& \multicolumn{2}{c}{\cellcolor{oodblue}\textbf{NRGBD}} \\

\cmidrule(lr){2-3}
\cmidrule(lr){4-5}
\cmidrule(lr){6-7}
\cmidrule(lr){8-9}

Model
& AbsRel$\downarrow$ & $\delta < 1.25$ $\uparrow$
& AbsRel$\downarrow$ & $\delta < 1.25$ $\uparrow$
& AbsRel$\downarrow$ & $\delta < 1.25$ $\uparrow$
& AbsRel$\downarrow$ & $\delta < 1.25$ $\uparrow$ \\
\midrule

\LVSMVGGT{}
& - & -
& 0.476 & 0.006
& \textbf{0.343} & 0.933
& \underline{0.127} & 0.828 \\

\LVSMVWM{}
& - & -
& 0.475 & 0.006
& \underline{0.346} & 0.935
& \textbf{0.121} & \textbf{0.842} \\

\ZThreeDVGGTDD{}
& 0.248 & 0.667
& 0.055 & 0.986
& 0.372 & 0.959
& 0.225 & 0.686 \\

\ZThreeDWMDD
& 0.215 & 0.735
& 0.059 & 0.979
& 0.383 & 0.943
& 0.147 & 0.812 \\

\ZThreeDVGGT{}
& \textbf{0.112} & \textbf{0.891}
& \textbf{0.024} & \textbf{0.995}
& \textbf{0.342} & \textbf{0.965}
& 0.223 & 0.715 \\

\ZThreeDWM{}
& \underline{0.118} & \underline{0.889}
& \underline{0.036} & \underline{0.987}
& 0.362 & \underline{0.962}
& 0.131 & \underline{0.830} \\

\bottomrule
\end{tabular}}
\label{tab:two-4_targets_GT_depth-eval}
\end{table}

\begin{table}[!t]
\centering
\caption{MegaDepth depth, point cloud metrics. 2 source $\rightarrow$ 4 targets. \textbf{Bold}: best, \underline{underline}: second best. M: Mean, Md: Median }
\label{tab:megadepth}
% \vspace{-3mm}
\scriptsize
\setlength{\tabcolsep}{2.5pt}
\begin{tabular}{lcccccc}
\toprule
&  $\downarrow$ &  $\uparrow$
& \multicolumn{2}{c}{Acc $\downarrow$} & \multicolumn{2}{c}{Comp $\downarrow$} \\
Model &Abs Rel  & $\delta{<}1.25$ & Mean & Med & Mean & Med \\
\midrule
\ZThreeDVGGTDD       & 
0.102 & 0.905 & 
0.203 & 0.058 & \underline{0.176} & \underline{0.014} \\
\ZThreeDWMDD       & 
0.111 & 0.896 &
0.241 & 0.060 & \textbf{0.163} & \textbf{0.013} \\

\ZThreeDVGGT{} & 
\textbf{0.039} & \underline{0.976} &
\textbf{0.091} & \textbf{0.020} & 0.213 & 0.023 \\
\ZThreeDWM{}   & 
\underline{0.041} & \textbf{0.977} & 
\underline{0.110} & \underline{0.025} & 0.209 & 0.026 \\
\bottomrule
\end{tabular}
\label{tab:two-4_targets_megadepthGT_point-eval}

\end{table}

\parnobf{Generalization to Newer 3DFMs}
We further evaluate Z3D on a recently introduced 3D foundation model, VGGT-$\Omega$\cite{wang2026vggtomega}, demonstrating that the proposed framework is largely model-agnostic. We train \VZThreeVGGTOmega{} under the one-source-view $\rightarrow$ one-target-view setting without any algorithmic modifications. Tables~\ref{tab:pointcloud_metrics-omega} and \ref{tab:depth_metrics-omega}  report the corresponding point cloud and depth estimation results. \VZThreeVGGTOmega{} achieves performance comparable to \ZThreeDVGGT{} and \ZThreeDWM{}, showing that Z3D can be readily adapted to new 3D foundation models while maintaining competitive performance.

\begin{table}[!t]
\centering
\caption{Point Cloud Metrics (1 source $\to$ 1 target). \textbf{Bold}: best (lowest) per column per dataset, \underline{underline}: second best.}\scriptsize
\label{tab:pointcloud_metrics-omega}
\setlength{\tabcolsep}{1.8pt}
\resizebox{\columnwidth}{!}{%
\begin{tabular}{
l
*{4}{c}
*{12}{>{\columncolor{oodblue}}c}
}
\toprule

& \multicolumn{4}{c}{~}
& \multicolumn{12}{c}{\cellcolor{oodblue}\bf Out of Domain} \\

& \multicolumn{4}{c}{\textbf{In-Domain}}
& \multicolumn{4}{c}{\cellcolor{oodblue}\textbf{DTU}}
& \multicolumn{4}{c}{\cellcolor{oodblue}\textbf{7-Scenes}}
& \multicolumn{4}{c}{\cellcolor{oodblue}\textbf{NRGBD}} \\

\cmidrule(lr){2-5}
\cmidrule(lr){6-9}
\cmidrule(lr){10-13}
\cmidrule(lr){14-17}

\textbf{Model}
& \multicolumn{2}{c}{Acc $\downarrow$}
& \multicolumn{2}{c}{Comp $\downarrow$}
& \multicolumn{2}{c}{\cellcolor{oodblue}Acc $\downarrow$}
& \multicolumn{2}{c}{\cellcolor{oodblue}Comp $\downarrow$}
& \multicolumn{2}{c}{\cellcolor{oodblue}Acc $\downarrow$}
& \multicolumn{2}{c}{\cellcolor{oodblue}Comp $\downarrow$}
& \multicolumn{2}{c}{\cellcolor{oodblue}Acc $\downarrow$}
& \multicolumn{2}{c}{\cellcolor{oodblue}Comp $\downarrow$} \\

& M & Md & M & Md
& M & Md & M & Md
& M & Md & M & Md
& M & Md & M & Md \\

\midrule

\ZThreeDVGGTDD
& 0.154 & 0.044 & 0.149 & \textbf{0.013}
& 6.337 & 4.386 & \textbf{1.644} & \textbf{1.155}
& 0.046 & 0.029 & 0.029 & \underline{0.010}
& 0.255 & 0.092 & \underline{0.097} & 0.025 \\

\ZThreeDWMDD
& 0.130 & 0.053 & \textbf{0.068} & \textbf{0.013}
& 11.282 & 7.379 & \underline{1.737} & \underline{1.335}
& 0.042 & 0.029 & \textbf{0.024} & \textbf{0.009}
& 0.159 & 0.084 & \textbf{0.084} & 0.019 \\

\ZThreeDVGGT
& 0.121 & \textbf{0.020} & 0.181 & \underline{0.021}
& \textbf{2.725} & \textbf{1.782} & 2.938 & 2.225
& 0.033 & \underline{0.018} & 0.046 & 0.022
& 0.340 & 0.056 & 0.156 & 0.053 \\

\ZThreeDWM
& \textbf{0.083} & \underline{0.022} & \underline{0.100} & \underline{0.021}
& 5.021 & \underline{2.408} & 3.827 & 2.610
& \textbf{0.029} & \textbf{0.017} & 0.038 & 0.019
& \underline{0.154} & \underline{0.050} & 0.139 & \underline{0.050} \\

\VZThreeVGGTOmega
& \underline{0.108} & 0.029 & 0.114 & 0.029
& \underline{4.834} & 3.117 & 4.279 & 3.386
& \underline{0.030} & \textbf{0.017} & \underline{0.032} & 0.019
& \textbf{0.119} & \textbf{0.039} & 0.127 & \textbf{0.034} \\

\bottomrule
\end{tabular}%
}
\end{table}

\begin{table}[!t]
\centering
\caption{Depth Estimation Metrics (1 sources $\to$ 1 target). \textbf{Bold}: best per column per dataset, \underline{underline}: second best.}
\label{tab:depth_metrics-omega}
\scriptsize
\resizebox{\columnwidth}{!}{%
\begin{tabular}{
l
cc
>{\columncolor{oodblue}}c
>{\columncolor{oodblue}}c
>{\columncolor{oodblue}}c
>{\columncolor{oodblue}}c
>{\columncolor{oodblue}}c
>{\columncolor{oodblue}}c
}
\toprule

& \multicolumn{2}{c}{~}
& \multicolumn{6}{>{\columncolor{oodblue}}c}{\textbf{Out-of-Domain}} \\

& \multicolumn{2}{c}{\textbf{In-Domain}}
& \multicolumn{2}{c}{\cellcolor{oodblue}\textbf{DTU}}
& \multicolumn{2}{c}{\cellcolor{oodblue}\textbf{7-Scenes}}
& \multicolumn{2}{c}{\cellcolor{oodblue}\textbf{NRGBD}} \\

\cmidrule(lr){2-3}
\cmidrule(lr){4-5}
\cmidrule(lr){6-7}
\cmidrule(lr){8-9}

\textbf{Model}
& AbsRel$\downarrow$ & $\delta<1.25$ $\uparrow$
& AbsRel$\downarrow$ & $\delta<1.25$ $\uparrow$
& AbsRel$\downarrow$ & $\delta<1.25$ $\uparrow$
& AbsRel$\downarrow$ & $\delta<1.25$ $\uparrow$ \\

\midrule

\ZThreeDVGGTDD
& 0.098 & 0.914
& 0.022 & \underline{0.998}
& 0.274 & 0.963
& 0.192 & \underline{0.743} \\

\ZThreeDWMDD
& 0.108 & 0.902
& 0.036 & 0.997
& \underline{0.272} & \underline{0.967}
& \underline{0.127} & \textbf{0.837} \\

\ZThreeDVGGT
& \underline{0.076} & \underline{0.935}
& \textbf{0.012} & \textbf{0.999}
& \textbf{0.260} & 0.963
& 0.260 & 0.676 \\

\ZThreeDWM
& \textbf{0.075} & \textbf{0.939}
& 0.021 & \textbf{0.999}
& 0.266 & \underline{0.967}
& 0.133 & 0.826 \\

\VZThreeVGGTOmega
& 0.096 & 0.925
& \underline{0.019} & \textbf{0.999}
& 0.305 & \textbf{0.978}
& \textbf{0.126} & \textbf{0.837} \\

\bottomrule
\end{tabular}%
}
\end{table}
\parnobf{Limitations} 
Z3D advances novel-view depth synthesis by leveraging 3DFM scene representations, but also has certain limitations. We observe that performance degrades when there is limited overlap between source and target views, making it difficult to recover fine-grained geometry in novel viewpoints. In our setup, the first source image is treated as the reference camera. Prior work on 3DFMs~\cite{wang2025pi} has shown that the choice of reference view can affect model performance. Z3D inherits this sensitivity and reliable 3D reconstruction becomes challenging when input views provide insufficient scene coverage. Thus, with low-overlap conditions between source and target views, both \ZThreeDVGGT{} and \ZThreeDWM{} struggle to produce accurate depth predictions for novel views. Figure \ref{fig:2-4_compare-4-z3d-limitations} illustrates a typical failure mode of Z3D, where the method struggles to predict novel-view depth with fine-grained details. In such cases, the resulting point clouds become inconsistent.

% --- Page 1 ---
\begin{figure}[!t]
    \centering
  \includegraphics[page=1,width=0.90\textwidth]{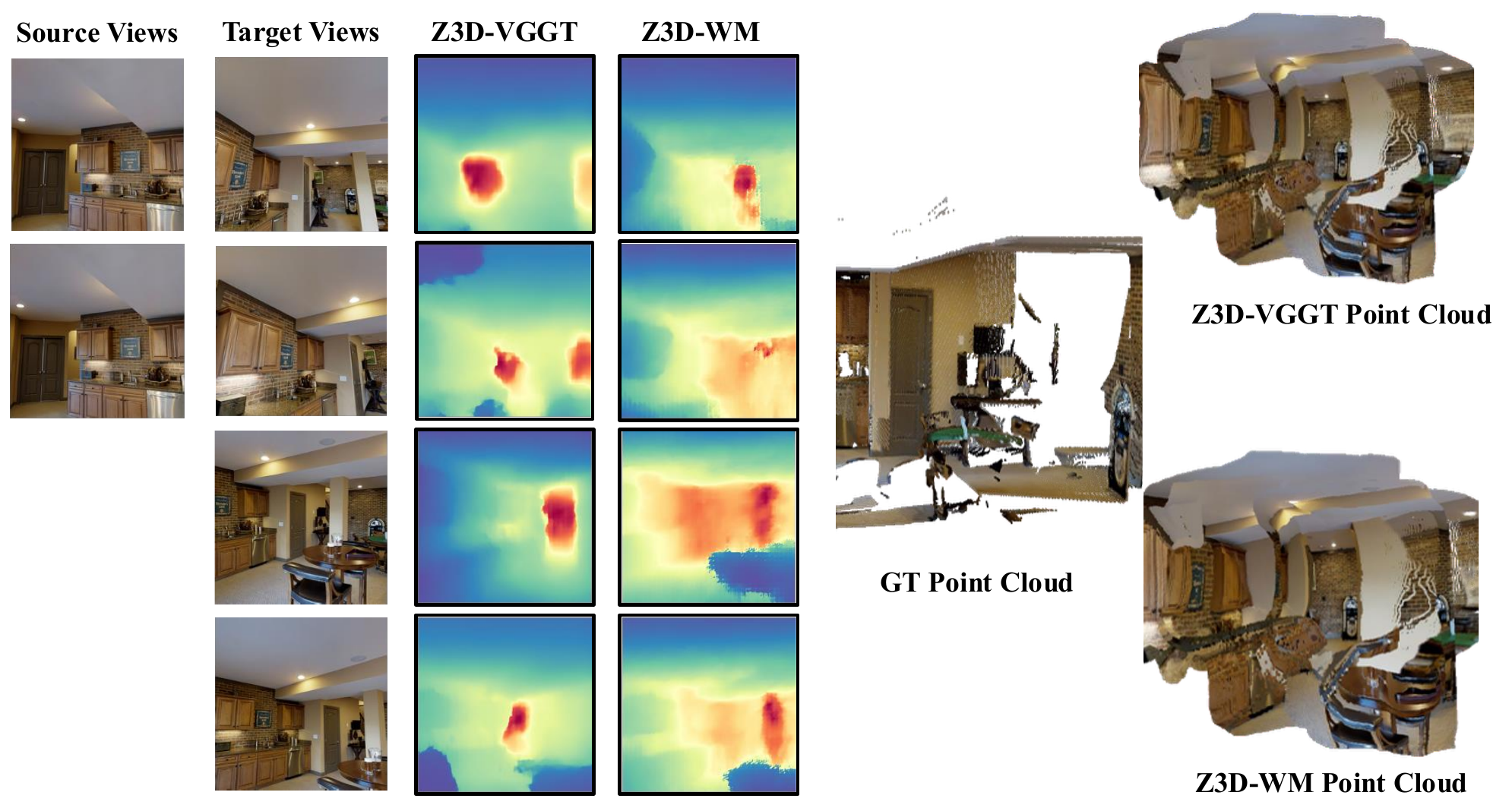}
    \caption{Point maps obtained from the depth maps predicted by \ZThreeDVGGT{} and \ZThreeDWM{}. Both \ZThreeDVGGT{} and \ZThreeDWM{} struggles to predict depths when source views have limited overlap with target views. The depth maps from \ZThreeDVGGT{} and \ZThreeDWM{} preserve the overall scene structure, but finer details are missing, and the resulting projected point clouds is not consistent.}
    \label{fig:2-4_compare-4-z3d-limitations}
    
\end{figure}
%}
\section{Conclusion}
We present \method, novel depth synthesis model that requires at atleast one view and a new view pose to predict the novel depth of that new view. Our method can take an arbitrary number of posed images and predict the depth of novel views conditioned on the camera pose and latent scene representation of the scene extracted from 3DFMs.

% \parnobf{Acknowledgments}
\section*{Acknowledgements}
This work was partially supported by NSF \#2437330
and NYU IT High Performance Computing resources, services,
and staff expertise. Thanks to Sarah Jabbour, Joseph Tung, Ruoyu Wang, and V. Samuel Pérez-Díaz for helpful feedback.

\clearpage


\bibliographystyle{splncs04}
\bibliography{main}

\clearpage

\appendix
\renewcommand{\thesection}{\arabic{section}}
\renewcommand{\baselinestretch}{1.02}

\definecolor{oodblue}{RGB}{240,248,255}

\title{Zero-Shot Novel Depth Synthesis Using 3D Foundation Models Scene Representations -Supplementary Material} 

\titlerunning{Zero-Shot Novel Depth Synthesis Using 3DFMs}

\author{Denis Mbey Akola\inst{1} \and
David F. Fouhey\inst{1, 2}}

\authorrunning{D. Akola and D. Fouhey}

\institute{New York University, Tandon School of Engineering,  1 MetroTech Center, Brooklyn, NY 11201, USA  \and
Courant Institute of Mathematical Sciences,  251 Mercer St, New York, NY 10012 }
\maketitle

\section{Introduction}

The supplementary material provides: (1) details on the curation of the datasets used for training and evaluation, (2) implementation details of our diffusion models, (3) additional results and (4) the limitations of Z3D.

\section{Training dataset curation}
To effectively train and evaluate Z3D for novel depth synthesis, we construct sequences of cameras with strong geometric overlap. We select a diverse set of datasets that provide depth maps and camera annotations, enabling the generation of such camera sequences. We considered the following datasets with the statistics shown in Table~\ref{tab:dataset_summary}. 
%these statistics.

\begin{table}[tb]
\centering
\begin{tabular}{l c c}
\toprule
Dataset & Total Scenes & Total Images \\
\midrule
Taskonomy\cite{Zamir_2018_CVPR} (Tiny) & 35 & 381,840 \\
Replica\cite{replica19arxiv} & 18 & 104,397 \\
Hypersim\cite{Roberts_2021_ICCV} & 457 & 74,619 \\
Habitat-Matterport\cite{9010745} & 900 &900,000 \\
MegaDepth\cite{Li_2018_CVPR} & 196 & 130,000 \\

\bottomrule
\end{tabular}
\caption{Total number of scenes and total number of images for each dataset used in our experiments.}
\label{tab:dataset_summary}
\end{table}

For each scene in the datasets, we use all available camera poses if the scene contains fewer than 1,000 images; otherwise, we evenly subsample 1,000 cameras. Depth maps are projected across all camera pairs to compute an N×N symmetric camera-overlap matrix. Each entry is obtained by measuring frustum overlap in both directions and taking the minimum value, ensuring mutual visibility (co-visibility) between the two views. This matrix is then used to extract camera sequences with strong geometric overlap for every view in the scene. For Habitat-Matterport, we used the Habitat 2.0 simulator to generate 1,000 RGBD observations per scene.

We follow a similar strategy for all evaluation datasets, including DTU~\cite{aanaes2016large}, 7‑Scenes\cite{10.1109/CVPR.2013.377}, and NRGB‑D\cite{Azinovic_2022_CVPR}. For both training and evaluation, we use the official splits provided by each dataset. Tab. ~\ref{tab:dataset_summary} summarizes the statistics of each training dataset.

\section{Implementation details}
\parnobf{Network Architecture}
Tab. \ref{tab:dit_architecture} summarizes the modified DiT model used in Z3D. 

\begin{table}[tb]
\centering
\caption{The diffusion timestep is encoded by the TimestepEmbedder (Step 1) and camera poses by the PoseEncoder (Step 2). Sampled Noise is modulated with positional embeddings and RoPE for spatial structure (Step 3). The DiTBlocks (Step 4) perform self-attention, cross-attention to source views,and the output is fed into an MLP layer. Finally, the FinalLayer (Step 5) applies adaLN-Zero modulation\cite{10377858}, cross-attention, and an MLP to produce the output features. Note   that $S_t, S,P, C$ are the number of target views, number of source views, number of tokens, and token dimension respectively.}
\resizebox{\textwidth}{!}{%
\begin{tabular}{l p{5cm} p{7cm} l}
\toprule
Step & Inputs & Operation & Output Shape \\
\midrule
1 & Timestep $t$ & TimestepEmbedder MLP & $C$ \\
2 & Camera poses $(S, 12)$ & PoseEncoder: Fourier features + MLP & $S \times C$ \\
2 & Pose embed $(S, C)$ & Pose Embedding is added to timestep embedding & $S \times C$ \\
3 & Target tokens/sampled noise $x$ & Add sin-cos  and RoPe positional embedding & $S_t \times P \times C$ \\
4 & (3), Time + pose embedding, Source tokens & DiTBlocks: self-attn. + cross attn. + MLP & $S_t \times P \times C$ \\
5 & (4) & FinalLayer: adaLN-Zero + cross-attn. + MLP & $ S_t \times P \times C$ \\
\bottomrule
\end{tabular}%
}
\label{tab:dit_architecture}
\end{table}
\parnobf{Training}
Our method is implemented in \textbf{PyTorch} and trained in two stages. 
In \textbf{Stage 1}, the model is trained on pairs of target and source views with an effective batch size of 128 for 98k steps. 
In \textbf{Stage 2}, we train the model using multiple views (2 source views and 4 target views) with an effective batch size of 32 for 156k steps. We train the model using AdamW optimizer with $\beta_{1} = 0.90, \beta_{2} = 0.95$ and learning rate, $lr$  of 2e-4.

We observe that training the model on high-dimensional latents, such as those produced by 3DFMs, does not work out-of-the-box with the default noise schedulers used in popular diffusion architectures. 
Following the empirical findings of~\cite{zheng2025diffusiontransformersrepresentationautoencoders}, we modify the \textit{width} of the DiT model to match the token dimension of the 3DFM latents.

Previous diffusion studies\cite{10.5555/3692070.3692573,zheng2025diffusiontransformersrepresentationautoencoders} have also shown that increasing the spatial resolution reduces information corruption at the same noise level, which otherwise degrades diffusion training. 
Following~\cite{zheng2025diffusiontransformersrepresentationautoencoders}, we therefore make the timestep shift in our noise scheduler data-dependent. 
Specifically, we adjust the timestep shift factor using a dimension-dependent scaling rule, $\alpha = \sqrt{m/n}$, where $m = n_{\text{target view}}\times P\times C$ represents the dimensionality of the latent tokens and $n = 4096$ is the base token dimension used for scaling. 
Here, $P$ denotes the number of spatial tokens per view ($P = 1369$ for both VGGT\cite{wang2025vggt} and WorldMirror\cite{liu2025worldmirror}), and $C$ is the channel dimension (C =2048 for both VGGT and WorldMirror) and $n_{\text{target view}}$ is the number of target views.

\section{Additional LDI Results}
As described in the main paper, we provide additional LDI prediction results in Figure \ref{fig:ldi_results} for both the pretrained VGGT and WorldMirror backbones.

\begin{figure}[!t]
    \centering
    \includegraphics[page=1,width=\linewidth]{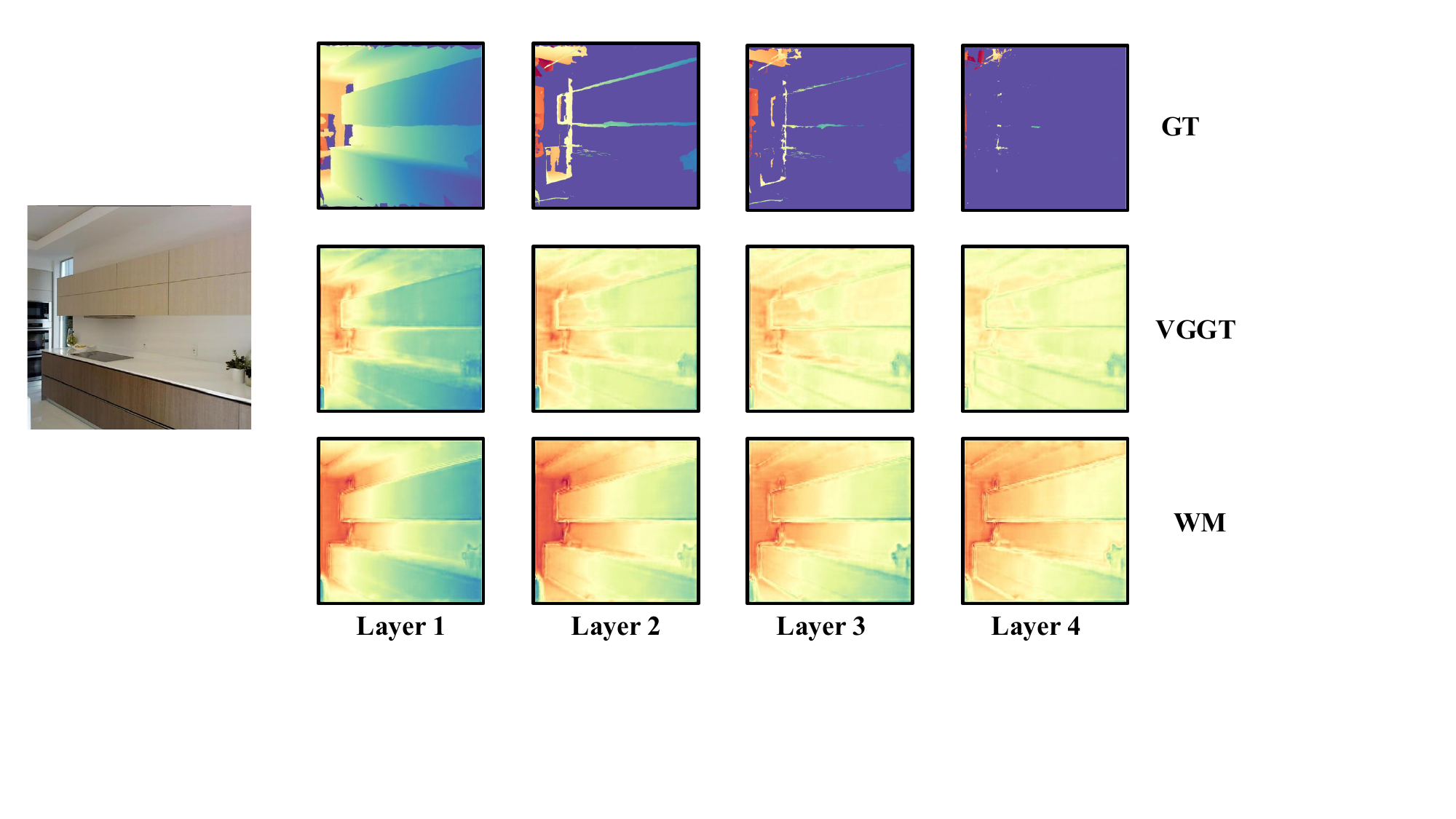}
    \caption{
    {\bf Linear Probe Layered Depth Image (LDI) prediction capability of 3DFMs.}
    To test whether 3DFMs implicitly contain information about unseen views, we train a linear model to predict LDIs. Good performance on LDI prediction after the first layer suggests that the model understands the full 3D scene, including hidden surfaces. For each example: (left) input RGB; (top) ground-truth LDI layers; (bottom) linearly decoded LDI layers. The ability to recover the non-trivial structure of the scene with only a linear layer suggests that 3DFMs implicitly encode this information.
    }
    \label{fig:ldi_results}
\end{figure}

\section{Which layer features of 3DFM backbones should be used for diffusion?}
Recent 3DFM architectures such as VGGT~\cite{wang2025vggt} and WorldMirror~\cite{liu2025worldmirror} aggregate transformer tokens from multiple intermediate layers for depth prediction. 
Each token representation contains high-dimensional embeddings (e.g., 2048 dimensions in both VGGT and WorldMirror), making it computationally expensive to apply diffusion directly to the concatenated multi-layer outputs used for depth estimation.

In this section, we analyze the contribution of tokens from different backbone layers to depth prediction. 
Our analysis is inspired by observations in~\cite{zhang2025emergentextremeviewgeometry3d}, which indicate that certain intermediate layers carry stronger geometric signals.
The analysis in~\cite{zhang2025emergentextremeviewgeometry3d} identifies layers 4, 11, 17, and 23 as the most informative for geometric reasoning. 
However, applying diffusion to the tokens from all four layers simultaneously is computationally intractable due to the high dimensionality of the representations. 
Therefore, we analyze which of these layers is most relevant for depth prediction in order to select an appropriate layer for diffusion.

To perform this analysis, we infer depth using tokens from all possible combinations of these layers. 
We then evaluate which individual layer or subset of layers produces depth predictions that best approximate the depth inferred using the full set of layers.

As shown in Figures~\ref{vggt_layer_analysis} and \ref{worldmirror_layer_analysis}, for both VGGT and WorldMirror, using only features from layer 17 produces depth predictions that are nearly as good as using features from all four selected layers. Moreover, combinations that include layer 17 consistently outperform those that do not. This trend holds across single-layer, two-layer, and three-layer configurations, with layer 17 producing noticeably sharper depth maps across all datasets. These results suggest that the layer 17 tokens provide the most informative features for depth diffusion.

Based on this analysis, we restrict the diffusion process to the scene tokens extracted from layer 17 for the target view(s). 
This design substantially reduces computational cost while retaining the most depth-informative representation.

To mitigate potential information loss from restricting the diffusion tokens, we use the backbone layer outputs from all four aggregation layers of the source views as conditioning signals. 
Through cross-attention, this conditioning provides the model with richer multi-layer geometric information while keeping the diffusion process computationally tractable.

\newcommand{\dpwidth}{0.07\textwidth}
% \begin{table}[t]
\begin{figure*}[t]
\centering
\centering
\caption{Depth predictions for a target view using VGGT. Rows correspond to individual layers and layer combinations predictions from VGGT backbone used to predict the depth. Columns correspond to different input images.}
\resizebox{\textwidth}{!}{%
\begin{tabular}{lccccc}
\toprule
\textbf{Layers} & \textbf{Image 1} & \textbf{Image 2} & \textbf{Image 3} & \textbf{Image 4} & \textbf{Image 5} \\
\midrule

Target Image
& \includegraphics[width=\dpwidth]{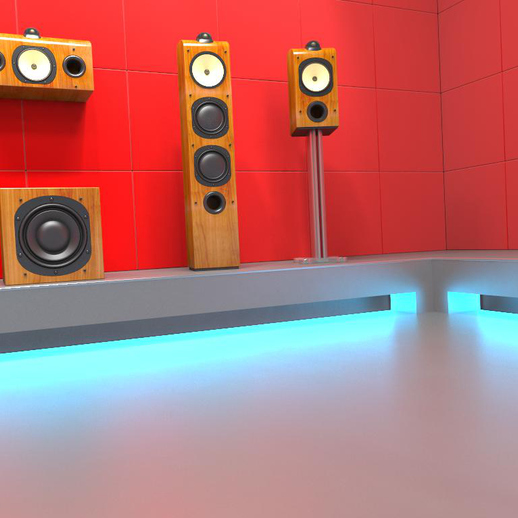}
& \includegraphics[width=\dpwidth]{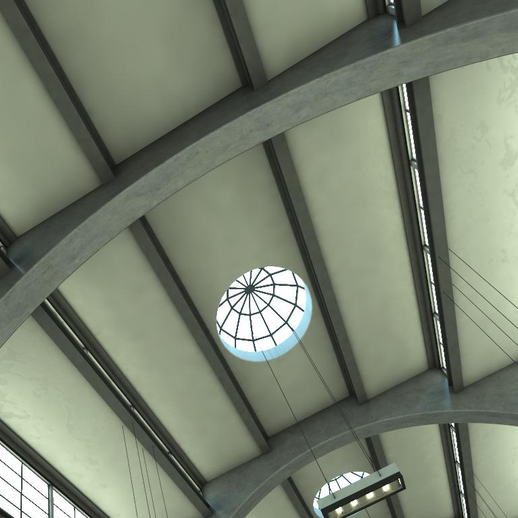}
& \includegraphics[width=\dpwidth]{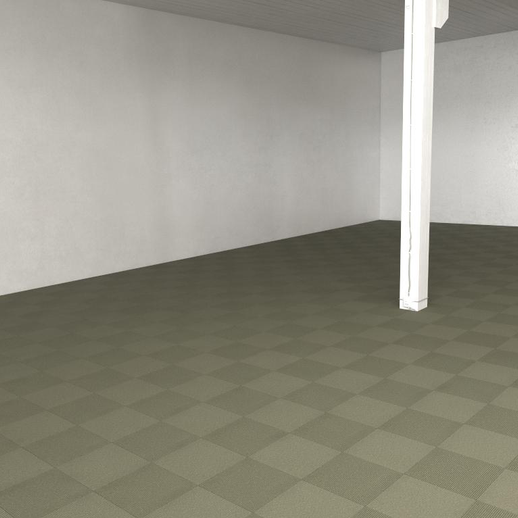}
& \includegraphics[width=\dpwidth]{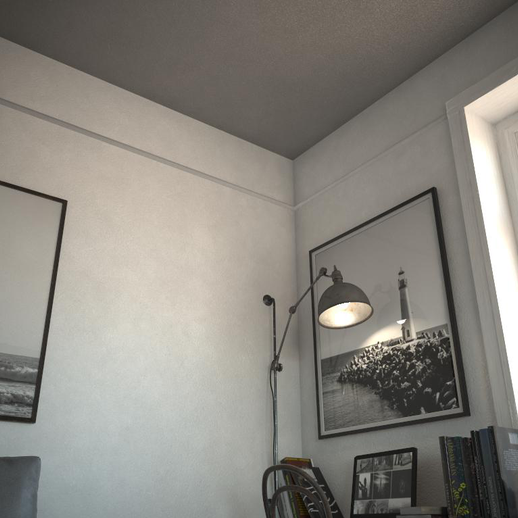}
& \includegraphics[width=\dpwidth]{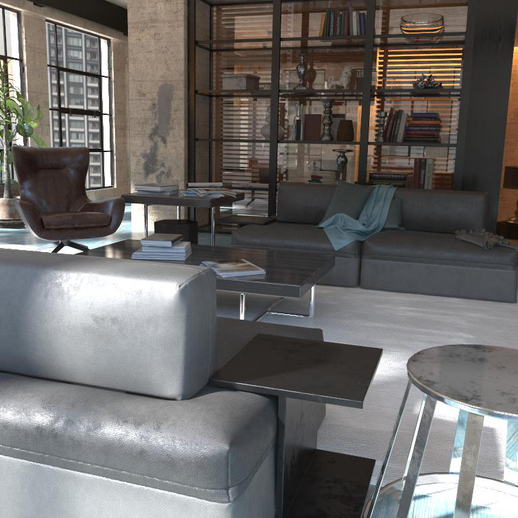} \\

\midrule

Layer 4
& \includegraphics[width=\dpwidth]{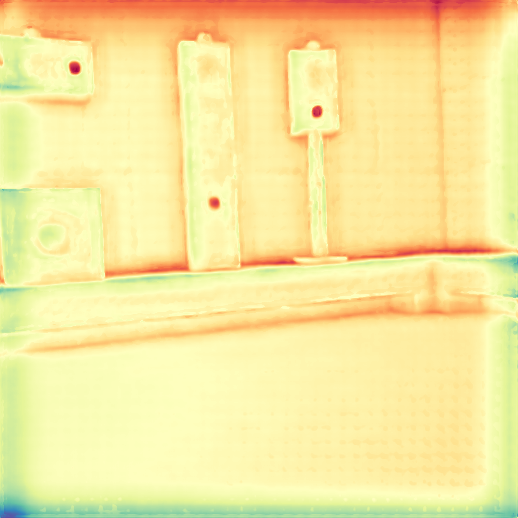}
& \includegraphics[width=\dpwidth]{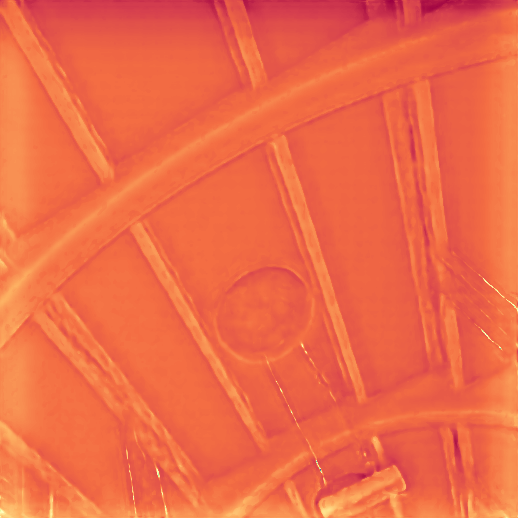}
& \includegraphics[width=\dpwidth]{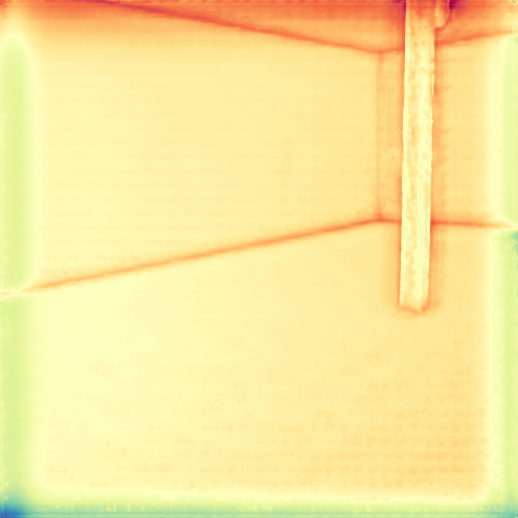}
& \includegraphics[width=\dpwidth]{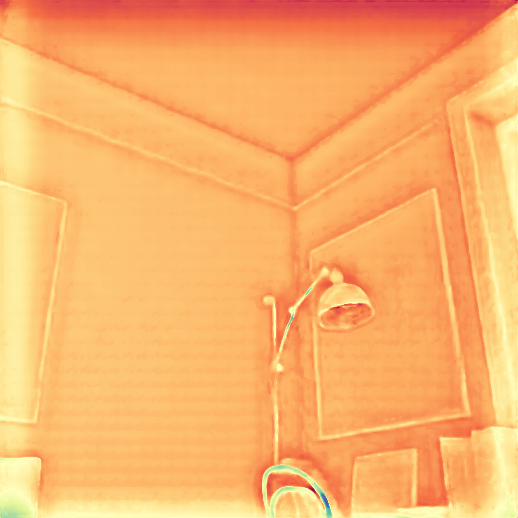}
& \includegraphics[width=\dpwidth]{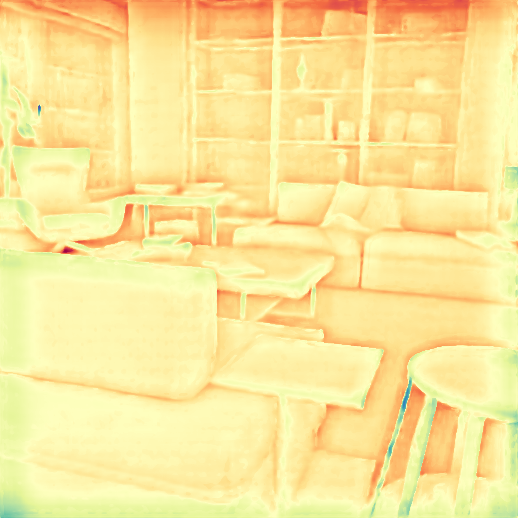} \\

Layer 11
& \includegraphics[width=\dpwidth]{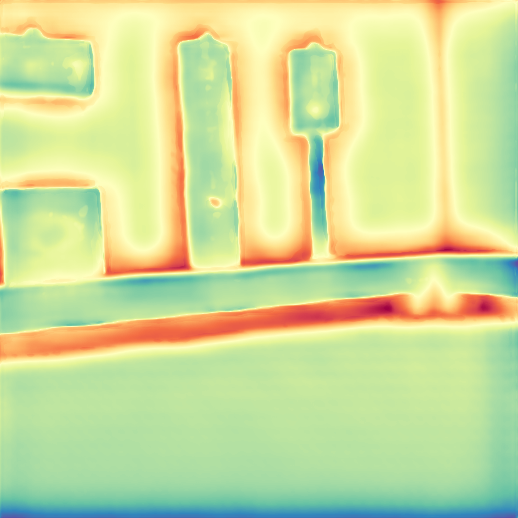}
& \includegraphics[width=\dpwidth]{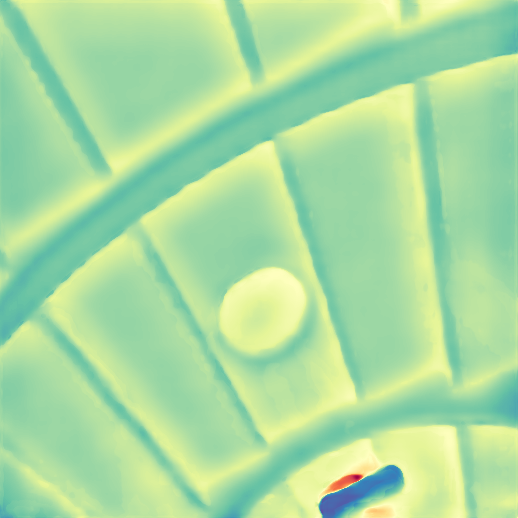}
& \includegraphics[width=\dpwidth]{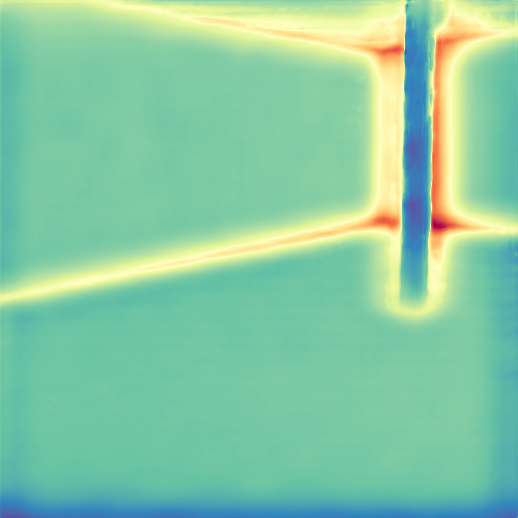}
& \includegraphics[width=\dpwidth]{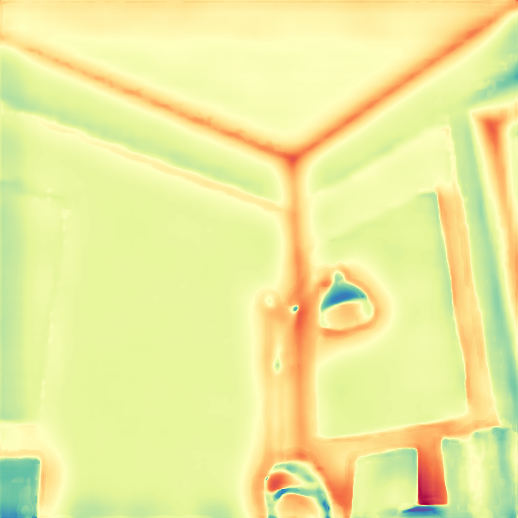}
& \includegraphics[width=\dpwidth]{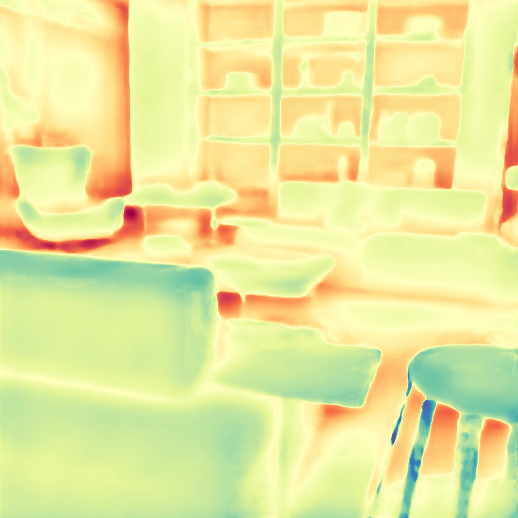} \\

Layer 17
& \includegraphics[width=\dpwidth]{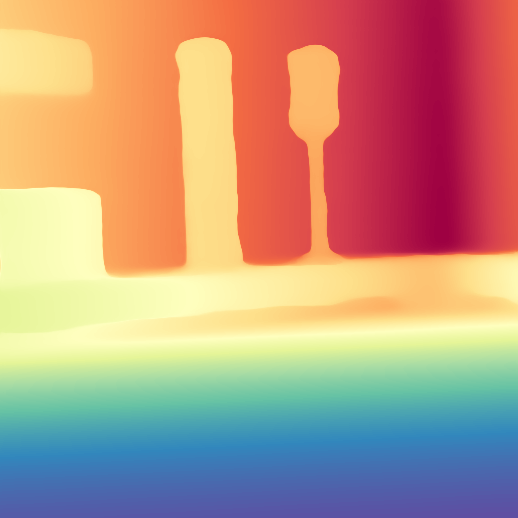}
& \includegraphics[width=\dpwidth]{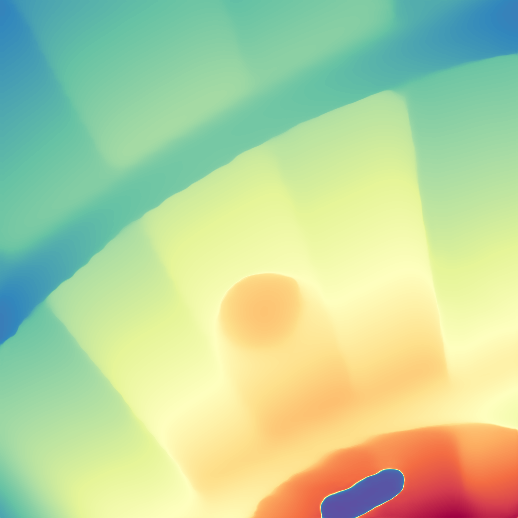}
& \includegraphics[width=\dpwidth]{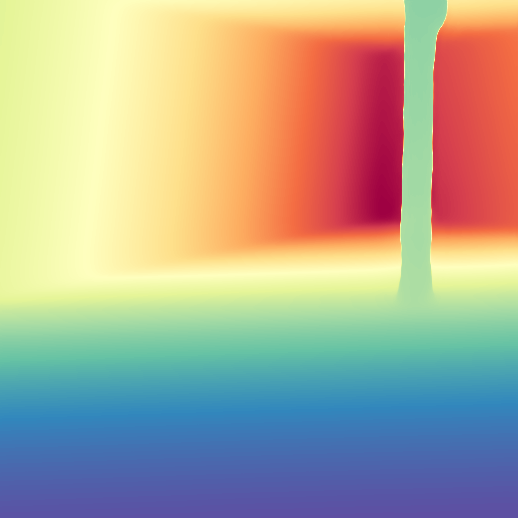}
& \includegraphics[width=\dpwidth]{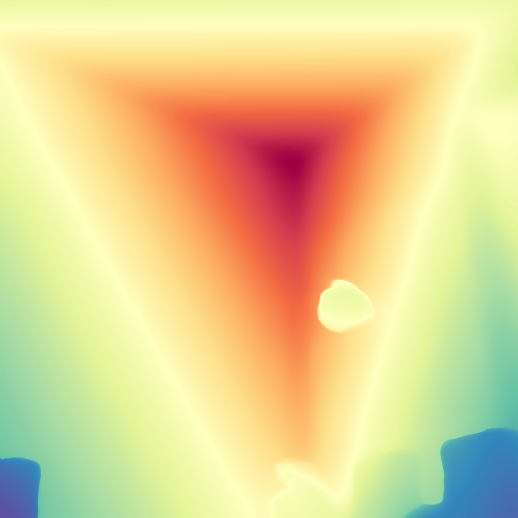}
& \includegraphics[width=\dpwidth]{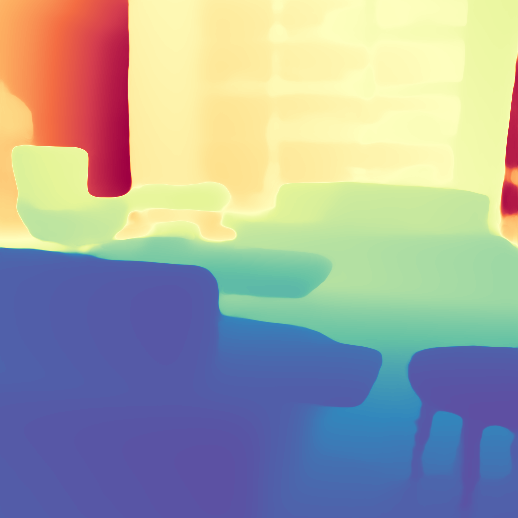} \\

Layer 23
& \includegraphics[width=\dpwidth]{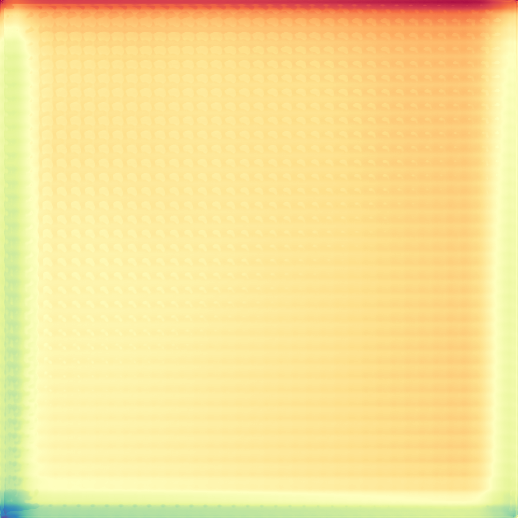}
& \includegraphics[width=\dpwidth]{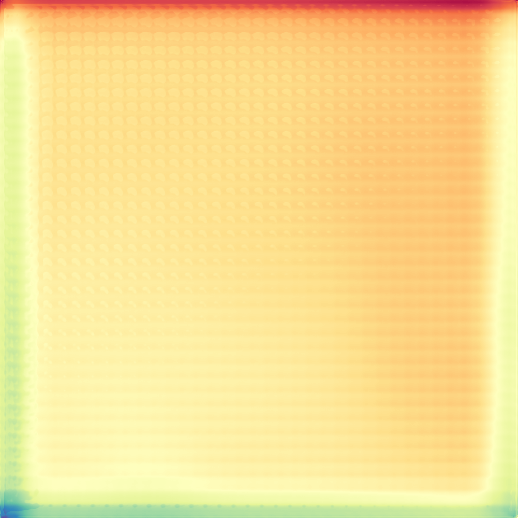}
& \includegraphics[width=\dpwidth]{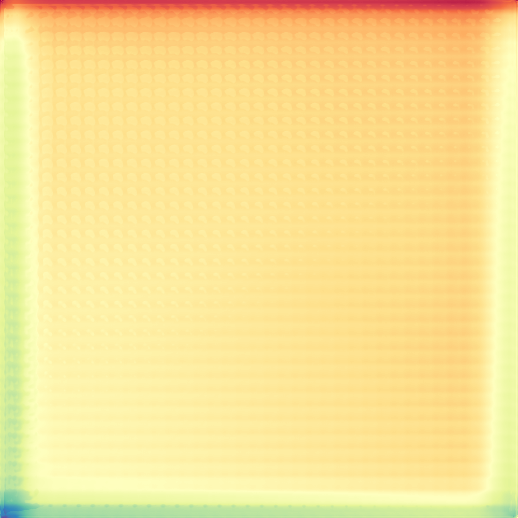}
& \includegraphics[width=\dpwidth]{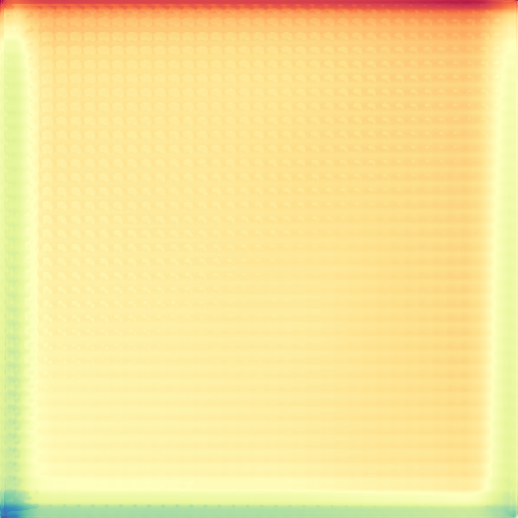}
& \includegraphics[width=\dpwidth]{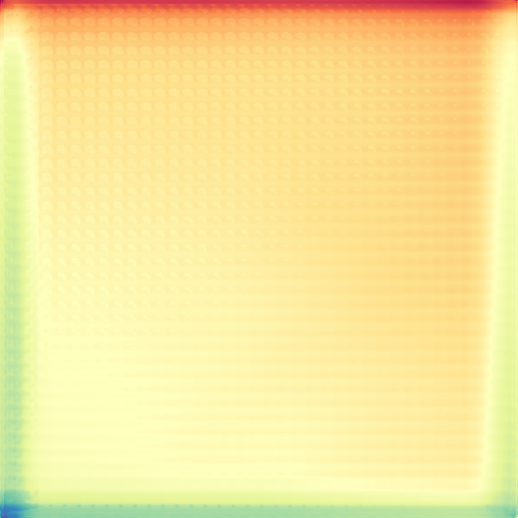} \\

4\_11
& \includegraphics[width=\dpwidth]{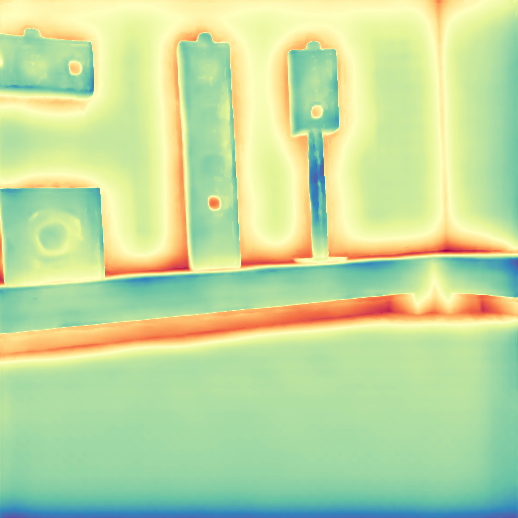}
& \includegraphics[width=\dpwidth]{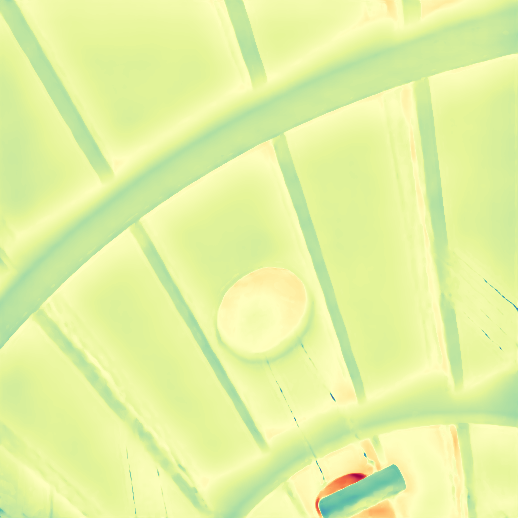}
& \includegraphics[width=\dpwidth]{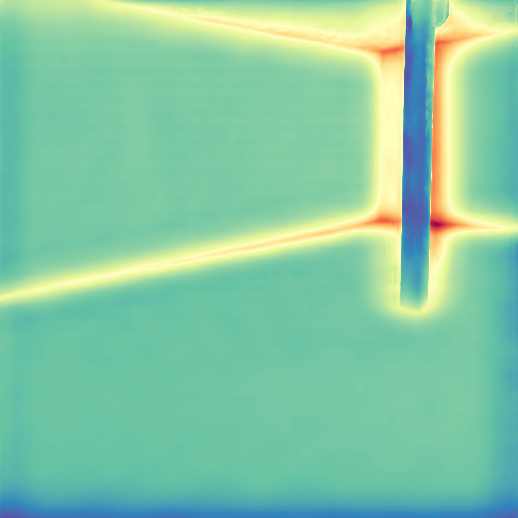}
& \includegraphics[width=\dpwidth]{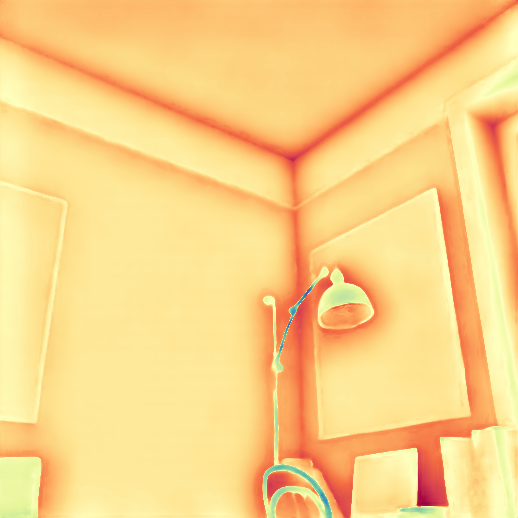}
& \includegraphics[width=\dpwidth]{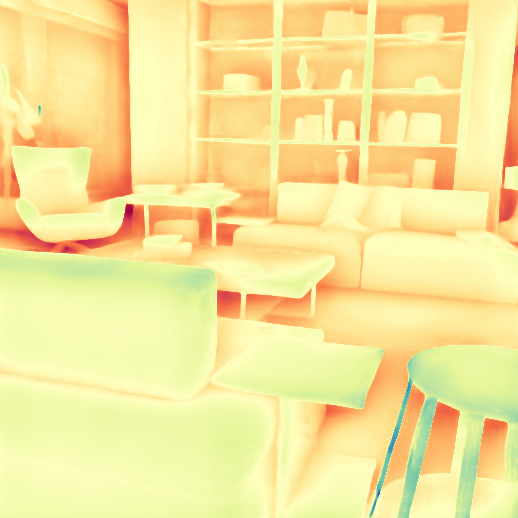} \\

4\_17
& \includegraphics[width=\dpwidth]{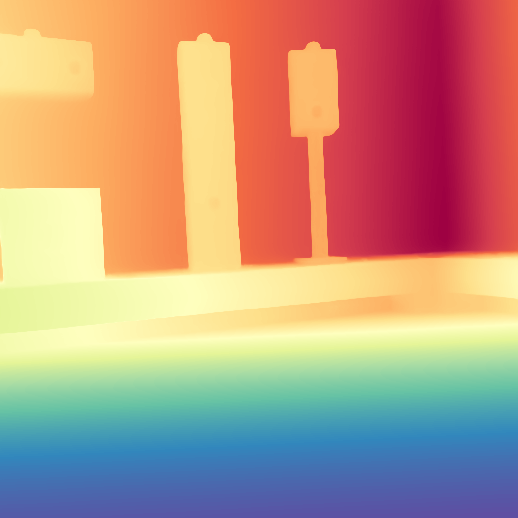}
& \includegraphics[width=\dpwidth]{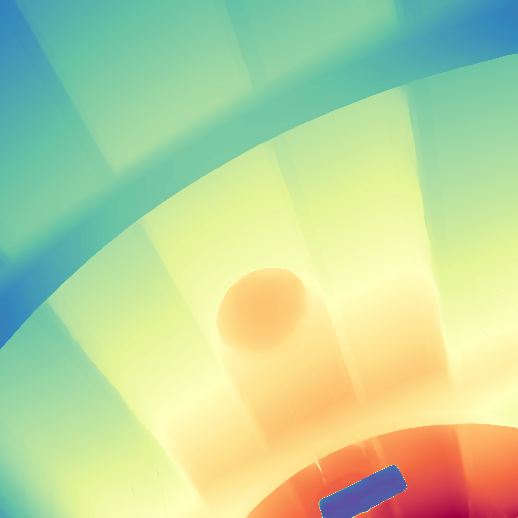}
& \includegraphics[width=\dpwidth]{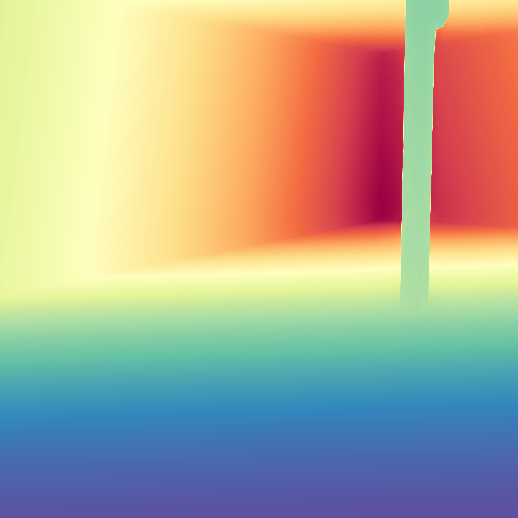}
& \includegraphics[width=\dpwidth]{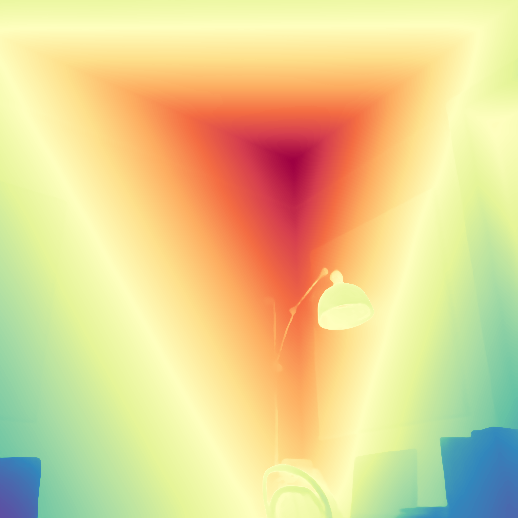}
& \includegraphics[width=\dpwidth]{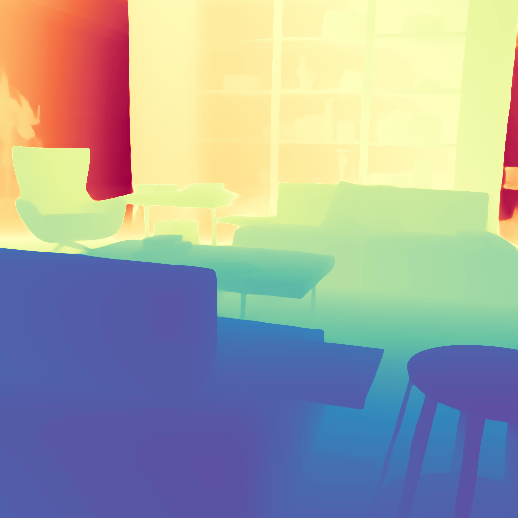} \\

11\_17
& \includegraphics[width=\dpwidth]{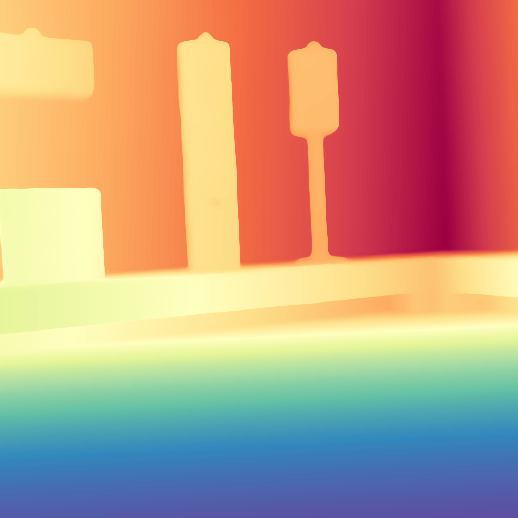}
& \includegraphics[width=\dpwidth]{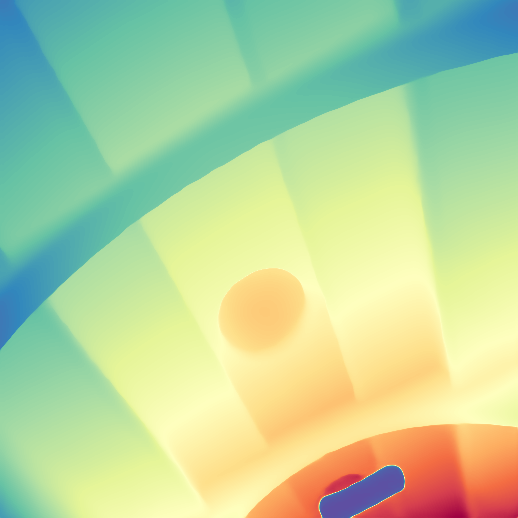}
& \includegraphics[width=\dpwidth]{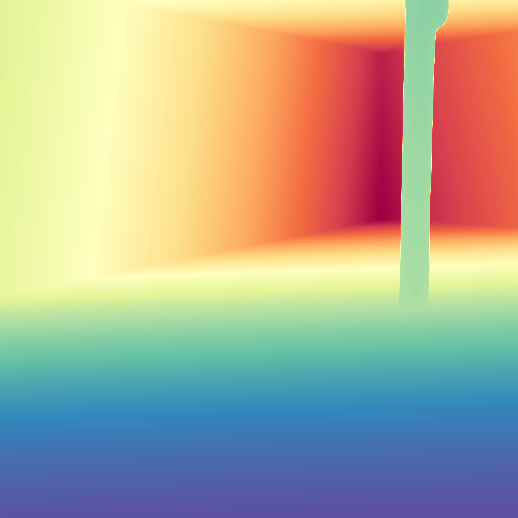}
& \includegraphics[width=\dpwidth]{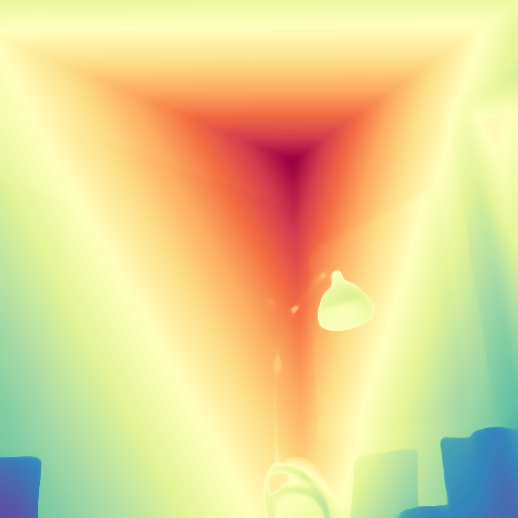}
& \includegraphics[width=\dpwidth]{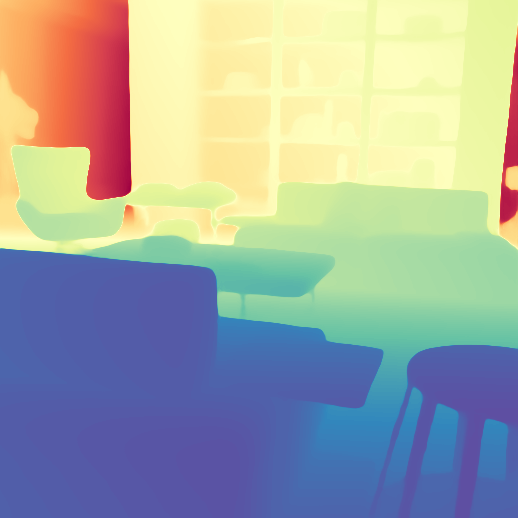} \\

4\_11\_17
& \includegraphics[width=\dpwidth]{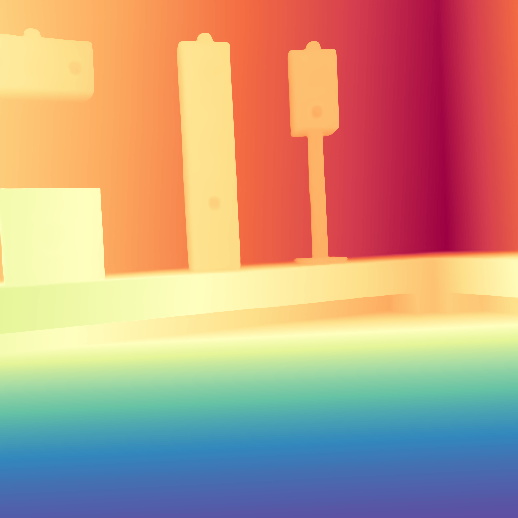}
& \includegraphics[width=\dpwidth]{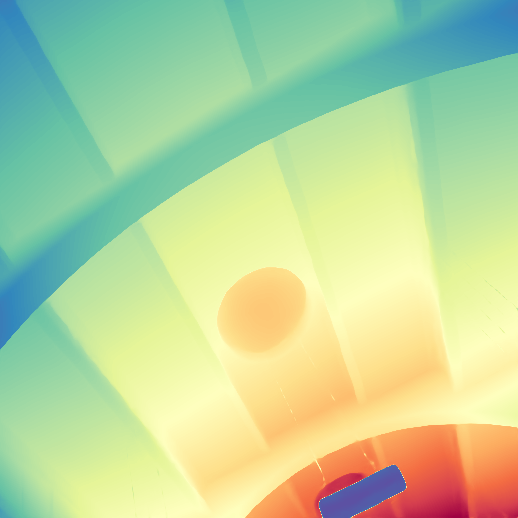}
& \includegraphics[width=\dpwidth]{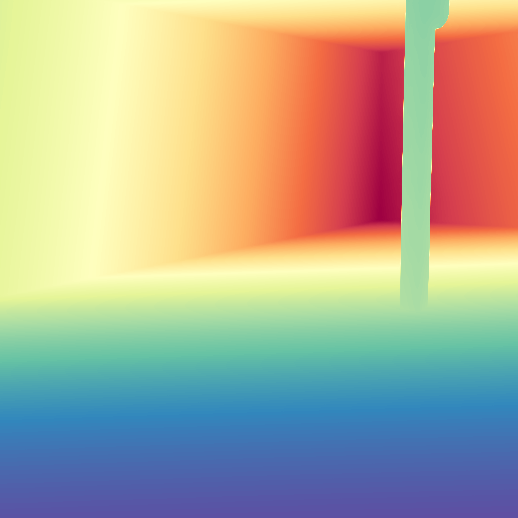}
& \includegraphics[width=\dpwidth]{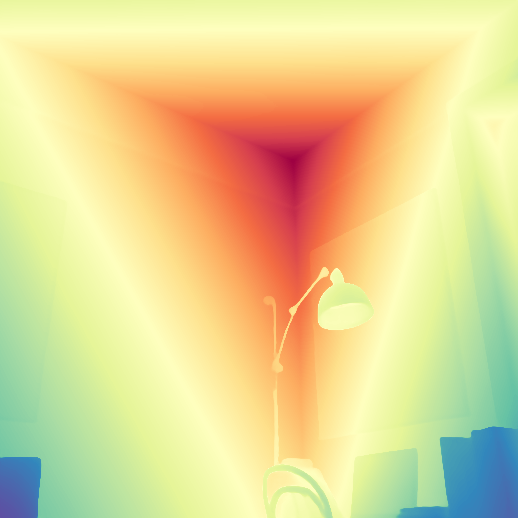}
& \includegraphics[width=\dpwidth]{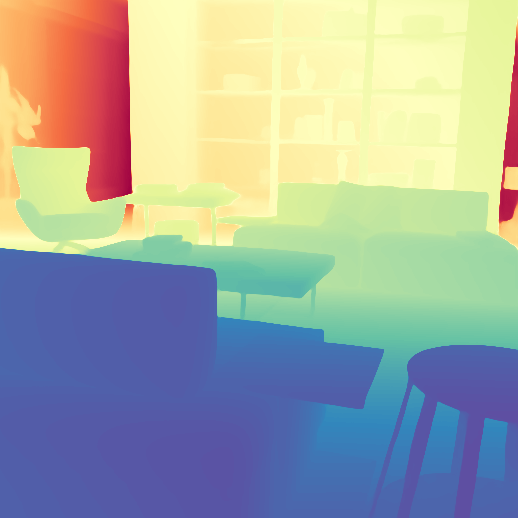} \\

4\_11\_23
& \includegraphics[width=\dpwidth]{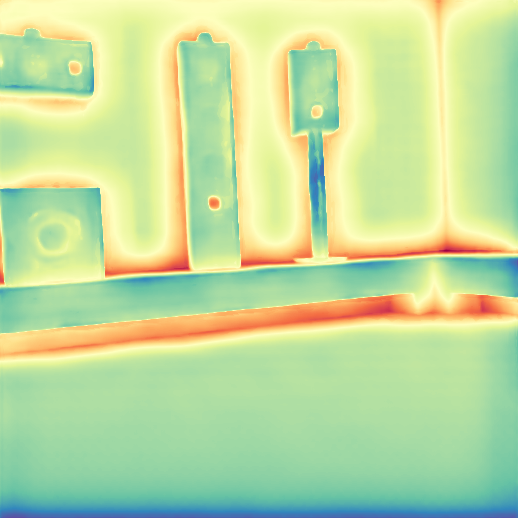}
& \includegraphics[width=\dpwidth]{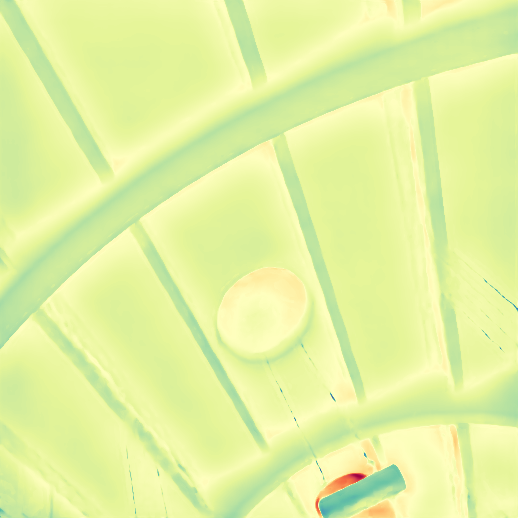}
& \includegraphics[width=\dpwidth]{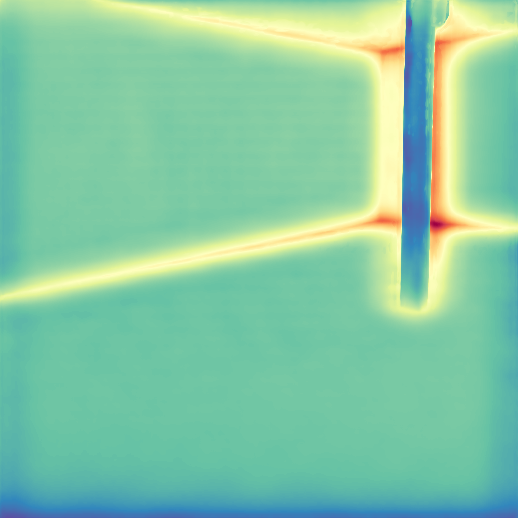}
& \includegraphics[width=\dpwidth]{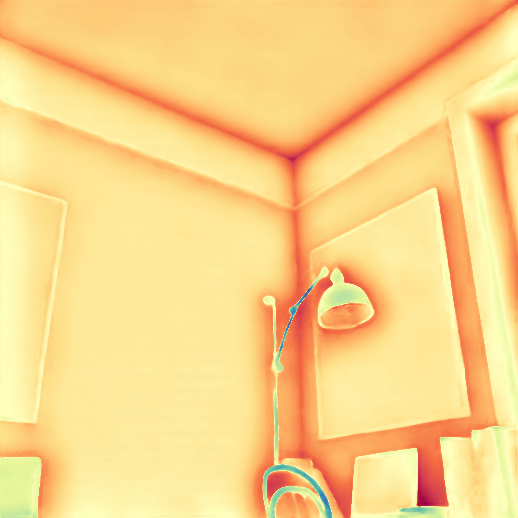}
& \includegraphics[width=\dpwidth]{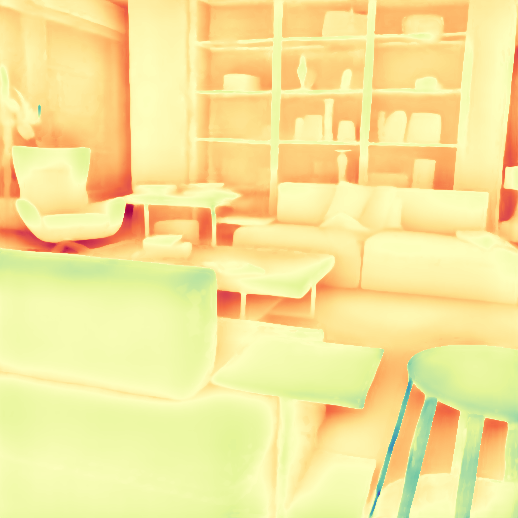} \\

4\_17\_23
& \includegraphics[width=\dpwidth]{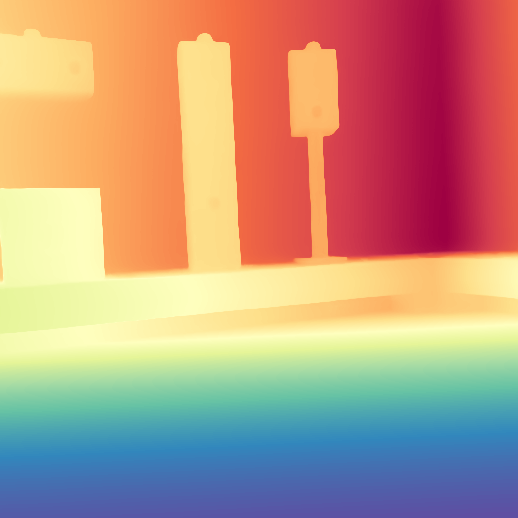}
& \includegraphics[width=\dpwidth]{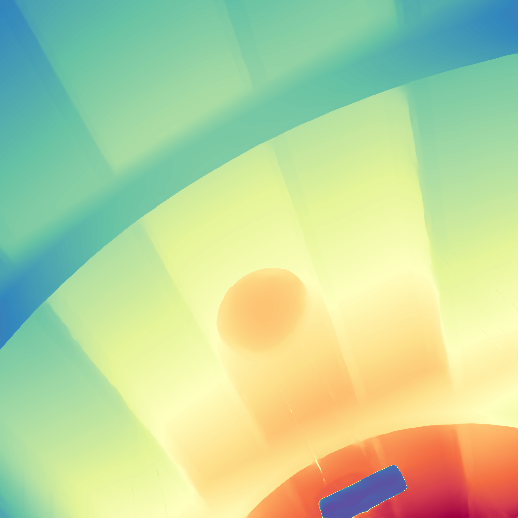}
& \includegraphics[width=\dpwidth]{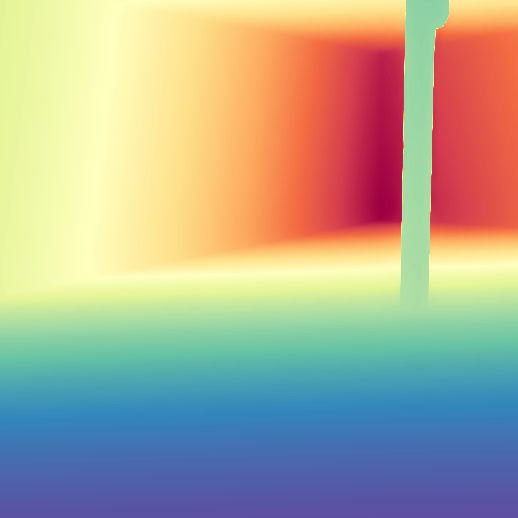}
& \includegraphics[width=\dpwidth]{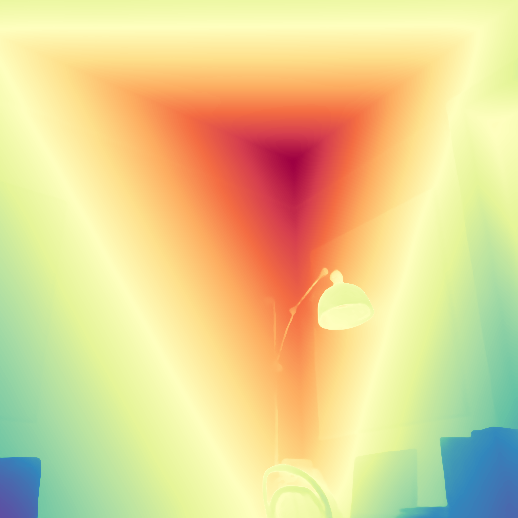}
& \includegraphics[width=\dpwidth]{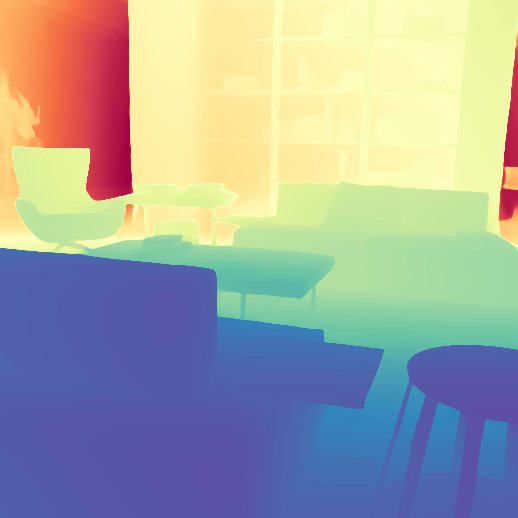} \\

All Layers
& \includegraphics[width=\dpwidth]{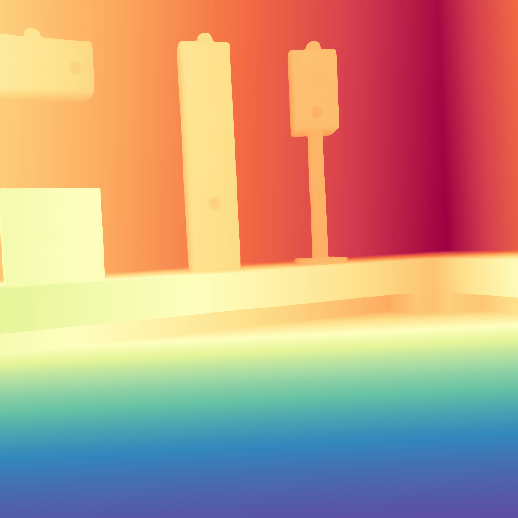}
& \includegraphics[width=\dpwidth]{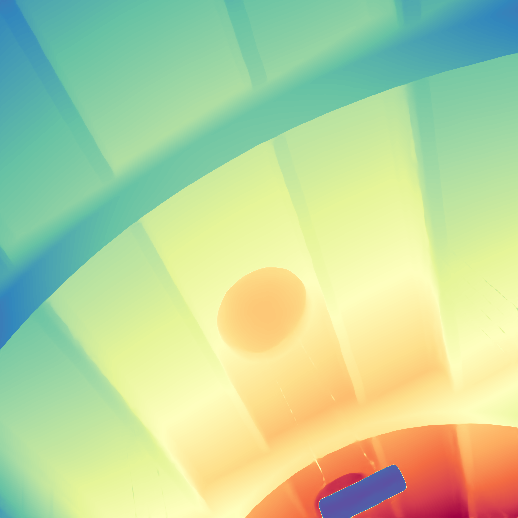}
& \includegraphics[width=\dpwidth]{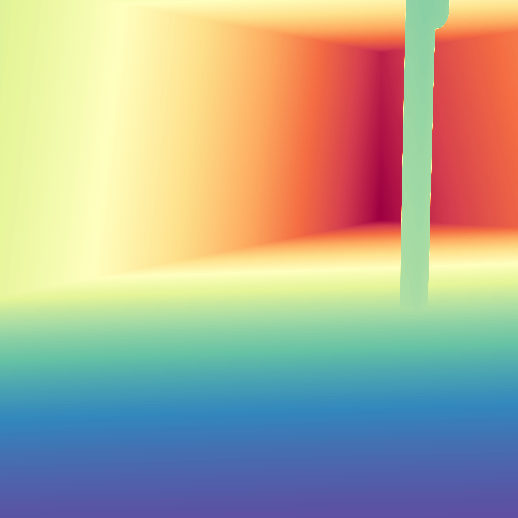}
& \includegraphics[width=\dpwidth]{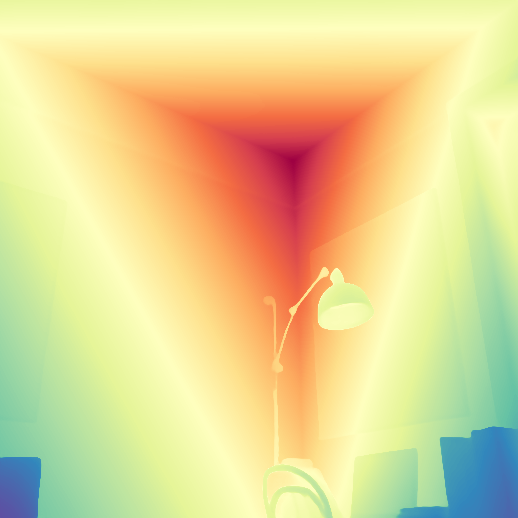}
& \includegraphics[width=\dpwidth]{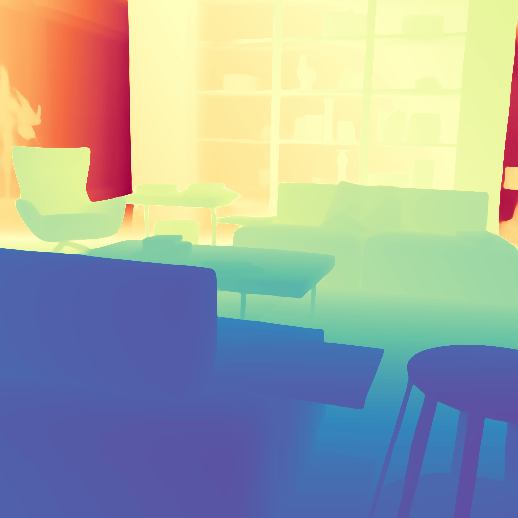} \\

\bottomrule
\end{tabular}
}
\label{vggt_layer_analysis}
\end{figure*}

\begin{figure*}[t]
\centering
\caption{Depth predictions for a target view using WorldMirror. Rows correspond to individual layers and layer combinations predictions from WorldMirror backbone used to predict the depth. Columns correspond to different input images.}
\resizebox{\textwidth}{!}{%
\begin{tabular}{lccccc}
\toprule
\textbf{Layers} & \textbf{Image 1} & \textbf{Image 2} & \textbf{Image 3} & \textbf{Image 4} & \textbf{Image 5} \\
\midrule

Target Image
& \includegraphics[width=\dpwidth]{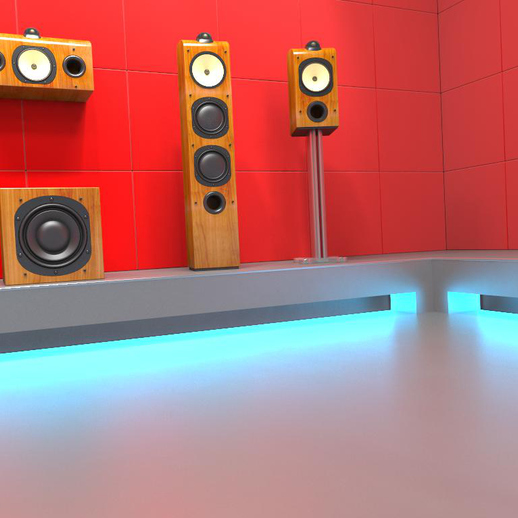}
& \includegraphics[width=\dpwidth]{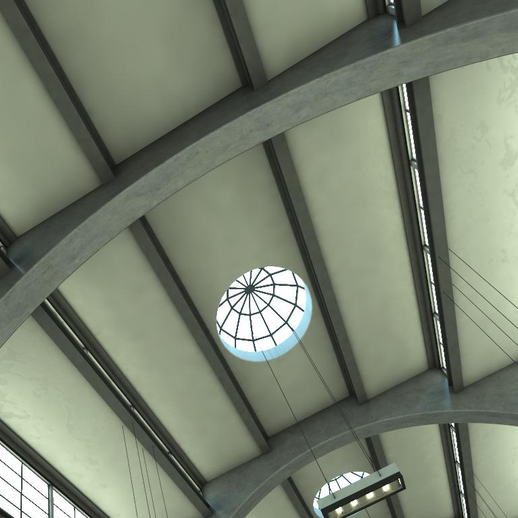}
& \includegraphics[width=\dpwidth]{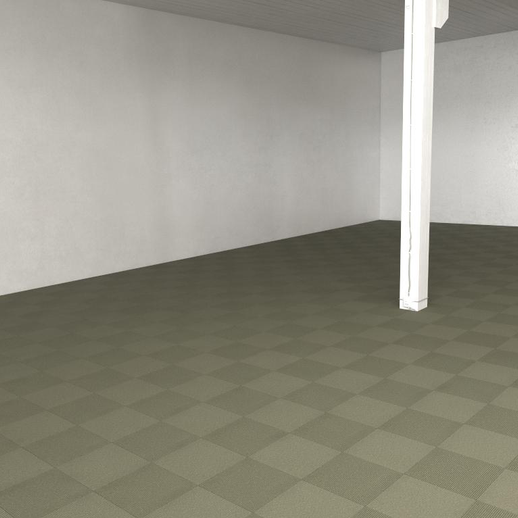}
& \includegraphics[width=\dpwidth]{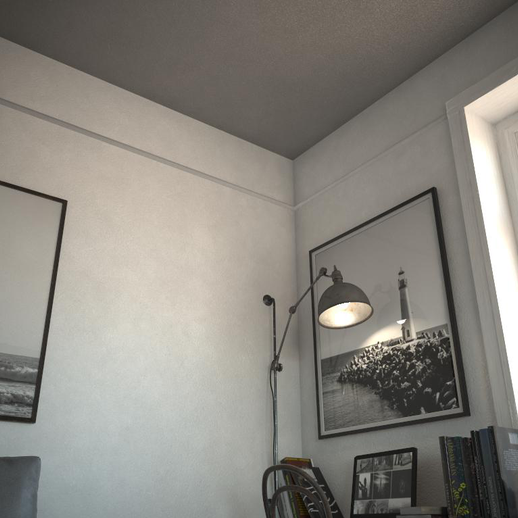}
& \includegraphics[width=\dpwidth]{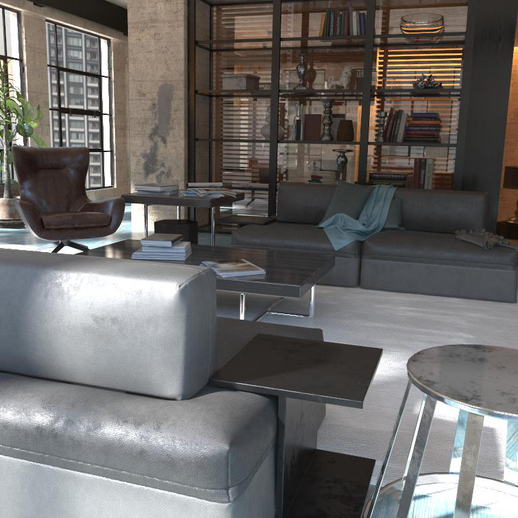} \\

\midrule

Layer 4
& \includegraphics[width=\dpwidth]{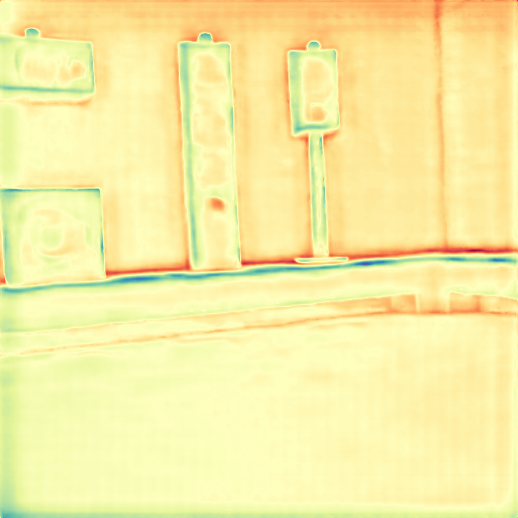}
& \includegraphics[width=\dpwidth]{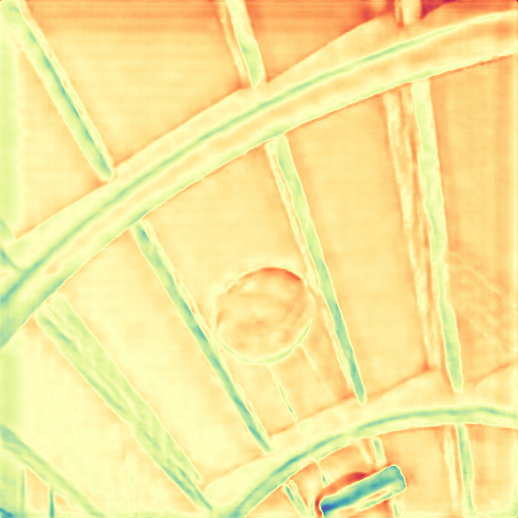}
& \includegraphics[width=\dpwidth]{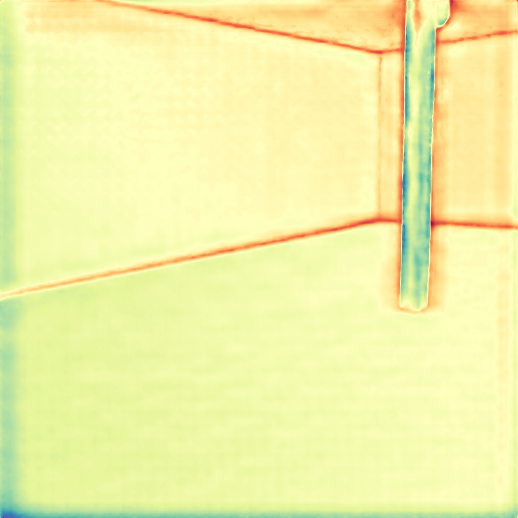}
& \includegraphics[width=\dpwidth]{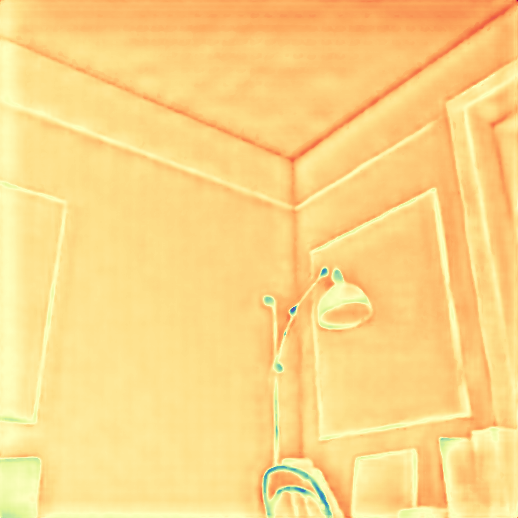}
& \includegraphics[width=\dpwidth]{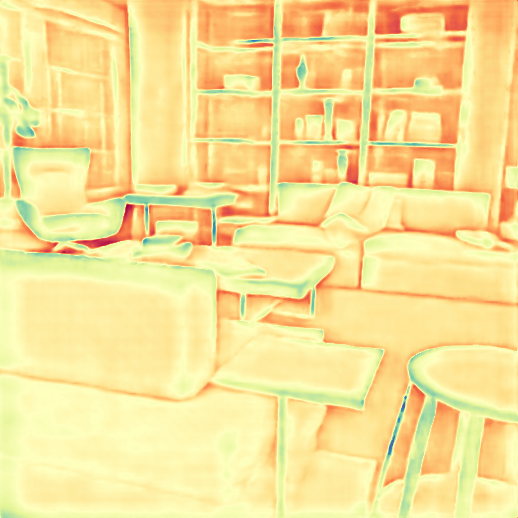} \\

Layer 11
& \includegraphics[width=\dpwidth]{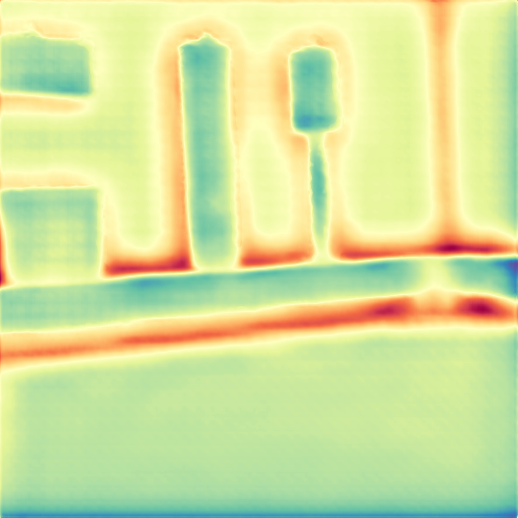}
& \includegraphics[width=\dpwidth]{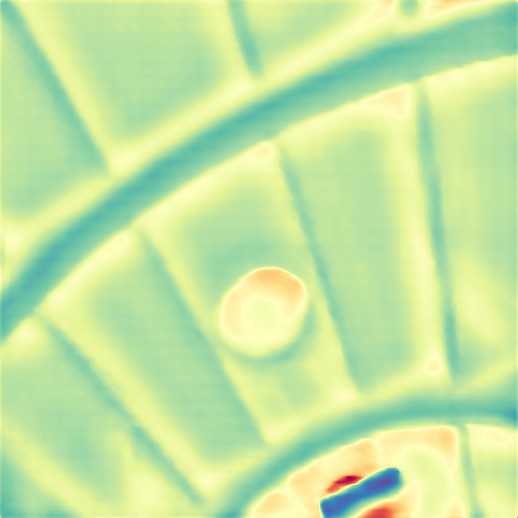}
& \includegraphics[width=\dpwidth]{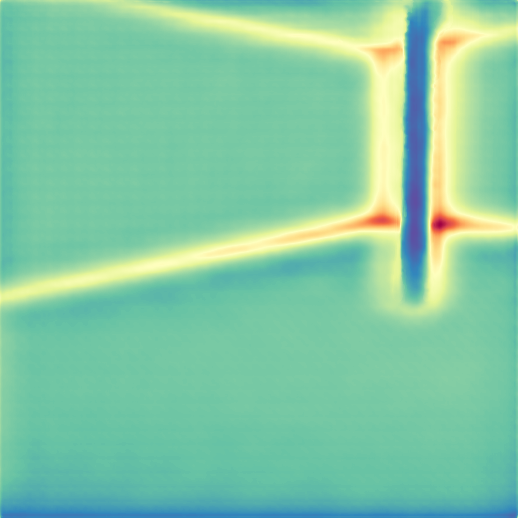}
& \includegraphics[width=\dpwidth]{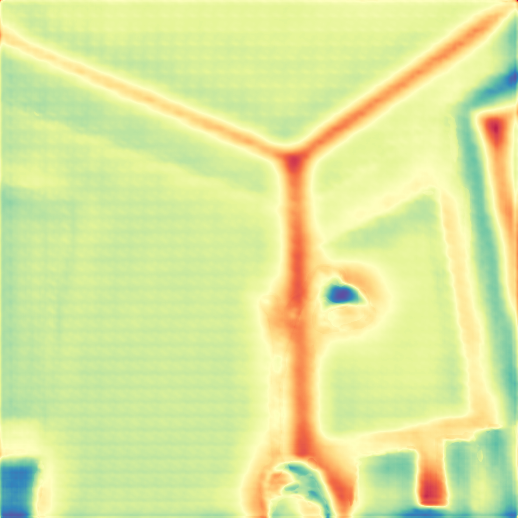}
& \includegraphics[width=\dpwidth]{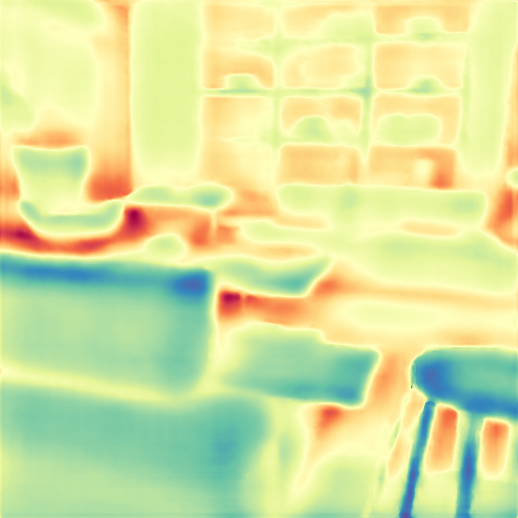} \\

Layer 17
& \includegraphics[width=\dpwidth]{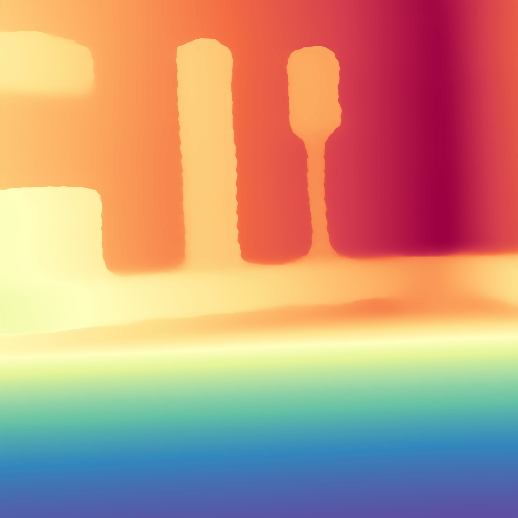}
& \includegraphics[width=\dpwidth]{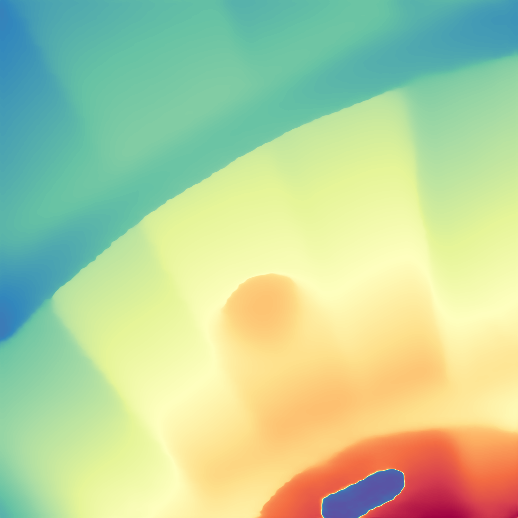}
& \includegraphics[width=\dpwidth]{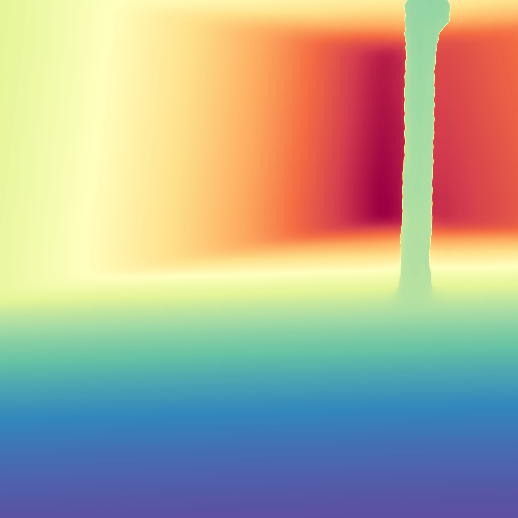}
& \includegraphics[width=\dpwidth]{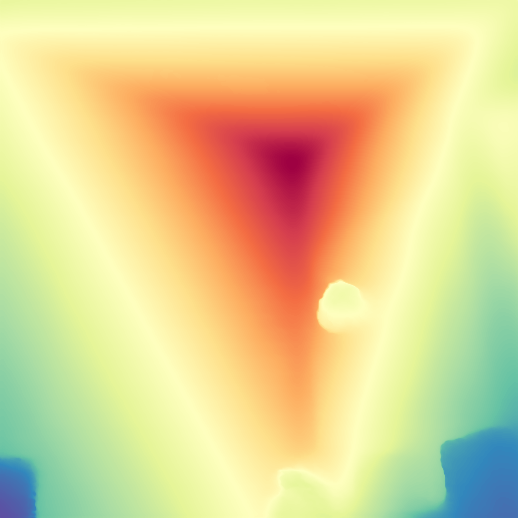}
& \includegraphics[width=\dpwidth]{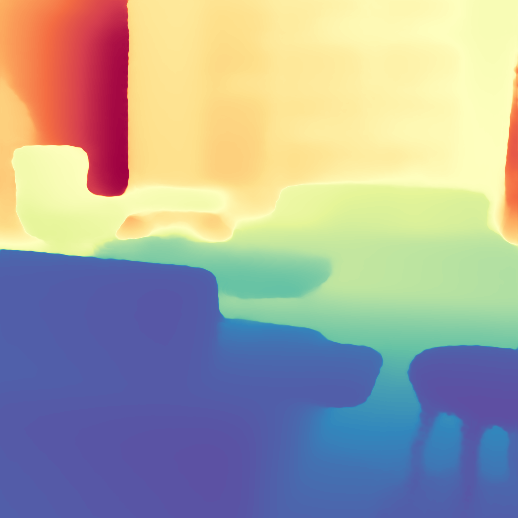} \\

Layer 23
& \includegraphics[width=\dpwidth]{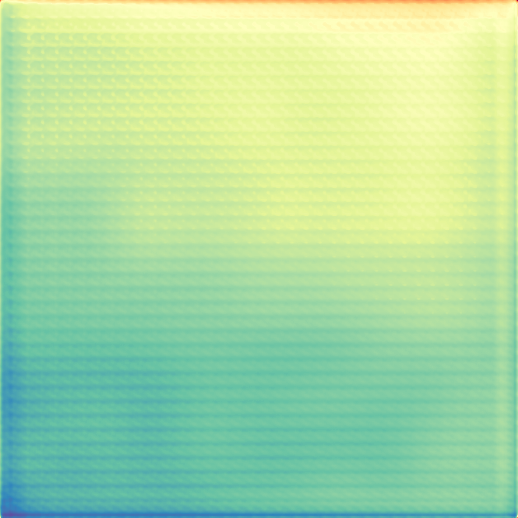}
& \includegraphics[width=\dpwidth]{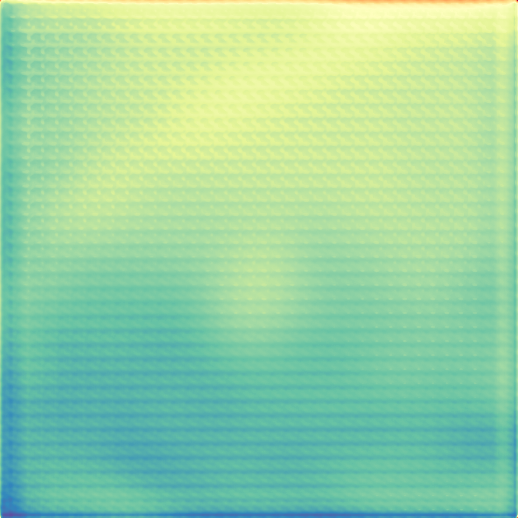}
& \includegraphics[width=\dpwidth]{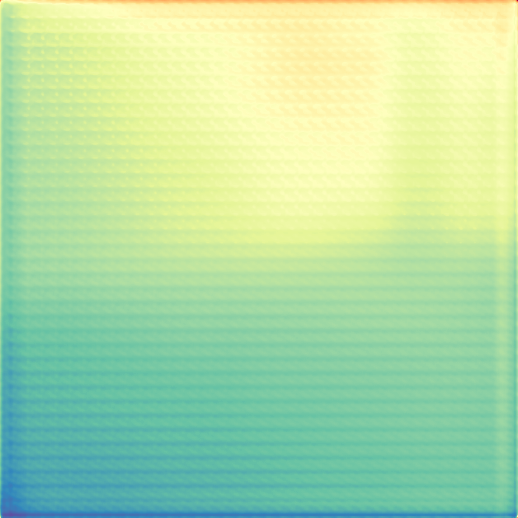}
& \includegraphics[width=\dpwidth]{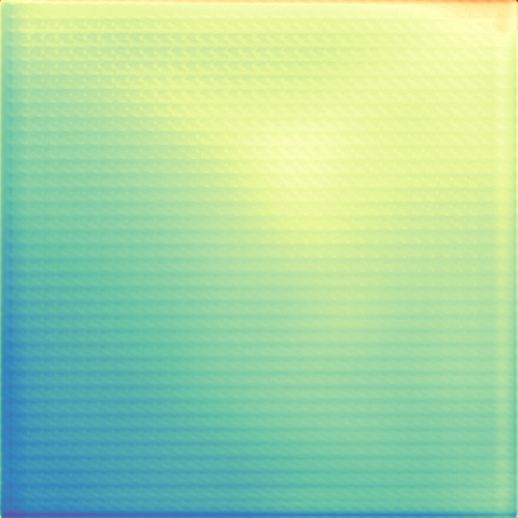}
& \includegraphics[width=\dpwidth]{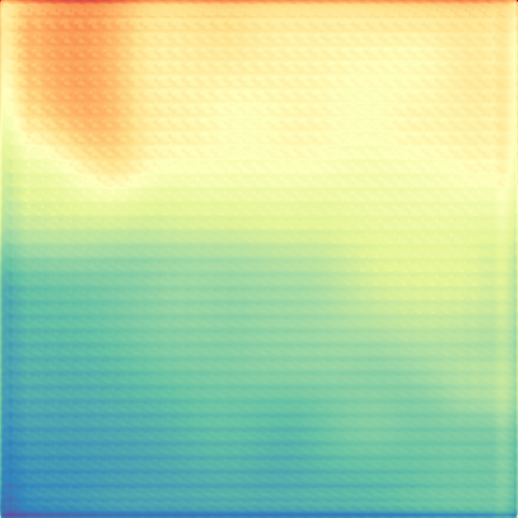} \\

4\_11
& \includegraphics[width=\dpwidth]{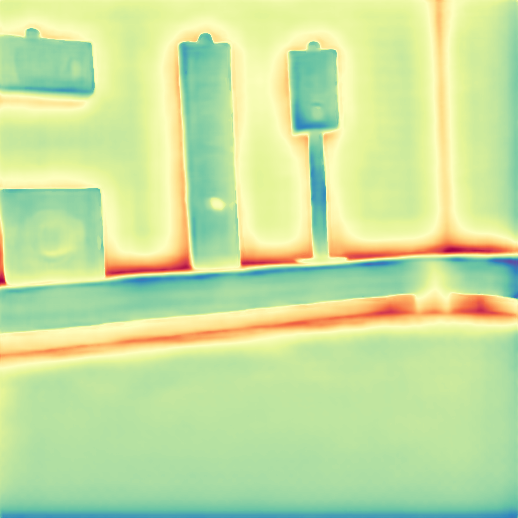}
& \includegraphics[width=\dpwidth]{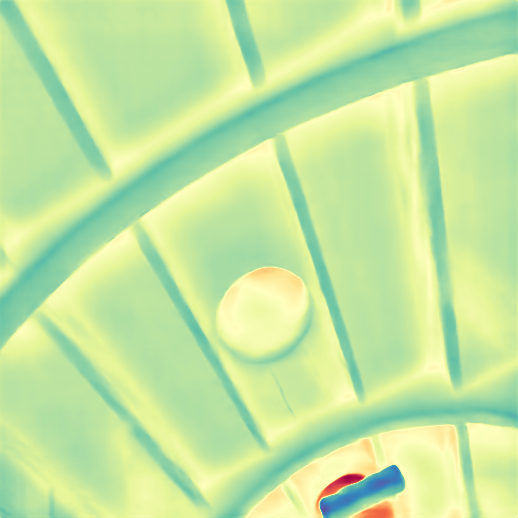}
& \includegraphics[width=\dpwidth]{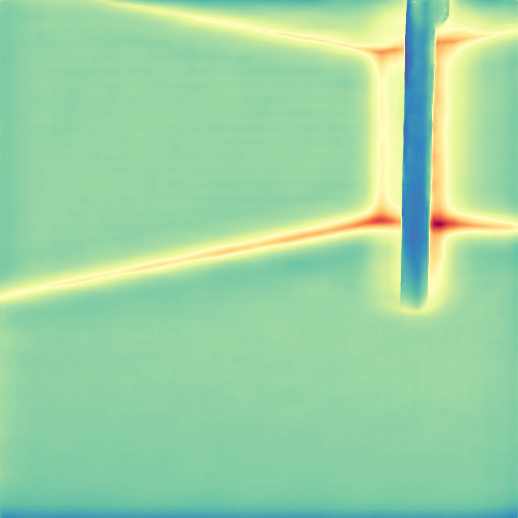}
& \includegraphics[width=\dpwidth]{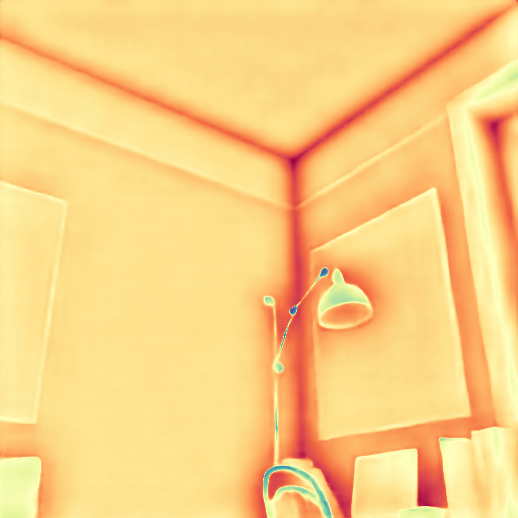}
& \includegraphics[width=\dpwidth]{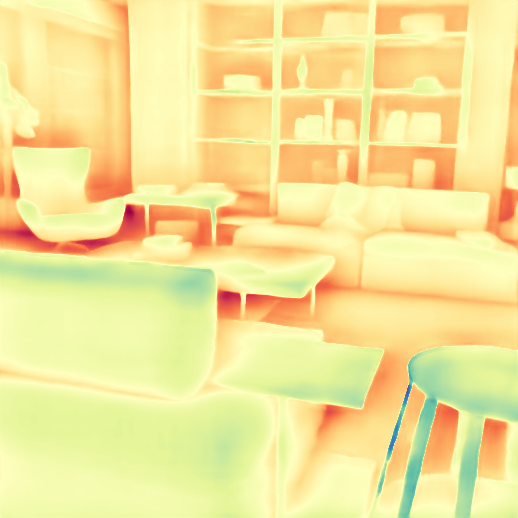} \\

4\_17
& \includegraphics[width=\dpwidth]{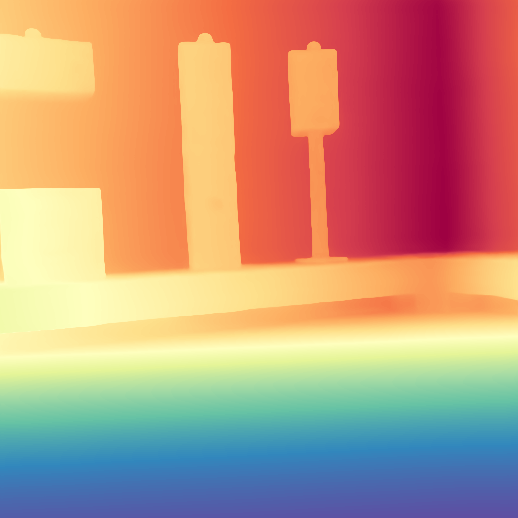}
& \includegraphics[width=\dpwidth]{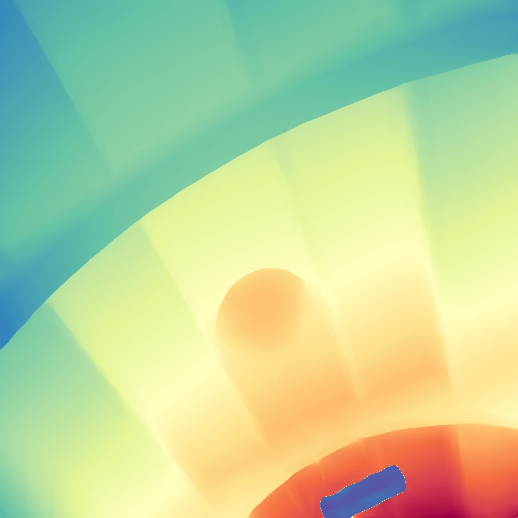}
& \includegraphics[width=\dpwidth]{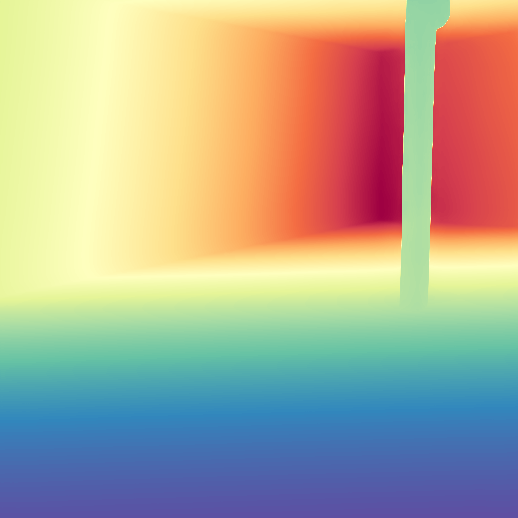}
& \includegraphics[width=\dpwidth]{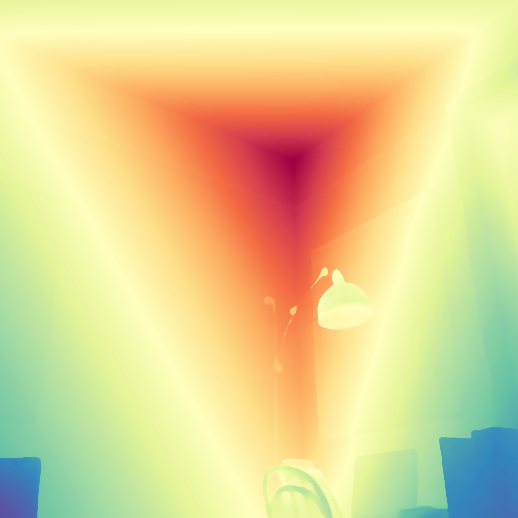}
& \includegraphics[width=\dpwidth]{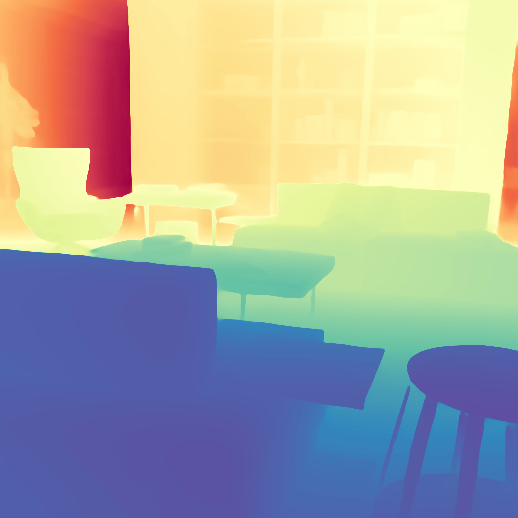} \\

11\_17
& \includegraphics[width=\dpwidth]{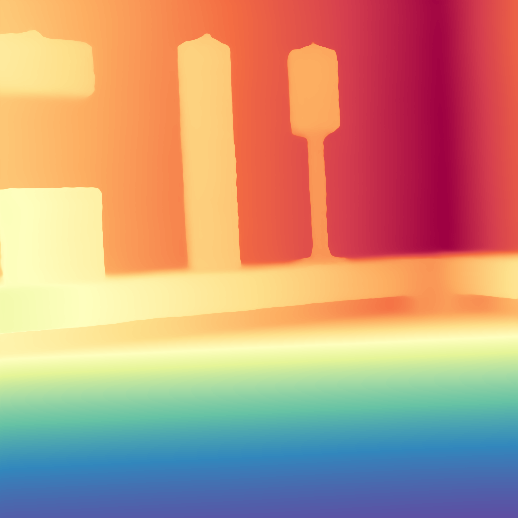}
& \includegraphics[width=\dpwidth]{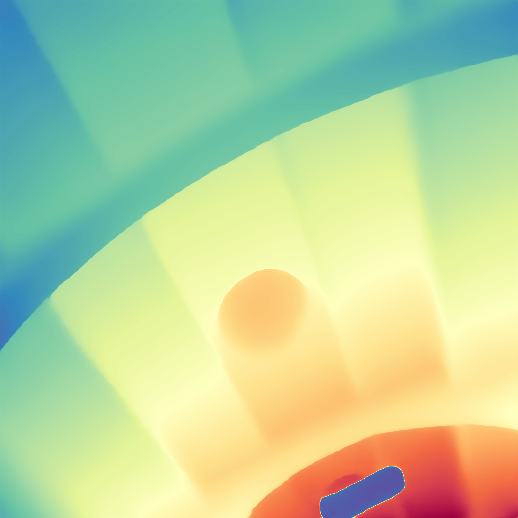}
& \includegraphics[width=\dpwidth]{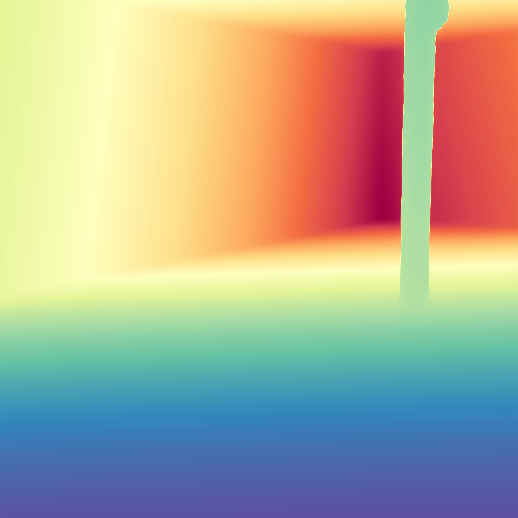}
& \includegraphics[width=\dpwidth]{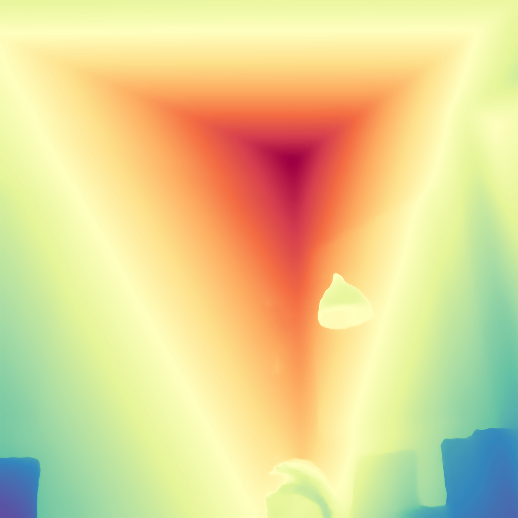}
& \includegraphics[width=\dpwidth]{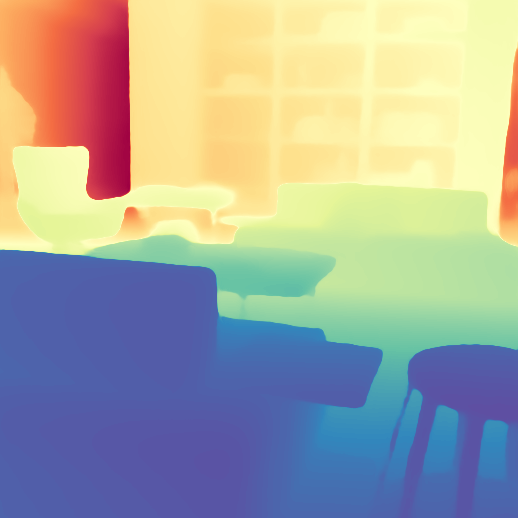} \\

4\_11\_17
& \includegraphics[width=\dpwidth]{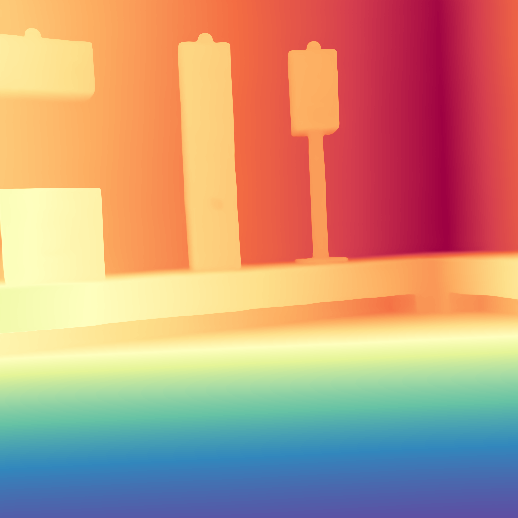}
& \includegraphics[width=\dpwidth]{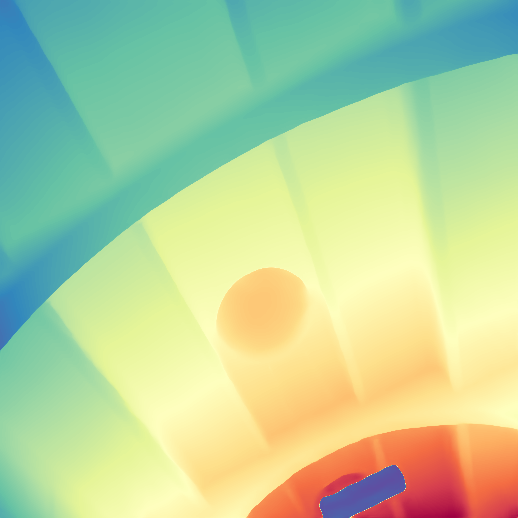}
& \includegraphics[width=\dpwidth]{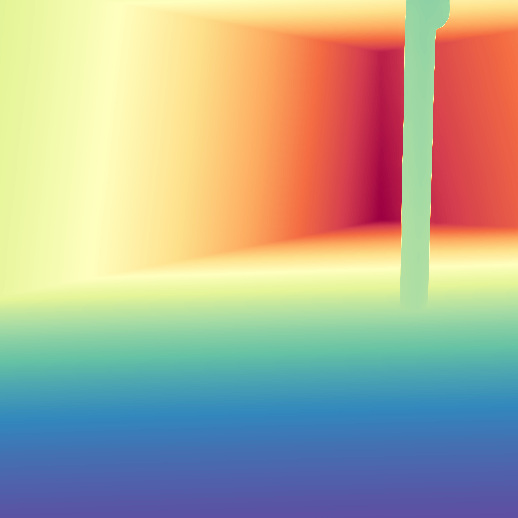}
& \includegraphics[width=\dpwidth]{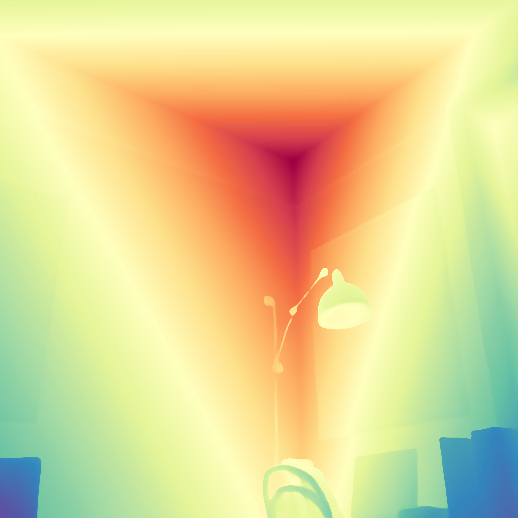}
& \includegraphics[width=\dpwidth]{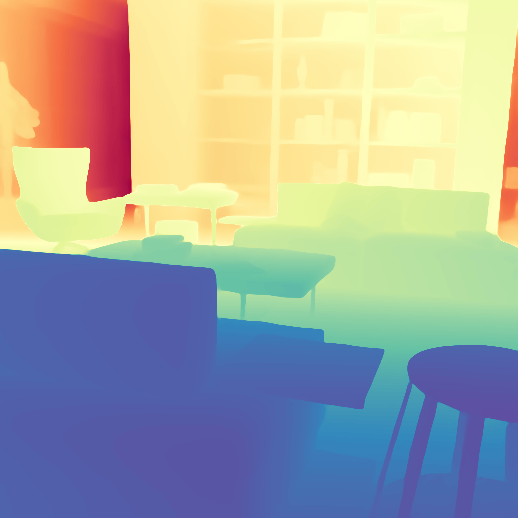} \\

4\_11\_23
& \includegraphics[width=\dpwidth]{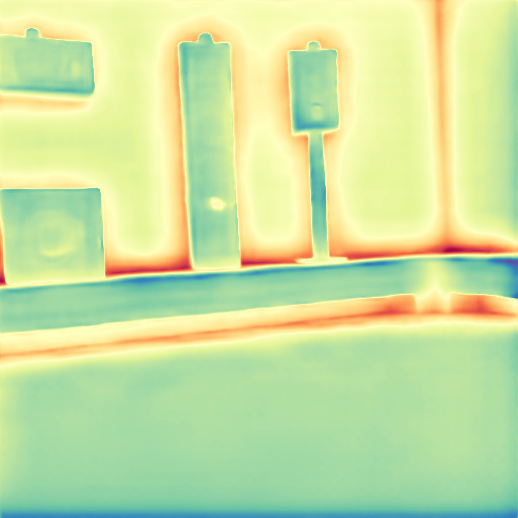}
& \includegraphics[width=\dpwidth]{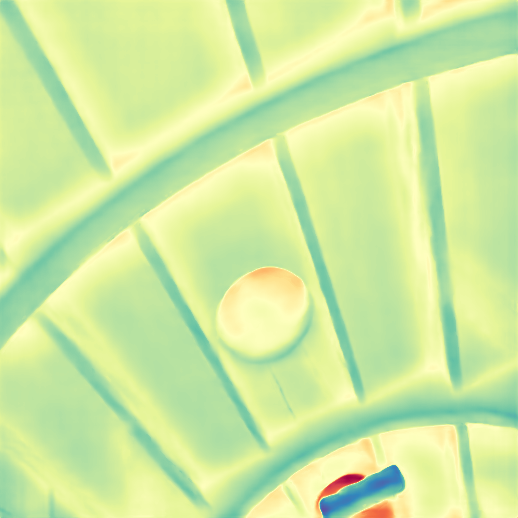}
& \includegraphics[width=\dpwidth]{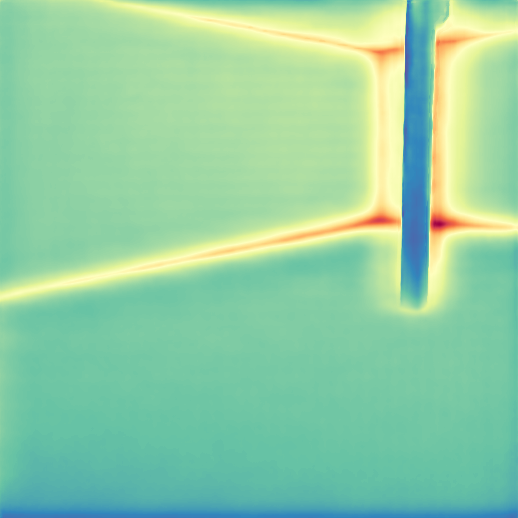}
& \includegraphics[width=\dpwidth]{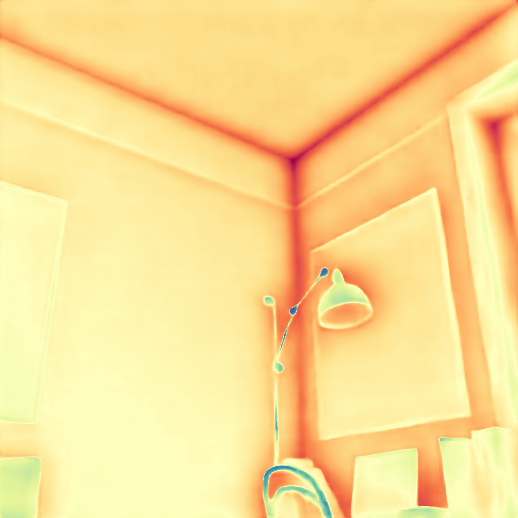}
& \includegraphics[width=\dpwidth]{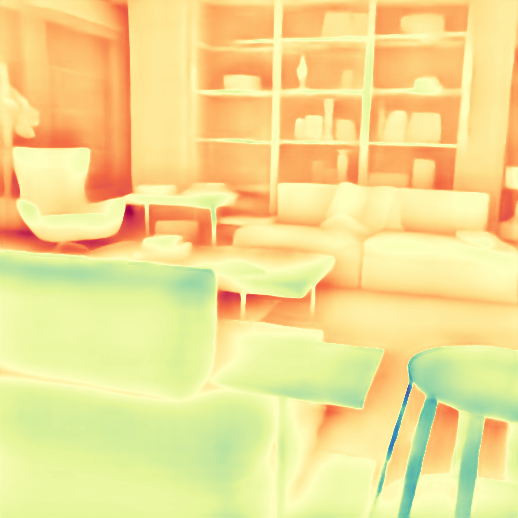} \\

4\_17\_23
& \includegraphics[width=\dpwidth]{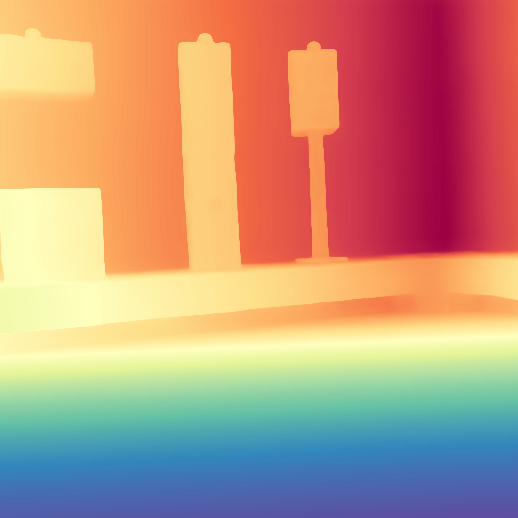}
& \includegraphics[width=\dpwidth]{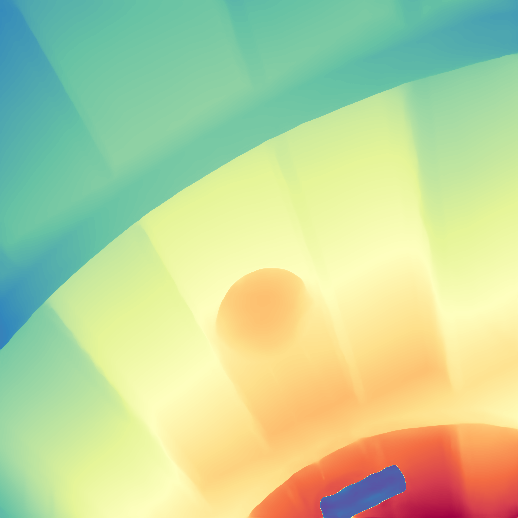}
& \includegraphics[width=\dpwidth]{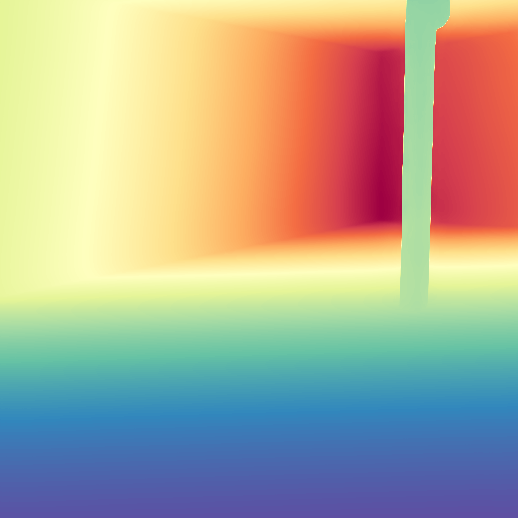}
& \includegraphics[width=\dpwidth]{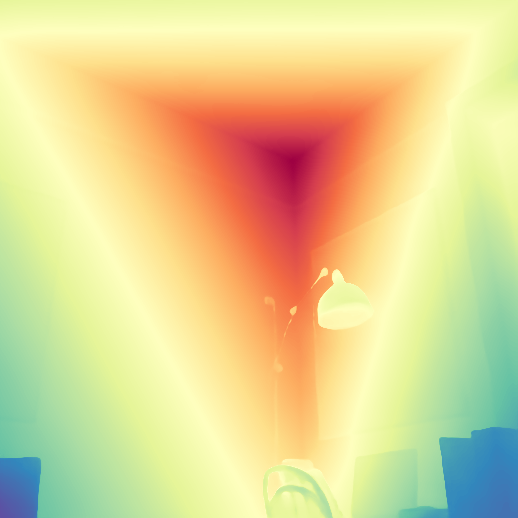}
& \includegraphics[width=\dpwidth]{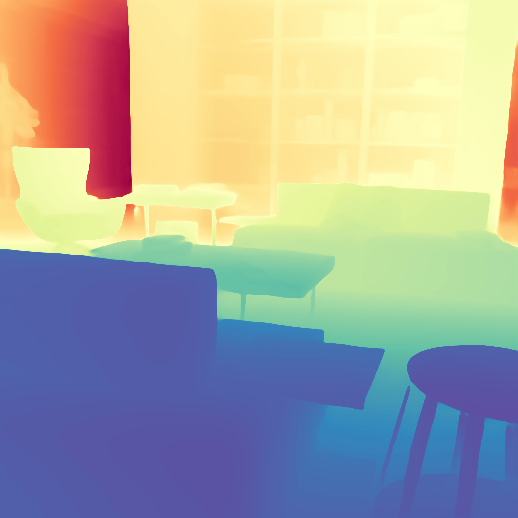} \\

All Layers
& \includegraphics[width=\dpwidth]{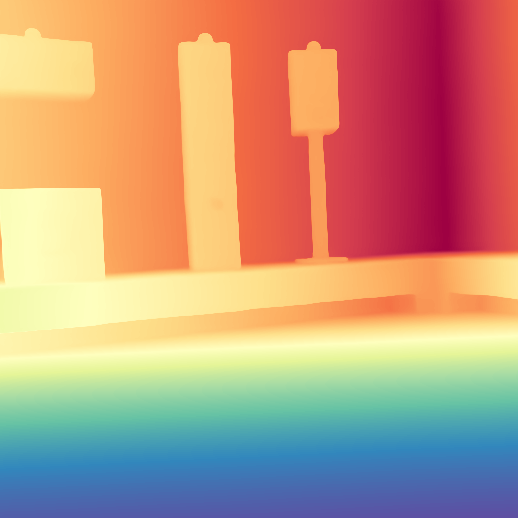}
& \includegraphics[width=\dpwidth]{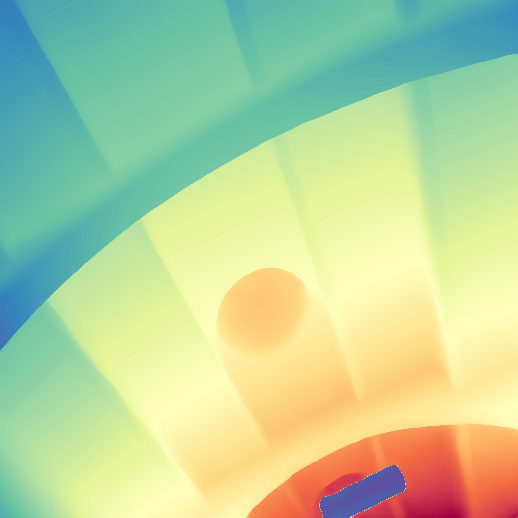}
& \includegraphics[width=\dpwidth]{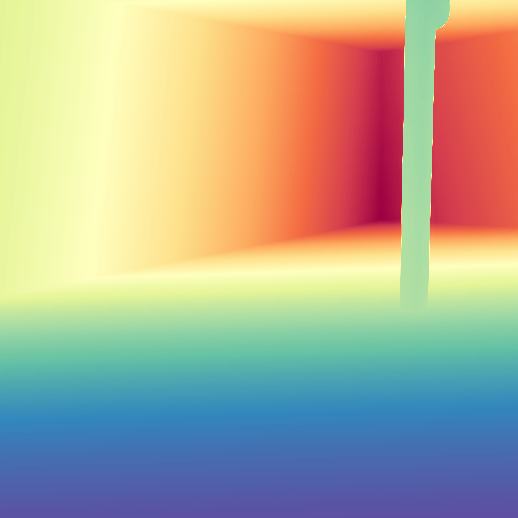}
& \includegraphics[width=\dpwidth]{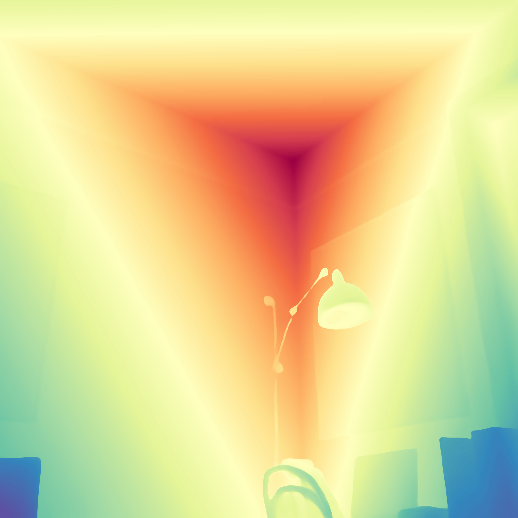}
& \includegraphics[width=\dpwidth]{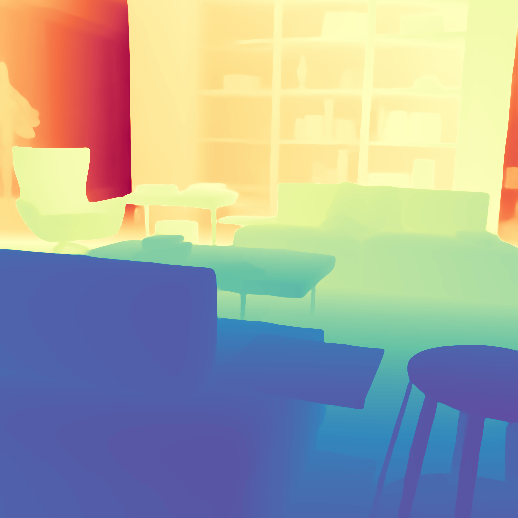} \\

\bottomrule
\end{tabular}
}
\label{worldmirror_layer_analysis}
\end{figure*}

\clearpage

\section{More Qualitative Results}

\begin{figure}[th]
    \centering
    \includegraphics[page=1,width=\textwidth]{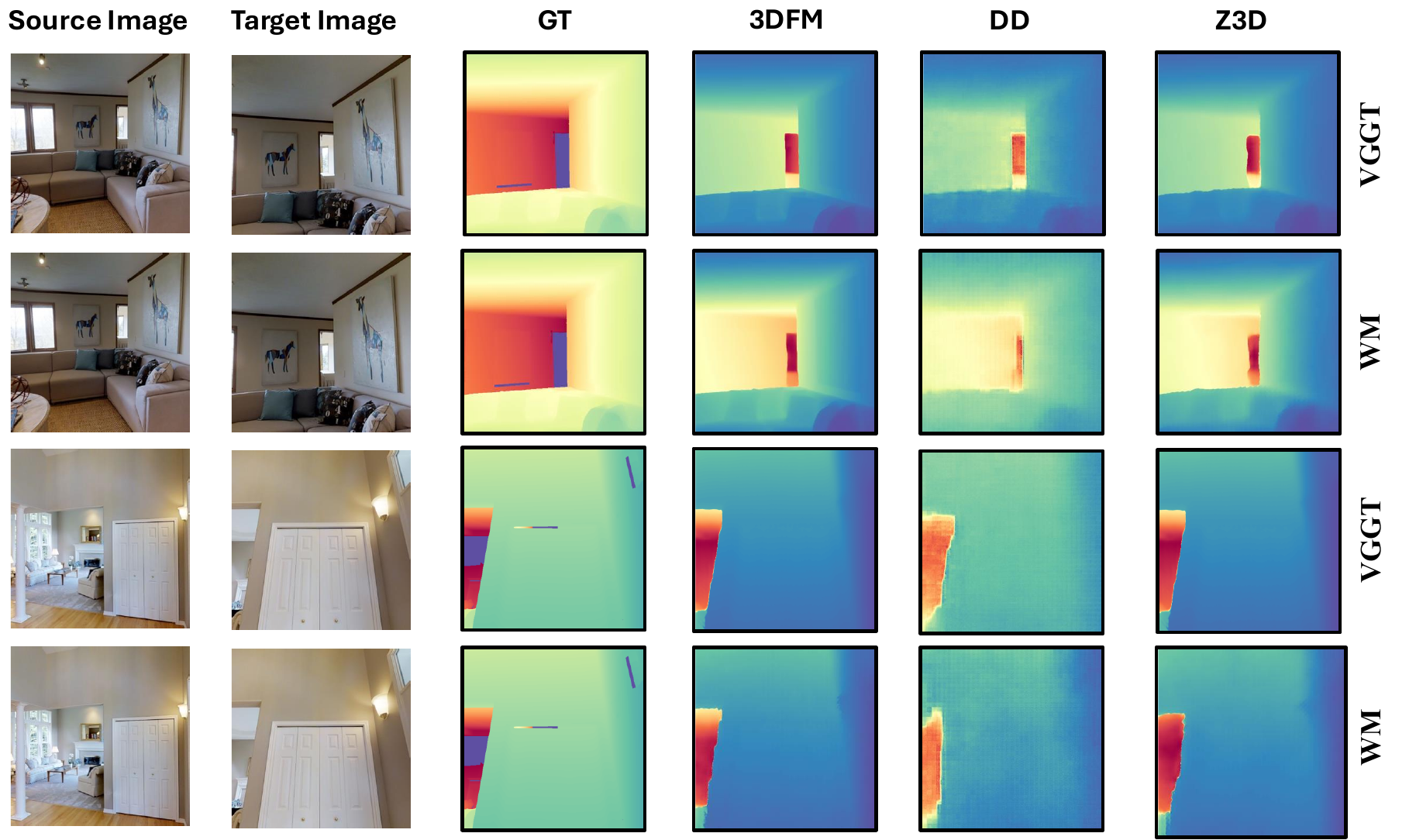}
       \caption{Qualitative comparison of Novel View Depth Predictions. Columns show: source RGB, target RGB, ground truth depth (GT), 3DFM predicted depth, Depth Diffusion (DD) prediction, and Z3D prediction. Z3D results are substantially more smooth as compared to results from a method that works directly in patch-space.}
    
    \label{fig:1-1_compare-1-depth}
\end{figure}

\begin{figure}[th]
    \centering
    \includegraphics[page=1,width=\textwidth]{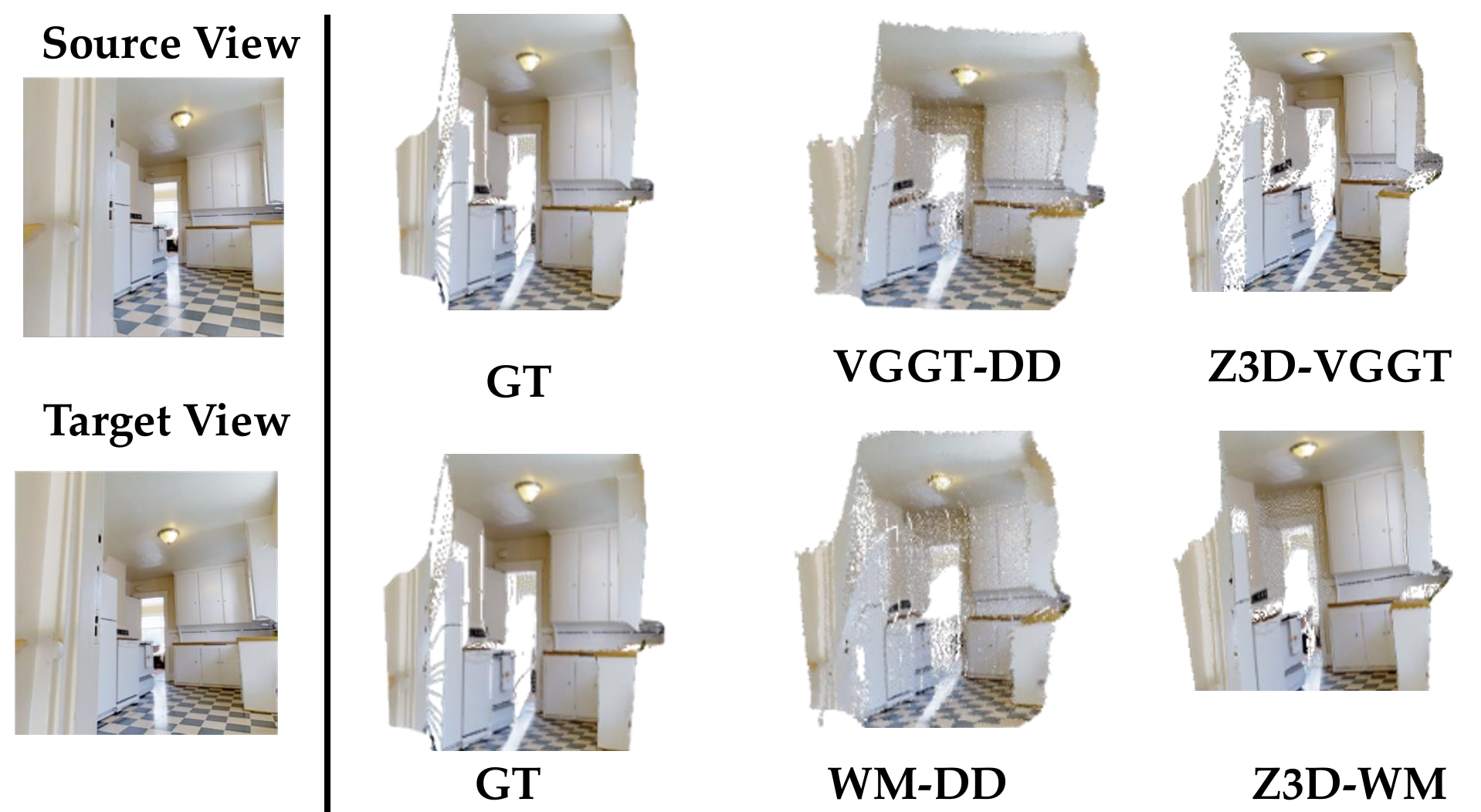}
   \caption{Qualitative comparison of novel-view point cloud reconstructions from a single source image and target pose. Reconstructions from \ZThreeDVGGTDD{} and \ZThreeDWMDD{} exhibit higher noise and reduced geometric sharpness, whereas \ZThreeDVGGT{} and \ZThreeDWM{} produce cleaner and more coherent 3D structures. Top left: source view (left of the vertical line). Bottom left: target view (not observed by the model). First row: Predictions of VGGT , \ZThreeDVGGTDD{} and \ZThreeDVGGT{} respectively. Second row: Predictions of WM , \ZThreeDWMDD{} and \ZThreeDWM{} respectively.
}
    \label{fig:1-1_compare-2}
\end{figure}

\begin{figure}[th]
    \centering
    \includegraphics[page=1,width=\textwidth]{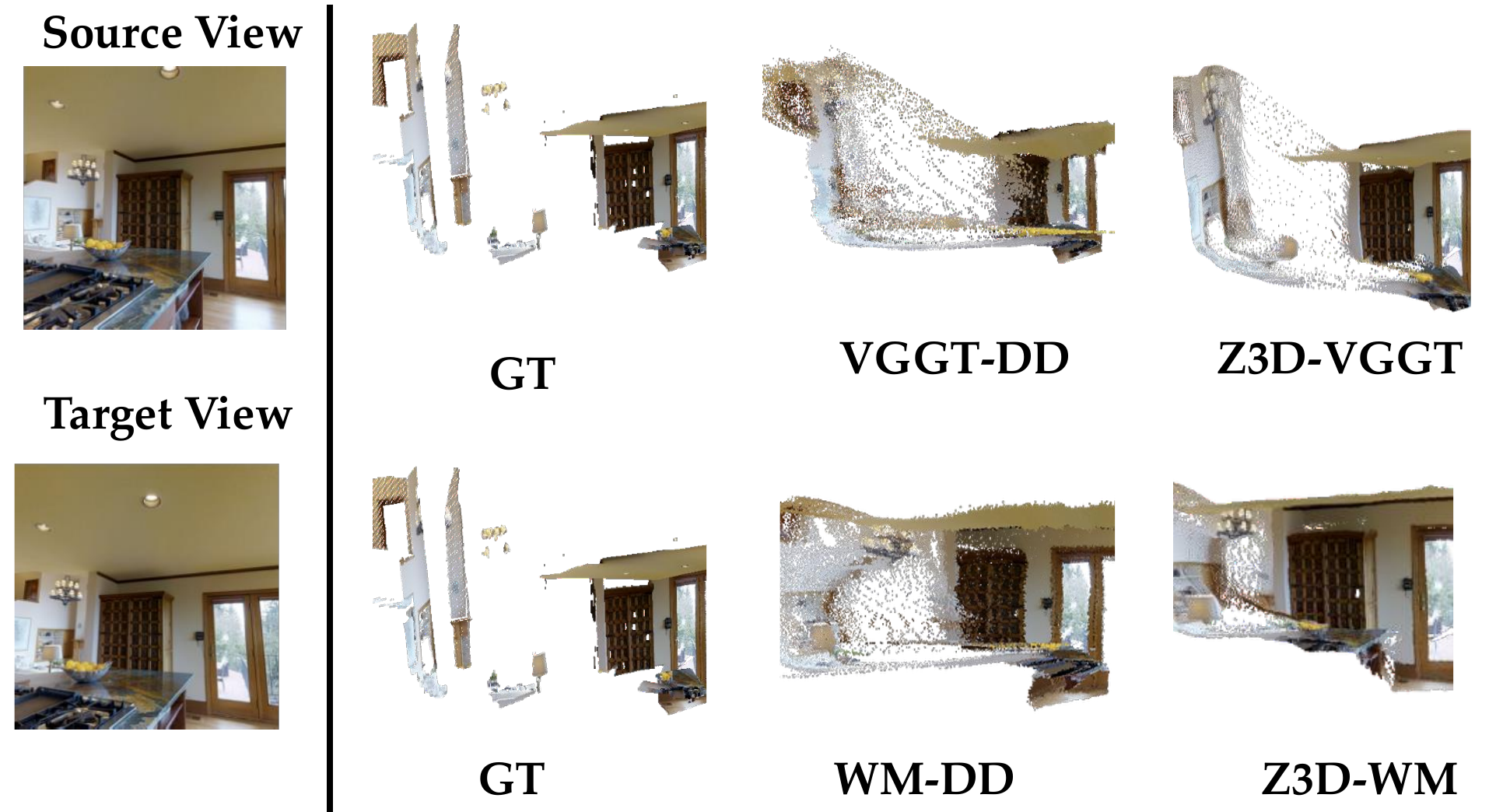}
    \caption{Qualitative comparison of novel-view point cloud reconstructions from a single source image and target pose. Reconstructions from \ZThreeDVGGTDD{} and \ZThreeDWMDD{} exhibit higher noise and reduced geometric sharpness, whereas \ZThreeDVGGT{} and \ZThreeDWM{} produce cleaner and more coherent 3D structures. Top left: source view (left of the vertical line). Bottom left: target view (not observed by the model). First row: Predictions of VGGT , \ZThreeDVGGTDD{} and \ZThreeDVGGT{} respectively. Second row: Predictions of WM, \ZThreeDWMDD{} and \ZThreeDWM{} respectively.
}
    \label{fig:1-1_compare-3}
\end{figure}

\begin{figure}[th]
    \centering
    \includegraphics[page=1,width=\textwidth]{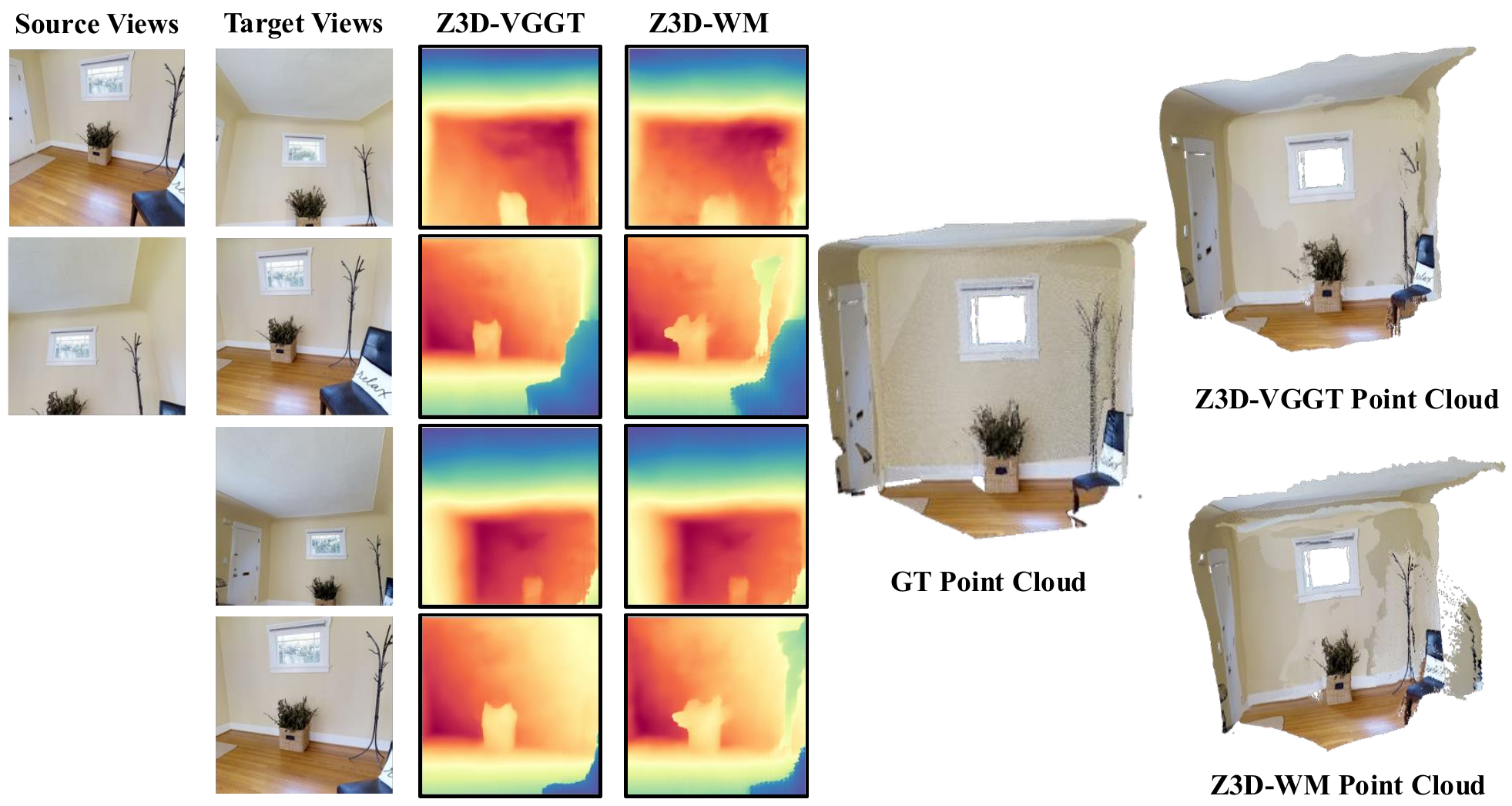}
    \caption{Novel-view depth prediction from 2 source views $\rightarrow$ 4 target views. Conditioned on the target camera poses, \ZThreeDVGGT{} and \ZThreeDWM{} generate geometrically consistent depth maps across all target views, resulting in a coherent reconstructed point cloud.}
      \label{fig:2-4_compare-1-example1}
\end{figure}

\begin{figure}[th]
    \centering
    \includegraphics[page=1,width=\textwidth]{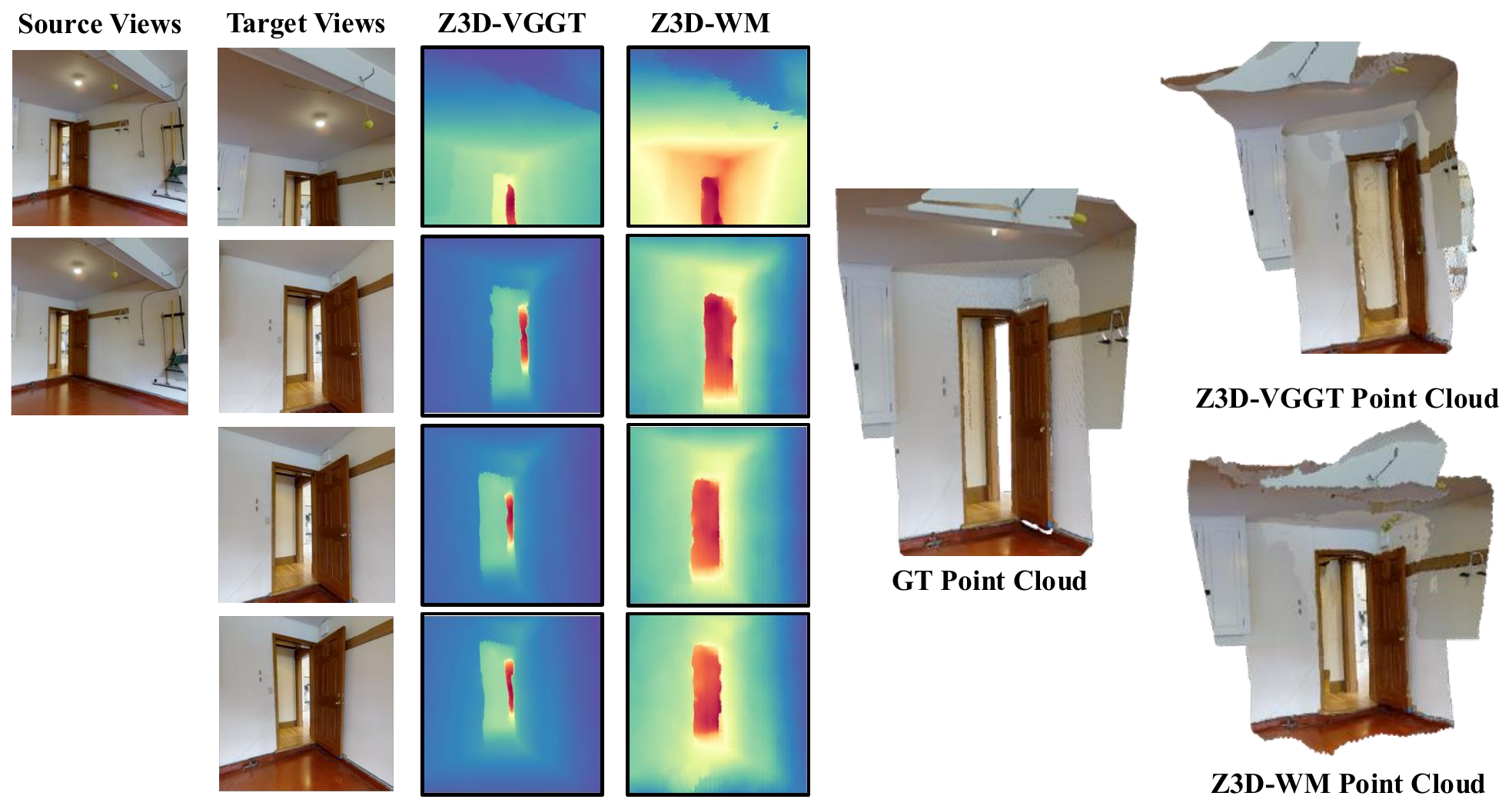}
    \caption{Novel-view depth prediction from 2 source views $\rightarrow$ 4 target views. Conditioned on the target camera poses, \ZThreeDVGGT{} and \ZThreeDWM{} generate geometrically consistent depth maps across all target views, resulting in a coherent reconstructed point cloud.}
        \label{fig:2-4_compare-1-example2}
\end{figure}

We provide additional qualitative results, comparing our model with our depth-diffusion (DD)-based baselines (\ZThreeDWMDD{} and \ZThreeDVGGTDD{}).
Figure \ref{fig:1-1_compare-1-depth} shows additional examples of novel depth inference using Z3D and the DD baselines.Z3D produces substantially smoother results compared to methods operating directly in patch space. Figures \ref{fig:1-1_compare-2} and \ref{fig:1-1_compare-3} show 3D point cloud projections of the predicted depth maps under the 1 source $\to$ 1 target setting. Reconstructions from \ZThreeDVGGTDD{} and \ZThreeDWMDD{} exhibit higher noise and reduced geometric sharpness, whereas \ZThreeDVGGT{} and \ZThreeDWM{} produce cleaner and more coherent 3D structures.  Figures \ref{fig:2-4_compare-1-example1} and \ref{fig:2-4_compare-1-example2} show the combined 3D point cloud projections of all predicted maps in the 2-source views $\to$ 4 target views setting. When there is sufficient overlap across views, the models (\ZThreeDVGGT{} and \ZThreeDWM{}) excel at novel view depth prediction, producing high-fidelity point clouds.

% --- Page 1 ---

% --- Page 2 ---

% --- Page 1 ---

\clearpage

\section{More Quantitative Results}
We provide additional quantitative evaluation using the depths predicted by the 3DFMs as pseudo ground truth. Specifically, the target-view depth maps produced by the 3DFMs are treated as reference depths, and all methods are evaluated against them using the same depth and point cloud metrics. We report results on both \textbf{in-domain} data (test split of the training datasets) and \textbf{out-of-domain}
datasets (used only for evaluation and never seen during training). This experiment evaluates whether the Z3D architecture learns scene geometry consistent with the geometric priors encoded by the 3DFMs.

Tab.~\ref{tab:pointcloud_metrics-3dfm-1-1} and Tab.~\ref{tab:depth_metrics-3dfm-1-1} report the point cloud and depth results, respectively, for the 1 source $\rightarrow$ 1 target setting. Tab.~\ref{tab:pointcloud_metrics-3dfm-2-4} and Tab.~\ref{tab:depth_metrics-3dfm-2-4} present the corresponding results for the 2 source $\rightarrow$ 4 target setting. Across both settings, Z3D predictions closely match the target-view depths produced by the foundation models, indicating that Z3D learns geometric representations that are highly consistent with the priors encoded by the 3DFMs.

\begin{table}[th]
\centering
\caption{1 source $\to$ 1 target Point Cloud Metrics (Mean / Median). Best results per dataset are shown in \textbf{bold}, second best are \underline{underlined}. M: Mean, Md: Median}
\label{tab:pointcloud_metrics-3dfm-1-1}
\setlength{\tabcolsep}{1.8pt}
\resizebox{\columnwidth}{!}{%
\begin{tabular}{
l
*{4}{c}
*{12}{>{\columncolor{oodblue}}c}
}
\toprule

& \multicolumn{4}{c}{~}
& \multicolumn{12}{c}{\cellcolor{oodblue}\bf Out of Domain}
\\

& \multicolumn{4}{c}{\textbf{In-Domain}}
& \multicolumn{4}{c}{\cellcolor{oodblue}\textbf{DTU}}
& \multicolumn{4}{c}{\cellcolor{oodblue}\textbf{7-Scenes}}
& \multicolumn{4}{c}{\cellcolor{oodblue}\textbf{NRGBD}} \\

\cmidrule(lr){2-5}
\cmidrule(lr){6-9}
\cmidrule(lr){10-13}
\cmidrule(lr){14-17}

\textbf{Model}
& \multicolumn{2}{c}{Acc $\downarrow$}
& \multicolumn{2}{c}{Comp $\downarrow$}
& \multicolumn{2}{c}{\cellcolor{oodblue}Acc $\downarrow$}
& \multicolumn{2}{c}{\cellcolor{oodblue}Comp $\downarrow$}
& \multicolumn{2}{c}{\cellcolor{oodblue}Acc $\downarrow$}
& \multicolumn{2}{c}{\cellcolor{oodblue}Comp $\downarrow$}
& \multicolumn{2}{c}{\cellcolor{oodblue}Acc $\downarrow$}
& \multicolumn{2}{c}{\cellcolor{oodblue}Comp $\downarrow$} \\

& M & Md & M & Md
& M & Md & M & Md
& M & Md & M & Md
& M & Md & M & Md \\

\midrule

\ZThreeDVGGTDD
& 0.079 & \underline{0.039} & \underline{0.046} & \textbf{0.013}
& 7.105 & 5.487 & \textbf{1.612} & \textbf{1.144}
& 0.027 & \underline{0.022} & \underline{0.011} & \underline{0.008}
& 
\underline{0.113} & 0.059 & \underline{0.090} &\underline{0.021} \\

\ZThreeDWMDD
& 0.130 & 0.053 & 0.068 & \textbf{0.013}
& 10.323 & 7.610 & \underline{1.672} & \underline{1.311}
& 0.028 & 0.023 & \textbf{0.009} & \textbf{0.007}
& \textbf{0.120} & 0.063 & \textbf{0.063} & \textbf{0.016} \\

\ZThreeDVGGT
& \underline{0.046} & \textbf{0.016} & 0.049 & \underline{0.016}
& \underline{2.924} & \underline{2.044} & 2.574 & 1.937
& \underline{0.012} & \textbf{0.009} & 0.013 & 0.009
& 0.122 & \underline{0.036} & 0.119 & 0.038 \\

\ZThreeDWM
& \textbf{0.042} & \textbf{0.016} & \textbf{0.043} & \underline{0.016}
& \textbf{2.644} & \textbf{1.777} & 2.314 & 1.707
& \textbf{0.011} & \textbf{0.009} & \underline{0.011} & \underline{0.008}
& 0.117 & \textbf{0.029} & 0.105 & 0.028 \\

\bottomrule
\end{tabular}%
}
\end{table}
\begin{table}[th]
\centering
\caption{Depth Estimation Metrics (AbsRel and $\delta < 1.25$) for 1 Source $\to$ 1 source using 3DFM predictions as GT. Best results per dataset in \textbf{bold}, second best \underline{underlined}.}
\label{tab:depth_metrics-3dfm-1-1}

\resizebox{\textwidth}{!}{%
\begin{tabular}{
l
cc
>{\columncolor{oodblue}}c
>{\columncolor{oodblue}}c
>{\columncolor{oodblue}}c
>{\columncolor{oodblue}}c
>{\columncolor{oodblue}}c
>{\columncolor{oodblue}}c
}
\toprule

& \multicolumn{2}{c}{~}
& \multicolumn{6}{>{\columncolor{oodblue}}c}{\textbf{Out-of-Domain}} \\

\cmidrule(lr){2-3}
\cmidrule(lr){4-9}

& \multicolumn{2}{c}{\textbf{In-Domain}}
& \multicolumn{2}{c}{\cellcolor{oodblue}\textbf{DTU}}
& \multicolumn{2}{c}{\cellcolor{oodblue}\textbf{7-Scenes}}
& \multicolumn{2}{c}{\cellcolor{oodblue}\textbf{NRGBD}} \\

\cmidrule(lr){2-3}
\cmidrule(lr){4-5}
\cmidrule(lr){6-7}
\cmidrule(lr){8-9}

\textbf{Model}
& AbsRel$\downarrow$ & $\delta<1.25$ $\uparrow$
& AbsRel$\downarrow$ & $\delta<1.25$ $\uparrow$
& AbsRel$\downarrow$ & $\delta<1.25$ $\uparrow$
& AbsRel$\downarrow$ & $\delta<1.25$ $\uparrow$ \\

\midrule

\ZThreeDVGGTDD
& 0.083 & 0.937
& \underline{0.022} & \underline{0.998}
& 0.033 & 0.993
& 0.737 & 0.727 \\

\ZThreeDWMDD
& 0.108 & 0.902
& 0.034 & 0.997
& 0.031 & 0.992
& \underline{0.112} & \textbf{0.871} \\

\ZThreeDVGGT
& \underline{0.050} & \underline{0.960}
& \textbf{0.012} & \textbf{0.999}
& \underline{0.016} & \underline{0.995}
& 0.451 & 0.730 \\

\ZThreeDWM
& \textbf{0.046} & \textbf{0.966}
& \textbf{0.012} & \textbf{0.999}
& \textbf{0.014} & \textbf{0.996}
& \textbf{0.105} & \underline{0.860} \\

\bottomrule
\end{tabular}%
}
\end{table}

\begin{table}[th]
\centering
\caption{2 source $\to$ 4 target Point Cloud Metrics (Mean / Median). Best results per dataset are shown in \textbf{bold}, second best are \underline{underlined}.}
\label{tab:pointcloud_metrics-3dfm-2-4}
\setlength{\tabcolsep}{1.8pt}
\resizebox{\columnwidth}{!}{%
\begin{tabular}{
l
*{4}{c}
*{12}{>{\columncolor{oodblue}}c}
}
\toprule

& \multicolumn{4}{c}{~}
& \multicolumn{12}{c}{\cellcolor{oodblue}\bf Out of Domain}
\\

& \multicolumn{4}{c}{\textbf{In-Domain}}
& \multicolumn{4}{c}{\cellcolor{oodblue}\textbf{DTU}}
& \multicolumn{4}{c}{\cellcolor{oodblue}\textbf{7-Scenes}}
& \multicolumn{4}{c}{\cellcolor{oodblue}\textbf{NRGBD}} \\

\cmidrule(lr){2-5}
\cmidrule(lr){6-9}
\cmidrule(lr){10-13}
\cmidrule(lr){14-17}

\textbf{Model}
& \multicolumn{2}{c}{Acc $\downarrow$}
& \multicolumn{2}{c}{Comp $\downarrow$}
& \multicolumn{2}{c}{\cellcolor{oodblue}Acc $\downarrow$}
& \multicolumn{2}{c}{\cellcolor{oodblue}Comp $\downarrow$}
& \multicolumn{2}{c}{\cellcolor{oodblue}Acc $\downarrow$}
& \multicolumn{2}{c}{\cellcolor{oodblue}Comp $\downarrow$}
& \multicolumn{2}{c}{\cellcolor{oodblue}Acc $\downarrow$}
& \multicolumn{2}{c}{\cellcolor{oodblue}Comp $\downarrow$} \\

& M & Md & M & Md
& M & Md & M & Md
& M & Md & M & Md
& M & Md & M & Md \\

\midrule

\ZThreeDVGGTDD
& 0.196 & 0.120 & 0.060 & 0.017
& 17.841 & 13.084 & \textbf{1.643} & \textbf{1.043}
& \underline{0.028} & 0.021 & \underline{0.009} & \textbf{0.005}
& 0.083 & 0.041 & \underline{0.077} & \textbf{0.014} \\

\ZThreeDWMDD
& \textbf{0.031} & 0.053 & 0.068 & \textbf{0.013}
& 10.323 & 7.610 & \underline{1.672} & \underline{1.311}
& \underline{0.028} & 0.023 &\underline{0.009} & \underline{0.007}
& 0.120 & 0.063 & \textbf{0.063} & \underline{0.016} \\

\ZThreeDVGGT
& \underline{0.067} & \textbf{0.026} & \textbf{0.045} & \underline{0.016}
& \underline{6.584} & \textbf{4.483} & 2.630 & 1.762
& \textbf{0.011} & \underline{0.008} & \underline{0.009} & \textbf{0.005}
& \underline{0.070} & \underline{0.026} & 0.079 & 0.025 \\

\ZThreeDWM
& \underline{0.067} & \underline{0.027} & \underline{0.047} & 0.017
& \textbf{9.069} & \underline{5.385} & 3.375 & 2.245
& \textbf{0.011} & \textbf{0.007} & \textbf{0.008} & \textbf{0.005}
& \textbf{0.068} & \textbf{0.023} & 0.084 & 0.021 \\

\bottomrule
\end{tabular}%
}
\end{table}

\begin{table}[th]
\centering
\caption{Depth Estimation Metrics (AbsRel and $\delta < 1.25$). Best results per dataset in \textbf{bold}, second best \underline{underlined}.}
\label{tab:depth_metrics-3dfm-2-4}

\resizebox{\textwidth}{!}{%
\begin{tabular}{
l
cc
>{\columncolor{oodblue}}c
>{\columncolor{oodblue}}c
>{\columncolor{oodblue}}c
>{\columncolor{oodblue}}c
>{\columncolor{oodblue}}c
>{\columncolor{oodblue}}c
}
\toprule

& \multicolumn{2}{c}{~}
& \multicolumn{6}{>{\columncolor{oodblue}}c}{\textbf{Out-of-Domain}} \\

\cmidrule(lr){2-3}
\cmidrule(lr){4-9}

& \multicolumn{2}{c}{\textbf{In-Domain}}
& \multicolumn{2}{c}{\cellcolor{oodblue}\textbf{DTU}}
& \multicolumn{2}{c}{\cellcolor{oodblue}\textbf{7-Scenes}}
& \multicolumn{2}{c}{\cellcolor{oodblue}\textbf{NRGBD}} \\

\cmidrule(lr){2-3}
\cmidrule(lr){4-5}
\cmidrule(lr){6-7}
\cmidrule(lr){8-9}

\textbf{Model}
& AbsRel$\downarrow$ & $\delta<1.25$ $\uparrow$
& AbsRel$\downarrow$ & $\delta<1.25$ $\uparrow$
& AbsRel$\downarrow$ & $\delta<1.25$ $\uparrow$
& AbsRel$\downarrow$ & $\delta<1.25$ $\uparrow$ \\

\midrule

\ZThreeDVGGTDD
& 0.286 & 0.604
& 0.144 & 0.780
& 0.043 & \underline{0.986}
& 0.542 & 0.686 \\

\ZThreeDWMDD
& \underline{0.108} & \textbf{0.902}
& \underline{0.034} & \textbf{0.997}
& 0.031 & 0.992
& \underline{0.112} & \textbf{0.871} \\

\ZThreeDVGGT
& \textbf{0.110} & \underline{0.891}
& 0.055 & 0.992
& \underline{0.024} & 0.991
& 0.460 & 0.731 \\

\ZThreeDWM
& 0.118 & 0.879
& \textbf{0.070} & \underline{0.966}
& \textbf{0.022} & \textbf{0.992}
& \textbf{0.115} & \underline{0.841} \\

\bottomrule
\end{tabular}%
}
\end{table}

\clearpage

\section{Limitations of Z3D}

While Z3D advances novel depth synthesis by leveraging 3DFM scene representations, our method has some limitations. First, we observed that when there is limited overlap between source and target views, Z3D struggles to predict fine-grained depths for novel views. In our setup, the first image in the source views is treated as the reference camera. Previous work \cite{wang2025pi} on 3DFMs have shown that the choice of reference view significantly impacts 3DFM performance. Since Z3D builds on 3DFMs, it inherits their limitations: predicting  3D geometry is challenging when scene images do not have sufficient overlap. As illustrated in Figures~\ref{fig:2-4_compare-4-z3d-limitations}  the source views have limited overlap with the target views, resulting in both \ZThreeDVGGT{} and \ZThreeDWM{} struggling to produce reasonable depth maps for the novel views.

% --- Page 1 ---
\begin{figure}[th]
    \centering
    \includegraphics[page=1,width=\textwidth]{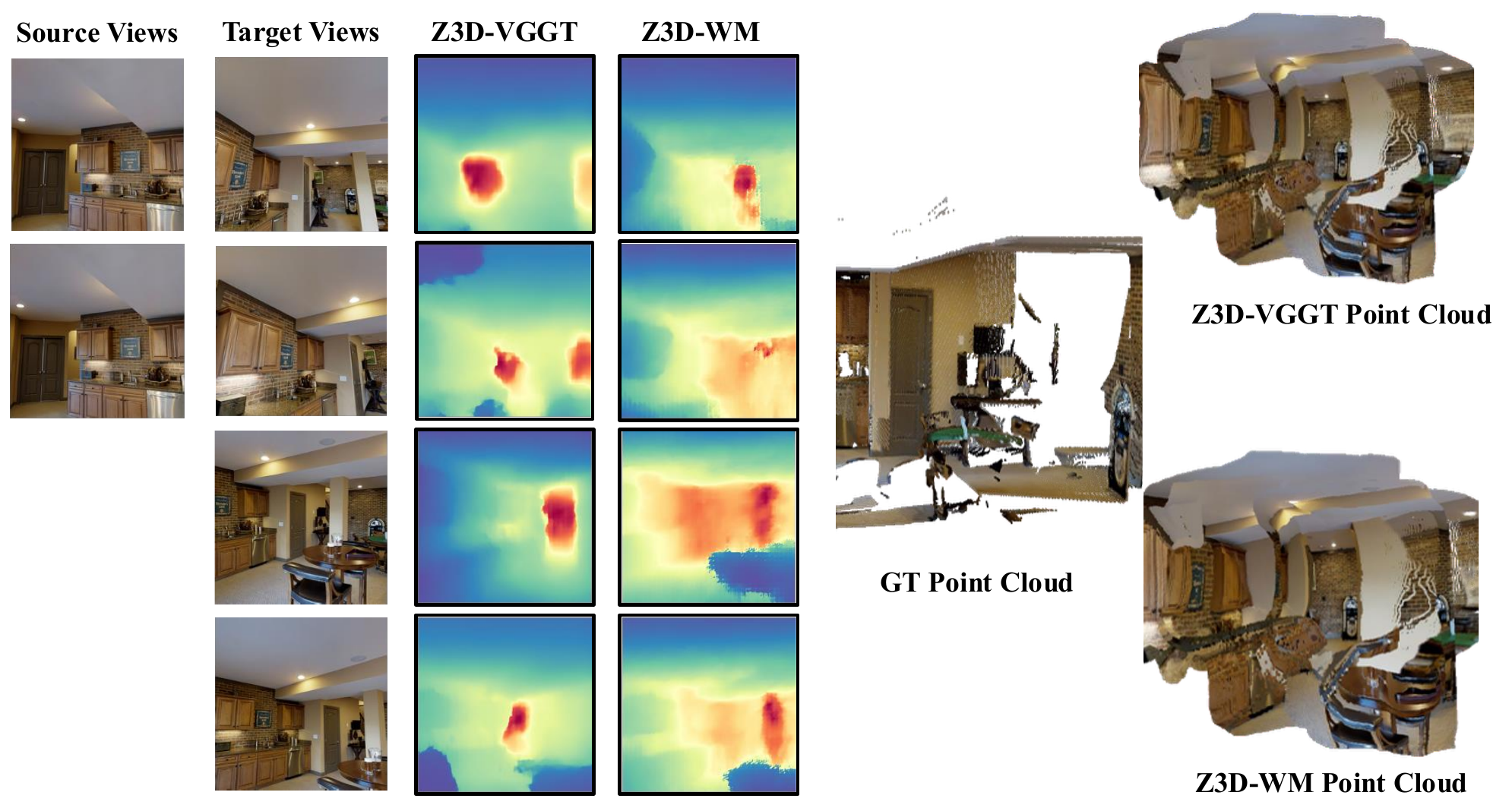}
    \caption{Point maps obtained from the depth maps predicted by \ZThreeDVGGT{} and \ZThreeDWM{}. Both \ZThreeDVGGT{} and \ZThreeDWM{} struggles to predict depths when source views have limited overlap with target views. The depth maps from \ZThreeDVGGT{} and \ZThreeDWM{} preserve the overall scene structure, but finer details are missing, and the resulting projected point clouds is not consistent.}
    \label{fig:2-4_compare-4-z3d-limitations}
    
\end{figure}

\clearpage  % TODO FINAL: This \clearpage needs to be removed from both review and camera-ready versions.

% % ---- Bibliography ----
% %
% % BibTeX users should specify bibliography style 'splncs04'.
% % References will then be sorted and formatted in the correct style.
% \bibliography{main}

\end{document}